\documentclass{article}

\PassOptionsToPackage{numbers, compress}{natbib} 

\usepackage[preprint]{neurips_2026}

\usepackage{amsmath,amsfonts,bm}

\def\eqref#1{equation~\ref{#1}}

\def\1{\bm{1}}

\DeclareMathAlphabet{\mathsfit}{\encodingdefault}{\sfdefault}{m}{sl}
\SetMathAlphabet{\mathsfit}{bold}{\encodingdefault}{\sfdefault}{bx}{n}

\usepackage[utf8]{inputenc} 
\usepackage[T1]{fontenc}    
\usepackage{hyperref}       
\usepackage{url}            
\usepackage{booktabs}       
\usepackage{amsfonts}       
\usepackage{nicefrac}       
\usepackage{microtype}      
\usepackage{xcolor}         
\usepackage{makecell}

\title{Adversarial Graph Neural Network Benchmarks: Towards Practical and Fair Evaluation}

\usepackage{microtype}
\usepackage{graphicx}
\usepackage{subfigure}
\usepackage{booktabs} 

\usepackage{algorithm}
\usepackage[noend]{algorithmic}

\usepackage{amsmath}
\usepackage{amssymb}
\usepackage{mathtools}
\usepackage{amsthm}

\usepackage[capitalize,noabbrev]{cleveref}

\theoremstyle{plain}

\theoremstyle{definition}

\theoremstyle{remark}

\usepackage[textsize=tiny]{todonotes}

\usepackage{multirow}       
\usepackage{lscape}
\usepackage{pdflscape}
\usepackage{csquotes}
\usepackage{wrapfig}
\usepackage{xspace}

\usepackage{hyperref}

\author{
\textbf{Tran Gia Bao Ngo}\textsuperscript{1}\quad
\textbf{Zulfikar Alom}\textsuperscript{2} \quad
\textbf{Federico Errica}\textsuperscript{3}\\
\textbf{Murat Kantarcioglu}\textsuperscript{4} \quad
\textbf{Cuneyt Gurcan Akcora}\textsuperscript{5} \\
\textsuperscript{1}Department of Computer Science, University of Manitoba,
\textsuperscript{2}University of Toledo, \\
\textsuperscript{3}NEC Laboratories Europe, 
\textsuperscript{4}Department of Computer Science, Virginia Tech, USA, \\
\textsuperscript{5}AI Initiative - University of Central Florida
}

\begin{document}
\maketitle


\newcommand{\etc}{et al.\xspace}
\renewcommand{\vec}[1]{\ensuremath{\mathbf{#1}}}
\newcommand{\vecs}[1]{\ensuremath{\mathbf{\boldsymbol{#1}}}}
\newcommand{\mat}[1]{\ensuremath{\mathbf{#1}}}
\newcommand{\mats}[1]{\ensuremath{\mathbf{\boldsymbol{#1}}}}
\newcommand{\ten}[1]{\mat{\ensuremath{\boldsymbol{\mathcal{#1}}}}}

\newcommand{\bao}[1]{{{\textcolor{blue}{\textbf{Bao: }}}{\textcolor{blue}{#1}}}}
\newcommand{\federico}[1]{{{\textcolor{blue}{\textbf{Federico: }}}{\textcolor{blue}{#1}}}}
\newcommand{\cuneyt}[1]{{{\textcolor{red}{\textbf{Cuneyt: }}}{\textcolor{red}{#1}}}}
\newcommand{\murat}[1]{{{\textcolor{blue}{\textbf{Murat: }}}{\textcolor{blue}{#1}}}}
\newcommand{\zulfikar}[1]{{{\textcolor{blue}{\textbf{Zulfikar: }}}{\textcolor{blue}{#1}}}}

\newcommand{\negative}[1]{{{\textcolor{red}{#1}}}}
\newcommand{\positive}[1]{{{\textcolor{blue}{#1}}}}

\newcommand{\propseRND}{L$^1$D-RND\xspace}
\newcommand{\cora}{CORA\xspace}
\newcommand{\citeseer}{CITESEER\xspace}
\newcommand{\pubmed}{PUBMED\xspace}

\newcommand{\squirrel}{SQUIRREL\xspace}
\newcommand{\chameleon}{CHAMELEON\xspace}
\newcommand{\ogb}{OGB-ARXIV\xspace}
\newcommand{\prbcdNA}{PR-BCD (NA)\xspace}

\newcommand{\note}[1]{{{\textcolor{blue}{\textbf{}}}{\textcolor{red}{#1}}}}

\newcommand{\cmt}[1]{{\color{gray}{\texttt{//} #1}}}

\definecolor{color1}{HTML}{B65A3E}
\definecolor{color2}{HTML}{1F232A}






\begin{abstract}
Adversarial learning and the robustness of Graph Neural Networks (GNNs) are topics of widespread interest in the machine learning community, as documented by the number of adversarial attacks and defenses designed for these purposes. 
While a rigorous evaluation of these adversarial methods is necessary to understand the robustness of GNNs in real-world applications, we posit that many works in the literature do not share the same experimental settings, leading to ambiguous and potentially contradictory scientific conclusions. \\
In this benchmark, we demonstrate the importance of adopting fair, robust, and standardized evaluation protocols in adversarial GNN research. 
We perform a comprehensive re-evaluation of seven widely used attacks and eight recent defenses under both poisoning and evasion scenarios, across six popular graph datasets. Our study spans over 453,000 experiments conducted within a unified framework. \\
We observe substantial differences in adversarial attack performance when evaluated under a fair and robust procedure. Our findings reveal that previously overlooked factors, such as target node selection and the training process of the attacked model, have a profound impact on attack effectiveness, to the extent of completely distorting performance insights. These results underscore the urgent need for standardized evaluations in adversarial graph machine learning.
\end{abstract}

 
\section{Introduction}
\label{intro}

Applying machine learning to graph-structured data, such as financial transaction networks and social graphs, requires models that can effectively embed non-Euclidean relationships. Graph Neural Networks (GNNs), introduced by \citet{franco2009the} and \citet{alessio2009neural}, have become foundational tools for this purpose. 
Over the past decade, GNNs have achieved strong performance across domains, but their vulnerability to adversarial attacks has raised growing concerns. A series of recent works~\citep{daniel2018nettack,jintang2023sga,kaidi2019pgd,simon2021robustness} demonstrate that even minor perturbations to the input graph can significantly degrade GNN performance.

As attack strategies have proliferated, inconsistencies in evaluation protocols have emerged as a serious obstacle to scientific progress. Many studies report substantial gains using differing experimental setups, making results difficult to compare and conclusions potentially misleading. \\
The reproducibility crisis in machine learning has highlighted the importance of standardized, rigorous empirical evaluations~\citep{lipton_troubling_2018}. In adversarial graph learning, \citet{felix2022robust} warns that the graph community has yet to learn the \textquote{bitter lesson} from the vision community, where overlooking adaptive attacks and evaluation rigor once led to a flood of unreliable defense mechanisms evaluations.
 
In this work, we identify several recurring issues in current evaluations. First, GNNs are often trained using attack-specific hyperparameters or fixed data splits, biasing results. Second, new attack models are frequently tested under more favorable conditions than their baselines. Third, evaluations commonly select target nodes in a way that underrepresents high-degree nodes, which are typically more resistant to attacks (see Figure~\ref{fig:node_cat_poison}). As a result, reported improvements may reflect favorable setups rather than true advances in method design. \\
{To further examine these issues, we re-evaluate several widely used gray-box attacks under a standardized, robust evaluation framework, demonstrating how variations in evaluation protocols can produce inconsistent or overstated findings. While a comprehensive re-evaluation of all attacks is infeasible, our focused effort aims to establish stronger evaluation practices for the community.}

To better contextualize performance claims, we also introduce \textit{a naive yet surprisingly good baseline}, \propseRND, which achieves competitive results at minimal computational cost. Its success reinforces the need for basic sanity checks when proposing complex new methods. \\
By demonstrating the limited scalability of many existing attacks and their declining effectiveness on high-degree nodes, our work highlights overlooked challenges in adversarial graph learning. We hope to encourage the development of more robust and scalable attack and defense strategies.

\textbf{Disclaimer.} This work advocates for rigorous evaluation practices. It is not intended to rank attacks or discredit prior contributions but to enable more reliable and reproducible comparisons across future studies.
\section{Related Work}\label{relate_works}

\textbf{Adversarial Attacks.} 
Recent studies on adversarial attacks on graph data have developed optimal strategies to minimally perturb the graph under a budget constraint while achieving the highest impact on a GNN's classification performance. Among the first methods is Nettack~\citep{daniel2018nettack}, a gradient-based adversarial attack strategy that generates perturbations on graph structure and node features. Upon the success of Nettack, novel adversarial attack strategies have been  proposed~\citep{jinyin2018fga,simon2021robustness}. Most early adversarial attacks focus on small-scale datasets with fewer than $5000$ nodes.
By only extracting a much smaller subgraph centered at the target nodes, \citet{jintang2020sga} proposed SGA as a scalable adversarial strategy. PR-BCD, another adversarial attack at scale by~\citet{simon2021robustness}, adopts the Randomized Block Coordinate Descent~\citep{yurii2012efficiency} for solving large-scale optimization problems to find optimal perturbations.  Meanwhile, in a recent study, GOttack~\citep{zulfikar2025gottack} uses graph structures by targeting topological equivalence groups and exploiting their influence in gradient-based adversarial models.

\textbf{Evaluation procedures.} We follow the good practice of~\cite{federico2020faircomparision} and~\cite{shchur2018pitfall}. Particularly, both works standardize the evaluation procedures and promote a reproducible experimental environment with a rigorous \textit{model selection} and \textit{assessment framework}, but in two different contexts. \citet{federico2020faircomparision} focuses on graph classification tasks while ~\citet{shchur2018pitfall}'s work is primarily on node classification. In addition, \citet{shchur2018pitfall} have shown that the train/validation/test split of choice used in evaluation significantly impacts the performance ranking, thus drawing community attention to the necessity of using different splits in the evaluation procedure. Differentiating from them, which focus on designing rigorous evaluation frameworks for GNN models, we propose a robust evaluation procedure to prevent over-optimistic and biased estimates of the true performance of adversarial attack strategies.

The Graph Robustness Benchmark (GRB)~\citep{zheng2021graph} was introduced years ago, and it mainly focuses on global evasion attacks. However, the GRB does not consider three valuable scenarios: (i) targeted attacks, (ii) poisoning scenarios, and (iii) the distinction between homophilic and heterophilic graphs. Our benchmark addresses these limitations by incorporating both targeted evasion and poisoning attacks, while explicitly evaluating performance on homophilic and heterophilic graphs, with victim models trained in each scenario.
\section{Preliminaries}\label{sec:preliminaries}

Let $\mathcal{G} = (\mathcal{V}, \mathcal{E}, \mathbf{X})$ denote a graph, where $\mathcal{V}$ is the set of $N$ nodes, $\mathcal{E} \subseteq \{(v, w) \mid v, w \in \mathcal{V} \}$ is the set of directed edges, and $\mathbf{X} = \{ \mathbf{x}_0, \mathbf{x}_1, \dots, \mathbf{x}_{N-1} \}$ is the set of node feature vectors. Each $\mathbf{x}_i \in \mathbb{R}^M$ encodes the $M$-dimensional attributes of node $v_i$. The graph structure is represented by an adjacency matrix $\mathbf{A} \in \{0,1\}^{N \times N}$, where $\mathbf{A}_{ij} = 1$ if $(v_i, v_j) \in \mathcal{E}$, and $0$ otherwise. Each node $v_i$ has an associated label vector $\mathbf{y}_i \in \{0,1\}^{|\mathcal{C}|}$ indicating its membership in one of $|\mathcal{C}|$ classes, forming the label matrix $\mathbf{Y} \in \{0,1\}^{N \times |\mathcal{C}|}$.

\textbf{Semi-supervised Node Classification.} We focus on node classification in a semi-supervised setting, where labels are available only for a subset of nodes. Let $\mathcal{V}_L \subset \mathcal{V}$ denote the set of labeled nodes with known labels $\mathbf{Y}^L$, and $\mathcal{V}_U = \mathcal{V} \setminus \mathcal{V}_L$ the set of unlabeled nodes. The goal is to learn a function $g: \mathcal{G}, \mathbf{Y}^L \to \mathbf{Y}^U$ that predicts a class probability distribution for each node in $\mathcal{V}_U$. The predicted label $\hat{y}_v$ for a node $v \in \mathcal{V}_U$ corresponds to the class with the highest predicted probability in $g(v)$.

\textbf{Node Classification Margin.} For a node $v$ with ground truth label $y$, the classification margin $M_v$ measures the confidence of the model in the correct class. It is defined as the difference between the model’s output score for the true class and the highest score assigned to any incorrect class~\citep{daniel2018nettack}:
\begin{equation}
M_v = g(v)_y - \max_{c \in \mathcal{C},\, c \neq y} g(v)_c
\end{equation}
A small or negative margin indicates that the prediction is uncertain or incorrect, making such nodes more susceptible to adversarial perturbation. 

\textbf{Risk Assessment.} Risk assessment refers to the empirical evaluation of model performance across multiple random splits~\citep{federico2020faircomparision}. Given $K$ random splits of $\mathcal{V}$ into disjoint subsets $\mathcal{V}_{\text{train}}$, $\mathcal{V}_{\text{valid}}$, and $\mathcal{V}_{\text{test}}$, the model is trained on $\mathcal{V}_{\text{train}}$ and tuned on $\mathcal{V}_{\text{valid}}$. For each split $k$, the best hyper-parameter configuration is selected based solely on validation performance. The empirical risk is then estimated by averaging the test performance across the $K$ splits.

\textbf{Model Selection.} Model selection aims to identify the hyper-parameter configuration that yields the highest validation accuracy. However, validation accuracy is often a biased estimator of generalization performance~\citep{federico2020faircomparision,gavin2010onOverfitting}. Overreliance on validation performance can lead to overfitting and inflated expectations. In adversarial GNN literature, model selection and final evaluation are often conflated, undermining fair comparisons across attack strategies. Proper separation between model selection and risk assessment is essential to avoid misleading conclusions.

\section{Graph Adversarial Attacks}
\label{sec:adversarial-attacks}

Adversarial attacks on graphs aim to perturb either the structure or features of a graph $\mathcal{G} = (\mathbf{A}, \mathbf{X})$ to degrade the performance of a GNN. We refer to the targeted model as the \textbf{victim} model. The attack modifies $\mathcal{G}$ into a perturbed version $\mathcal{G}' = (\mathbf{A}', \mathbf{X}')$, leading the victim to misclassify selected nodes.

\textbf{Attacker’s Capacity.}
The adversarial attack can introduce perturbations to data either in the inference or training phases. In the \textbf{evasion} setting, the victim model trains on clean graph data $\mathcal{G}$ to perform inference on the perturbed data $\mathcal{G}'$. In the \textbf{poisoning} setting, adversarial attacks create a modified graph $\mathcal{G}'$, which is then used to train a model.

\textbf{Perturbation Type.} We perturb $\mathcal{G}$ within a given budget $\Delta$ by adding or removing edges from $\mathcal{E}$ or perturbing node features. Formally, we can write
\begin{equation}
\sum_{u} \sum_{v} \left| \mathbf{X}_{uv} - \mathbf{X}'_{uv} \right| + \sum_{u<v} \left| \mathbf{A}_{uv} - \mathbf{A}'_{uv} \right|   \leq \Delta
\end{equation}
 

\textbf{Attacker’s Knowledge.} Attacks differ in the information available to the adversary. In \textbf{black-box} settings, the attacker lacks access to model parameters and labels. \textbf{White-box} attacks assume full access to both, a strong but often unrealistic assumption. In \textbf{gray-box} settings, the attacker can access the training data and labels, allowing them to train a \textbf{surrogate} model that approximates the victim. We adopt the gray-box setting, as it balances realism with the ability to diagnose vulnerabilities. Unlike prior work that uses fixed surrogates, we also evaluate \textbf{adaptive} attacks where perturbations are directly optimized against defended victim models, simulating stronger adversaries. Note that our evaluation pipeline is modular and extensible to all attack types.

\textbf{Attacker’s Target.} We focus on \textbf{targeted} attacks, where a chosen subset of nodes $\mathcal{V}_T \subseteq \mathcal{V}_{test}$ are perturbed to induce misclassification, as they are often harder to detect in real systems.

\textbf{Victim Models.} We define two classes of victim models: \textbf{vanilla} GNNs~\citep{bacciu_gentle_2020}, which are not trained with adversarial robustness in mind, and \textbf{defended} GNNs, which incorporate explicit defense mechanisms. Attacks against vanilla models define the baseline vulnerability, while defended scenarios test the effectiveness of robustness interventions. We emphasize that defense approaches generally operate without any prior knowledge of specific attacks.

\subsection{Attack Models and Pitfalls of Evaluation}\label{subsec:evaluation_issue}

Many attack evaluations in the literature suffer from inconsistent experimental setups, limiting fair comparison. Details of evaluation pitfalls for specific attacks are discussed in Appendix~\ref{sec:adversarial_pitfall}; here we formalize criteria for rigorous assessment.

\textbf{Target Node Selection.} Most prior works follow the Nettack~\citep{daniel2018nettack} strategy, selecting (i) the 10 nodes with the highest margin of classification, indicating evident correctness; (ii) the 10 nodes with the lowest margin (still correctly classified); (iii) 20 additional nodes randomly chosen.  This strategy may underrepresent high-degree nodes, which are harder to attack due to their richer neighborhood context (Figure~\ref{fig:node_cat_poison}). This bias inflates attack performance and skews conclusions.

\textbf{Evaluation Criteria.} A high-quality evaluation should satisfy the following: (i) the victim model has undergone a model selection process, as it usually happens in real-world scenarios; (ii) results are averaged over $K$ random splits with standard deviations and public splits; (iii) target nodes include diverse structural types; and (iv) evaluations include both vanilla and defended victims. Our benchmark adheres to all of these conditions.

\textbf{Attack Models.} We benchmark seven widely cited attack methods, selected based on peer-review status, architectural diversity, and citation count. These are Nettack~\citep{daniel2018nettack}, FGA~\citep{jinyin2018fga}, SGA~\citep{jintang2020sga}, GOttack~\citep{zulfikar2025gottack}, PR-BCD~\citep{simon2021robustness}, and PGD~\citep{kaidi2019pgd}. Full summaries and surrogate configurations appear in Appendix Section~\ref{sec:summary_attacks}.

\textbf{Victim Models.} Vanilla victim models are three standard GNNs: GCN~\citep{thomas2017gcn}, GSAGE~\citep{william2017gsage}, and GIN~\citep{keyula2019gin}, each using a single aggregation function, and a fourth vanilla model, PNA~\citep{gabriele2020pna}, which combines multiple aggregation operations. We also evaluate {nine} defended victim models, selected according to the taxonomy in Appendix Table~\ref{tab:categorization}, with selection criteria detailed in Appendix~\ref{gnn}.

\textbf{Adaptive Attacks.} Adaptive attacks are designed with full awareness of the defense, producing stronger and more targeted perturbations. We evaluate PR-BCD in both its fixed-surrogate (\prbcdNA) and adaptive variants. Though non-adaptive PR-BCD may underestimate its true capability, we include it due to its scalability, popularity, and baseline strength (later results in Tables~\ref{appendixTable:prbcdDefendedHomophily} and \ref{appendixTable:prbcdDefendedHeterophily} will show marginal differences between the variants). {The scope of this work is to benchmark adversarial attacks and defenses in a practical setting, where neither attackers nor defenders have access to the opponent’s backbone model or strategy. Consequently, evaluating defenses against fully adaptive attacks specifically crafted to circumvent their core mechanisms falls outside the focus of this study.}

\textbf{Naïve Baseline.} We introduce \propseRND, a simple yet effective baseline attack. Instead of using gradients or learned surrogates, \propseRND perturbs the graph by modifying edges connected to nodes selected using their degree and features. Despite its simplicity and low computational cost, it achieves surprisingly strong results, underscoring the importance of including naïve baselines to contextualize claimed improvements. Algorithm~\ref{alg:l1d-rnd} provides implementation details and in-depth analysis on factors driving the effectiveness of \propseRND is discussed in Appendix~\ref{sec:rndv2}.

\subsection{Risk Assessment in Adversarial Evaluation} \label{sec:method}
\begin{figure*}[]
    \centering
    \includegraphics[width=1\linewidth]{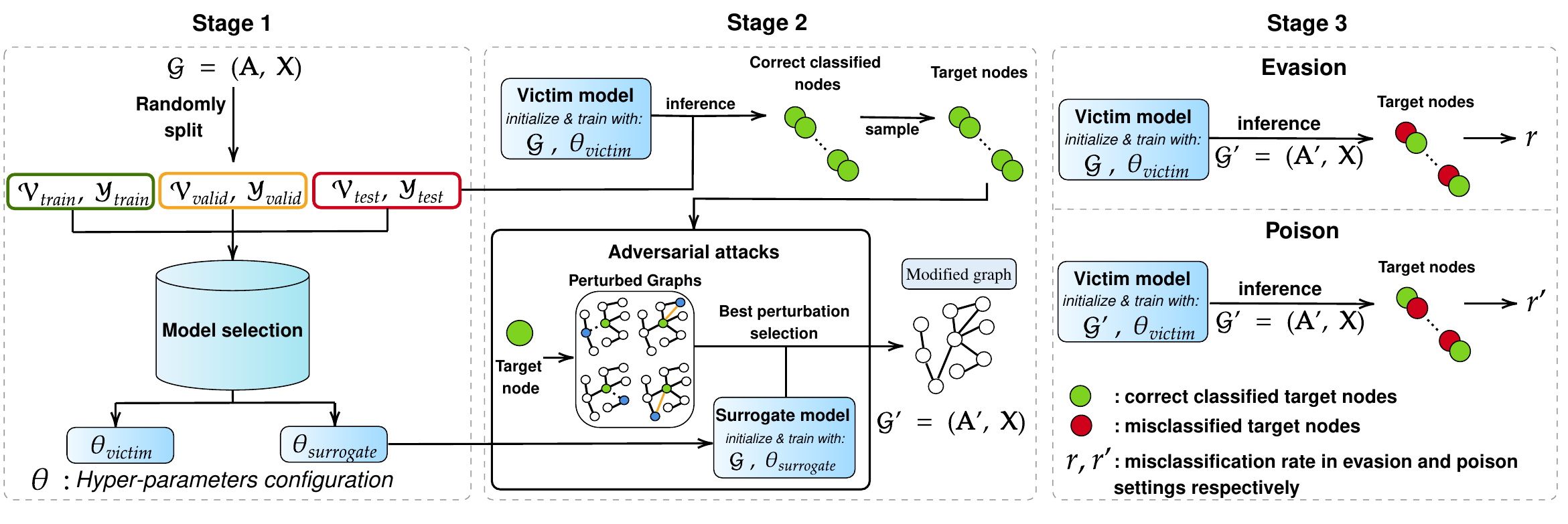}
    \vspace{-1em}
    \caption{Overview of our risk assessment framework for adversarial GNN evaluation.}
    \label{fig:overal_pipeline}
    \vspace{-2.3em}
\end{figure*}

Unifying the good practices of~\citet{federico2020faircomparision} and \citet{shchur2018pitfall}, the pseudo-algorithm of our proposed adversarial attacks evaluation is provided in Algorithm~\ref{alg:evaluation}. 

We first obtain $K$ different random splits from datasets, 
(Line~\ref{line:split}). The victim model's hyperparameters are first tuned on the $i$-th split's training set, and the best victim model \textit{for that split} is chosen based on the performance on the validation set (Line~\ref{line:victimtrain}). 

The model selection process relies solely on the training and validation sets to ensure an unbiased risk estimation. Notably, model selection of all models is performed on clean data (data without perturbation). Given the predictions of the best victim model, we sample a subset of target test nodes $\mathcal{V}_T$ that have been correctly classified (Line~\ref{line:testselect}). An adversarial example on a given target node is considered a successful attack if it causes the victim model to flip its prediction about the target node. 

The adversarial attack performance is evaluated based on the misclassification rate on a specific budget $\Delta$, averaged over $K$ different splits and $R$ risk assessment runs for each split. 
 Figure~\ref{fig:overal_pipeline} visualizes the overall proposed evaluation pipeline. As the hyper-parameter configurations of victim models are carefully selected through the model selection process, we also perform model selection on surrogate models used in adversarial attacks to ensure that the process is realistic. The model selection process on victim models and surrogate models is kept on the same hyper-parameter grids.

\begin{wrapfigure}{R}{0.5\textwidth}
\vspace{-2em}
\begin{minipage}{0.5\textwidth}
\begin{algorithm}[H]
   \caption{\small Adversarial attack/defense evaluation}
   \label{alg:evaluation}
   \scriptsize
\begin{algorithmic}[1]
   \STATE \textbf{Input:} Dataset $\mathcal{D}$, Configs $\mathbf{\Theta}$,
    Attack method \textbf{Attack}, Budget $\Delta$,\\
   \hspace{1.6em} Splits $K$, Runs $R$
   \STATE \textbf{Output:} Avg. misclassification rates in both evasion and poison settings
   \STATE Create $K$ train/val/test splits $F_1,..,F_K$ from $\mathcal{D}$  \label{line:split}
   \FOR{$i=1,...,K$}
        \STATE $\mathcal{V}^i_{train}$, $\mathcal{V}^i_{valid}$, $\mathcal{V}^i_{test}$ = $F_i$
        \STATE $\mathbf{\theta}^i_{best}$ = select($\mathbf{\Theta}$, $\mathcal{V}^i_{train}$, $\mathcal{V}^i_{valid}$, $\mathcal{D}$)
        \cmt{Alg. \ref{alg:select}} \label{line:victimtrain}
        \FOR{$r = 1,... ,R$} 
            \STATE $f$ = train($\mathbf{\theta}_{best}^i$, $\mathcal{V}^i_{train}$, $\mathcal{V}^i_{valid}$, $\mathcal{D}$)
            \STATE $\mathcal{V}_T$ = node\_select($f$, $\mathcal{V}^i_{test}$, $\mathcal{D}$) \label{line:testselect}
            \FOR{$v$ in $\mathcal{V}_T$}
                \STATE $\mathcal{D'}$ = \textbf{Attack}($v$, $\mathcal{D}$, $\Delta$) 
                \STATE ${s^{i}_{v,r}}$ = 1 if $f(v) = y_v$, else 0 otherwise \cmt{evasion}
                \STATE $f'$ = train($\mathbf{\theta}_{best}^i$, $\mathcal{V}^i_{train}$, $\mathcal{V}^i_{valid}$, $\mathcal{D'}$) \cmt{retrain}
                \STATE ${s^{',i}_{v,r}}$ = 1 if $f'(v) = y_v$, else 0  \cmt{poison}
                \STATE reset \textbf{Attack}
            \ENDFOR
       \ENDFOR
    \ENDFOR
    \STATE success rate = $\frac{\sum_{i=1}^{K}\sum_{r}^{R}\sum_{t}^{\mathcal{T}}s_{t,r}^i}{K \times R \times {|\mathcal{T}|}}$ \cmt{evasion}
    \STATE success rate$'$ = $\frac{\sum_{i=1}^{K}\sum_{r}^{R}\sum_{t}^{\mathcal{T}}{s_{t,r}^{',i}}}{K \times R \times {|\mathcal{T}|}}$ \cmt{poison}
    \STATE \textbf{Return:} success rate, success rate$'$
\end{algorithmic}
\end{algorithm}
\end{minipage}
\vspace{-2em}
\end{wrapfigure}

\vspace{-1em}
\section{Experiments}\label{Experiments}
 
We conduct extensive experiments to re-evaluate adversarial attacks under the standardized framework described in Section~\ref{sec:method}.

\textbf{Experimental Setup.} We adopt a transductive, semi-supervised node classification setting. 
For the evaluation procedure defined in Section~\ref{sec:method}, we set $K=5$ and $R= 3$. Following common practice~\citep{daniel2018nettack,zulfikar2025gottack}, the $\mathcal{V}_{train}/\mathcal{V}_{valid}/\mathcal{V}_{test}$ ratio is set to 10/10/80.
 
We implement early stopping with the patience parameter $n$, where training stops if $n$ epochs have passed without improvement on the validation set. Importantly, the same data split $F_i$ (Line~\ref{line:split}) is shared across different models to ensure a fair comparison.
 
We perform model selection for all victim and surrogate models, in both vanilla and defended scenarios, based on their performance on the validation sets. For each split, we evaluate adversarial attacks equipped with surrogate models. We report the average misclassification rate, percentage of nodes misclassified by the model, of adversarial attacks on vanilla and defended models across initialization seeds and splits. We fixed the same model as a surrogate model for each adversarial method (Table~\ref{tab:atta_surro}) for all evaluation settings, regardless of the choice of victim models, as attackers may not always know the classifier's architecture prior to performing the attack in practice. 
 
\textbf{Hyper-parameters.} Model selection varies hyperparameters, including the number of layers, embedding dimensions, learning rate, etc., based on ranges provided in original publications. Additional model-specific parameters are included as needed. Full details are in Appendix~\ref{sec:Hyper-parameters}.

\textbf{Target Node Selection.} For each experiment, we evaluate on 50 target nodes selected to ensure diversity in classification margin and structural role: i) $10$ correctly classified nodes with the highest degree, ii) $10$ correctly classified nodes with the lowest degree, iii) $10$ correctly classified nodes with the highest margin, iv) $10$ nodes with the lowest margin (but still correctly classified)  and v) $10$ randomly chosen nodes.

\textbf{Datasets.} We evaluate on six datasets: three homophilous graphs (\cora, \citeseer, \pubmed~\citep{yang2016revisiting}), two heterophilous graphs (\squirrel, \chameleon~\citep{benedek2021multi}), and one large-scale benchmark from OGB~\citep{weihua2020open}. Detailed description of these datasets can be found Appendix~\ref{sec:dataset}. Many adversarial and defense methods do not scale well to large graphs, and we highlight such limitations where relevant (see Appendix~\ref{sec:costs}).

\begin{wraptable}{r}{0.55\linewidth}
\vspace{-20px}
\setlength{\tabcolsep}{2.5pt}
\caption{Descriptive statistics of datasets.}
\label{tab:dataset_stats}
\centering
\scriptsize
\begin{tabular}{l l r r r r}
\toprule  
 \textbf{Type} & \textbf{Dataset} & \textbf{Nodes} & \textbf{Edges}  & \textbf{Features} & \textbf{Labels} \\
\midrule
\multirow{3}{*}{{\textbf{Homophilic}}}
& \cora & $2,708$ & $5,069$ & $1,432$  & $7$ \\
& \citeseer & $3,327$ & $3,668$ & $3,703$  & $6$ \\
& \pubmed  & $19,717$ & $44,325$ & $500$  & $3$ \\
\midrule
\multirow{2}{*}{\textbf{Heterophilic}}
& \chameleon & $2,277$   & $36,101$   &  $3,132$  & $5$ \\
& \squirrel  & $5,201$  & $217,073$   &  $3,148$  & $5$ \\
\midrule
{\textbf{Large scale}}
& \ogb & $169,343$   & $1,166,243$  & 128  & $40$ \\
\bottomrule 
\vspace{-2em}
\end{tabular}
\end{wraptable}

 \textbf{Computational Environment.} Experiments were run using Python 3.8.19 and PyTorch 2.3.0 on a Linux cluster with Intel Xeon Gold 6338 CPUs, 251 GB RAM, and NVIDIA RTX A40 GPUs with 44 GB memory. GNNs were implemented using PyTorch Geometric 2.5.3, and we reused code from DeepRobust~\citep{li2020deeprobust}, GreatX~\citep{greatX}, and author-provided repositories for additional defense methods.

\textbf{Reproducibility.} We release all code at \url{https://github.com/FDataLab/GAB}, including dataset splits and hyperparameters, for reproducible benchmarking with minimal overhead.

\section{Results and Discussion} \label{sec:results}

This section provides an in-depth discussion of our results. We discuss vanilla models' attacks in Section~\ref{sec:vanilla_attacks} and defense models' attacks in  Section~\ref{sec:eval_attack_defended}. Notably, higher misclassification rates reflect more effective attack models. The comprehensive results are summarized in Figures~\ref{fig:homophily_radar_plots} and~\ref{fig:heterophily_radar_plots}.  Time and GPU cost results are detailed in the Appendix~\ref{sec:costs} and \ref{sec:costs_defense}.

 \noindent\textbf{Computational considerations.} Our experiments include up to 453,461 training runs (see Appendix~\ref{appendix:numberBreakdown} for a breakdown). In some cases, model selection or attacks on a single split exceeded 120 hours, making full experiments infeasible. We capped training time at 120 hours; results exceeding this are marked as OOR (Out of Resource).

\subsection{Vanilla Evasion and Poisoning Attacks} \label{sec:vanilla_attacks}
\begin{table*}[t]
    \centering
    \caption{\textbf{Homophily Results.} Evaluating adversarial attacks with budget $\Delta = 1$ in both evasion and poison settings on GCN (vanilla attack) and GNNGuard (defended attack). NA indicates a non-adaptive variant.
    }
    \small
    \renewcommand\arraystretch{2.5}
    \setlength{\tabcolsep}{2.5pt}
    \resizebox{0.95\linewidth}{!}{
        \label{tab:budget1meta_homophily}
        \begin{tabular}{cc | ccccccc | ccccccc}
            \toprule
            && \multicolumn{7}{c}{Attack Model for Evasion} & \multicolumn{7}{c}{Attack Model for Poisoning} \\
            \cline{3-16}
            & Victim & \propseRND & FGA & NETTACK & PGD & \prbcdNA & SGA & GOttack  & \propseRND & FGA & NETTACK & PGD & \prbcdNA & SGA & GOttack\\
            \hline

            \multirow{2}{*}{\rotatebox{90}{\textbf{\cora}}} & GCN & 13.20 $\pm$ 0.04 &  27.87 $\pm$ 0.04 & \underline{29.60 $\pm$ 0.05} & 29.33 $\pm$ 0.04 & \textbf{32.13 $\pm$ 0.04} & 26.27 $\pm$ 0.04 & 28.53 $\pm$ 0.04 &
            15.47 $\pm$ 0.04 & 30.00 $\pm$ 0.06  & \textbf{33.47 $\pm$ 0.04} & 31.73 $\pm$ 0.04 & 32.80 $\pm$ 0.06 & 29.33 $\pm$ 0.04 & \underline{33.33 $\pm$ 0.07}\\
            &GNNGuard & 6.27 ± 4.13 & 6.67 ± 3.44  & 6.80 ± 4.77 & 6.80 ± 2.60 & 8.13 ± 3.34 & \textbf{8.40 ± 4.29 }  & \underline{8.40 ± 4.97 }
            & 6.93 ± 4.40 & 7.47 ± 3.96 & 7.47 ± 5.37  & 6.93 ± 3.01 & 8.93 ± 3.99 & \underline{9.60 ± 3.64 } & \textbf{10.27 ± 6.54 }\\
            \cline{1-16}
            \multirow{2}{*}{\rotatebox{90}{\textbf{\citeseer}}} & GCN & 15.20 $\pm$ 0.04 & 25.47 $\pm$ 0.04 & \underline{28.13 $\pm$ 0.07} & 25.47 $\pm$ 0.05 & \textbf{34.53 $\pm$ 0.07} & 23.47 $\pm$ 0.03 & 25.60 $\pm$ 0.04 & 16.27 $\pm$ 0.04 & 31.87 $\pm$ 0.07 & \textbf{36.40 $\pm$ 0.07} & 30.80 $\pm$ 0.07 & \underline{34.27 $\pm$ 0.06} & 25.20 $\pm$ 0.04 & 34.27 $\pm$ 0.08 \\
            &GNNGuard & \textbf{4.67 ± 3.68}  & 3.33 ± 3.18 & \underline{4.67 ± 2.35} & 3.07 ± 2.49 & 3.07 ± 2.12   & 4.00 ± 2.73 & 4.67 ± 2.89 
            & \underline{4.80 ± 3.84} & 4.40 ± 3.31  & \textbf{6.00 ± 2.93 } & 3.20 ± 2.11 &3.20 ± 1.97  & 4.80 ± 3.00 & 4.80 ± 2.70  \\

            \cline{1-16}
            \multirow{2}{*}{\rotatebox{90}{\textbf{\pubmed}}} & GCN & 10.93 $\pm$ 0.03 & \underline{35.60 $\pm$ 0.03} & 33.73 $\pm$ 0.03 & 34.13 $\pm$ 0.04 & 29.60 $\pm$ 0.03 & 34.27 $\pm$ 0.04 & \textbf{35.87 $\pm$ 0.03} & 9.73 $\pm$ 0.03 & \underline{35.33 $\pm$ 0.03} & 34.13 $\pm$ 0.05 & 33.60 $\pm$ 0.03 & 29.60 $\pm$ 0.03 & 34.13 $\pm$ 0.04 & \textbf{35.60 $\pm$ 0.03}\\
            &GNNGuard & \textbf{6.53 ± 4.63} & 3.60 ± 1.88   & 2.93 ± 1.67 & 2.53 ± 1.60 & \underline{4.27 ± 2.25} & 3.47 ± 1.92 & 3.07 ± 2.25
            &\textbf{6.53 ± 4.93 } & 4.80 ± 3.19  & 4.40 ± 2.53 & 4.27 ± 2.12 & \underline{5.60 ± 4.08} & 4.93 ± 3.20 & 4.93 ± 3.10 7.87\\
            \bottomrule
        \end{tabular}
    }
    \vspace{-1.5em}
\end{table*}

\begin{table*}[t]
    \centering
    \caption{\textbf{Heterophily Results.} Evaluating adversarial attacks with budget $\Delta = 1$ in both evasion and poison settings on GCN (vanilla attack) and RUNG (defended attack). NA indicates a non-adaptive variant.
    }
    \small
    \renewcommand\arraystretch{3}
    \setlength{\tabcolsep}{2.5pt}
    \resizebox{0.95\linewidth}{!}{
        \label{tab:budget1meta_heterophily}
        \begin{tabular}{cc | ccccccc | ccccccc}
            \toprule
            && \multicolumn{7}{c}{Attack Model for Evasion} & \multicolumn{7}{c}{Attack Model for Poisoning} \\
            \cline{3-16}
            & Victim & \propseRND & FGA & NETTACK & PGD & \prbcdNA & SGA & GOttack  & \propseRND & FGA & NETTACK & PGD & \prbcdNA & SGA & GOttack\\
            \hline

            \multirow{2}{*}{\rotatebox{90}{\textbf{\squirrel}}} & GCN & 24.93 $\pm$ 33.09  &  \underline{62.40 $\pm$ 14.64} & 1.87 $\pm$ 3.34  & 47.73 $\pm$ 10.25 & \textbf{69.87 $\pm$ 10.76} & 52.00 $\pm$ 6.19 & 13.33 $\pm$ 4.64  &
            34.67 $\pm$ 27.36 & \underline{63.87 $\pm$ 12.25} & 2.80 $\pm$ 2.24  & 52.27 $\pm$ 8.21 & \textbf{70.27 $\pm$ 11.16} & 53.47 $\pm$ 5.97 & 13.60 $\pm$ 3.79\\

            & RUNG & 2.13 $\pm$ 1.77 & 1.87 $\pm$ 2.88 & 0.27 $\pm$ 1.03 & 2.00 $\pm$ 1.85 & 1.73 $\pm$ 2.49 & \textbf{2.67 $\pm$ 3.68} & 0.93 $\pm$ 2.25 & 11.33 $\pm$ 8.64 & \textbf{20.53 $\pm$ 9.69} & 6.53 $\pm$ 4.44 & 17.33 $\pm$ 7.81 & 15.07 $\pm$ 6.18 & 18.40 $\pm$ 9.33 & 6.27 $\pm$ 5.18 \\

            \cline{1-16}
            \multirow{2}{*}{\rotatebox{90}{\textbf{\chameleon}}} & GCN & 21.87 $\pm$ 28.89  & \textbf{62.40 $\pm$ 7.72} & 3.07 $\pm$ 2.60  & 44.00 $\pm$ 20.95 & \underline{58.00 $\pm$ 18.53} & 45.47 $\pm$ 10.38 & 23.47 $\pm$ 8.16  & 35.60 $\pm$ 22.31 & \textbf{66.40 $\pm$ 8.25} & 7.87 $\pm$ 4.98 & 53.47 $\pm$ 17.98 & \underline{64.80 $\pm$ 14.69} & 51.47 $\pm$ 9.12  & 27.07 $\pm$ 8.48 \\

            & RUNG & \textbf{2.27 $\pm$ 3.28} & 0.93 $\pm$ 1.83 & 0.27 $\pm$ 0.70 & 1.87 $\pm$ 4.69 & 0.80 $\pm$ 2.24 & 2.13 $\pm$ 4.31 & 0.67 $\pm$ 1.23 & 12.00 $\pm$ 8.88 & \textbf{16.40 $\pm$ 7.53} & 7.33 $\pm$ 2.89 & 13.73 $\pm$  6.41 & 11.73 $\pm$ 5.90 & 14.27 $\pm$ 6.76 & 10.80 $\pm$ 4.89 \\

            \bottomrule
        \end{tabular}
    }
    \vspace{-0.5em}
\end{table*}

We evaluate seven adversarial attack models under evasion and poisoning settings on vanilla GNNs across homophily and heterophily datasets. Appendix Tables~\ref{tab:eval_attack_non_defense_homo},~\ref{tab:non_defense_heterophily_p1}, and~\ref{tab:non_defense_heterophily_p2} report complete results. For conciseness, we also provide reduced summaries in Tables~\ref{tab:budget1meta_homophily} and~\ref{tab:budget1meta_heterophily}.

\noindent\textbf{Homophily datasets.} \prbcdNA has the highest misclassification rates (i.e., best attack model) in Table~\ref{tab:eval_attack_non_defense_homo} for a budget of $\Delta = 1$ in $4$ out of $12$ evasion scenarios (spanning three datasets and four victim models). The remaining cases are distributed among other methods, with \propseRND surprisingly yielding the highest misclassification rates in \pubmed and \citeseer when PNA is the victim model. When the budget is increased ($\Delta =2,\ldots, 5$), Nettack demonstrates superior performance in $36$ out of $48$ cases in Table~\ref{tab:eval_attack_non_defense_homo}. Unlike the low-budget case, \prbcdNA achieves the best results only in \citeseer and \cora for GCN. In scenarios where GCN is the victim model, \prbcdNA, which uses GCN as the surrogate model, delivers competitive performance, surpassing Nettack in $10$ out of $15$ settings; these are the $\Delta=1,\ldots,5$ budgets in \cora and \citeseer in Table~\ref{tab:eval_attack_non_defense_homo}. However, \prbcdNA shows limited adversarial effectiveness on victim models that differ significantly from the surrogate ones. For example, \prbcdNA no longer outperforms FGA on GIN, GSAGE and PNA in evasion attacks. In Table~\ref{tab:eval_attack_non_defense_homo}, the baseline \propseRND exhibits the lowest performance based on average rank on homophily datasets (i.e., $7$ out of $7$ attack models). Notably, unlike the attack models, the baseline does not achieve high misclassification rates with increasing budgets.  

As shown in the lower rows of Table~\ref{tab:eval_attack_non_defense_homo}, poisoning attacks are significantly more effective than evasion attacks. For $\Delta = 1$, the best-attack model, Nettack, achieves a $4.72\%$ relative increase, rising from an average of $27.76\%$ in evasion to $32.48\%$ in poisoning across three datasets and four victim models. Even the baseline, \propseRND, experiences a $8.9\%$ improvement in poisoning attacks. Nettack remains the best-performing model when ranks are averaged over all five budgets, with an overall rank of $1.65$ across three datasets and four models. FGA follows as the second-best model, with an average rank of $3.23$.

Among three homophily datasets, \pubmed has the lowest misclassification rates for $\Delta=5$, whereas attack models reach 70\% in \cora and \citeseer in Table~\ref{tab:eval_attack_non_defense_homo}. With \cora and \citeseer, even the \propseRND baseline makes considerable gains in misclassification with increasing budgets. 

In addition, attacks provide a critical lens to evaluate the robustness of victim models under adversarial conditions. As Table~\ref{tab:eval_attack_non_defense_homo} shows, \textbf{GraphSAGE is the most resilient victim model in both evasion and poisoning attacks}. In evasion, Nettack yields the lowest average misclassification rate of $44.8\%$ against GraphSAGE across $15$ budgets (three datasets and five budgets). GIN follows with an average misclassification rate of $48.5\%$.  In poisoning,  Nettack has an average of $47.97\%$ misclassification rate on GraphSAGE models in three datasets across all budgets; other victim models have misclassification rates in $[53.4\%,57.4\%]$.

\noindent\textbf{Heterophily datasets.} As shown in Table~\ref{tab:non_defense_heterophily_p1} and ~\ref{tab:non_defense_heterophily_p2}, under budget 1, the average misclassification rate across seven attacks on four non-defense models is $33.96\%$ for heterophily datasets and $27.78\%$ for homophily datasets, while the average misclassification rate is  $52.21\%$ for homophily datasets and $46.86\%$ for heterophily datasets under budget 5. This suggests that the first \textbf{perturbation has a greater adversarial impact in heterophily settings}. However, \textbf{with increasing perturbation budgets, attacks tend to yield larger gains on homophily datasets}.
Nettack demonstrates the highest effectiveness on homophily datasets with an average rank of $1.64$, but its performance significantly drops on heterophily datasets, where it ranks $5.48$, the second worst. Interestingly, another evaluation shows that FGA achieves an average rank of $3.10$ on homophily datasets (second-best), but rises to $1.92$ on heterophily datasets, making it the top-performing attack in that setting. Similarly, the naïve random attack \propseRND shows the opposite trend; it performs surprisingly well on heterophily datasets with an average rank of $2.33$, but performs the worst, with an average rank of $6.27$, on homophily datasets among the seven adversarial attack methods.

\noindent\textbf{Large-scale dataset.} On the moderately sized \ogb dataset, which contains fewer than 200K nodes and is still considered small by industry standards, only three attack methods (\propseRND, \prbcdNA, and SGA) and two victim models (GCN and GSAGE) could be fully evaluated within the 120-hour compute limit. As shown in Table~\ref{tab:ogb_results}, SGA consistently achieves the highest misclassification rates across budgets 1 to 5 in both evasion and poisoning settings. \textbf{This result underscores a critical limitation: most existing adversarial attacks are not scalable enough to be applied even to modestly large graphs, raising concerns about their practicality in real-world deployments.} {Notably, this bottleneck is not due to limited computational resources: our experiments were conducted on a high-performance cluster with multiple NVIDIA RTX A40 GPUs (44 GB memory, high memory bandwidth). To provide valuable lessons for designing future scalable attack algorithms, we provide a detailed analysis of each adversarial attack’s runtime by breaking down the algorithm in Section~\ref{sec:costs}}

{\noindent\textbf{Feature attack.} While focused on structural attacks, our experimental pipeline can be extended to feature-based attacks. We conduct and report the results on feature-attack variants in Appendix~\ref{sec:feature-attack}. Overall, feature attacks largely mirror the trends observed in structural attacks on homophily datasets, where NETTACK and FGA remain the best and the second-best methods.}

\begin{table*}[ht]
    \centering
    \caption{Misclassification rate ($\uparrow$) on \ogb with budget $\Delta=1$ to $5$ in both evasion and poison setting on GCN and GSAGE.
    }
    \small
    \renewcommand\arraystretch{1.5}
    \setlength{\tabcolsep}{2pt}
    \resizebox{0.95\linewidth}{!}{
        \label{tab:ogb_results}
        \begin{tabular}{cc | ccccc | ccccc}
            \toprule
            && \multicolumn{5}{c}{GCN} & \multicolumn{5}{c}{GSAGE} \\
            \cline{3-12}
            & Attack & 1 & 2 & 3 & 4 & 5 & 1 & 2 & 3 & 4 & 5\\
            \hline

            \multirow{3}{*}{\rotatebox{90}{\textbf{Evasion}}} & \propseRND & 16.00 $\pm$ 2.00& 30.67 $\pm$ 7.02& 36.00 $\pm$ 2.00& 38.00 $\pm$ 2.00& 36.67 $\pm$ 3.06&16.67 $\pm$ 8.08& 27.33 $\pm$ 7.02& \underline{33.33 $\pm$ 7.57}& \underline{29.33 $\pm$ 9.45}& \underline{34.67 $\pm$ 7.02} \\
            & PR-BCD (NA)& \underline{23.33 $\pm$ 3.06} & \underline{34.00 $\pm$ 3.46} & \underline{38.00 $\pm$ 2.00} & \underline{40.67 $\pm$ 3.06}& \underline{39.33 $\pm$ 1.15} & \underline{20.67 $\pm$ 1.15}& \underline{24.67 $\pm$ 1.15}& 22.00 $\pm$ 3.46& 24.67 $\pm$ 3.06& 28.00 $\pm$ 2.00\\
            & SGA& \textbf{36.67 $\pm$ 5.03}& \textbf{48.67 $\pm$ 7.02} & \textbf{56.00 $\pm$ 2.00} & \textbf{57.33 $\pm$ 1.15}& \textbf{58.67 $\pm$ 1.15} & \textbf{40.67 $\pm$ 2.31}& \textbf{56.00 $\pm$ 10.00}& \textbf{63.33 $\pm$ 4.16}& \textbf{72.00 $\pm$ 7.21} & \textbf{ 71.33 $\pm$ 5.77}\\
            \cline{1-12}
            
            \multirow{3}{*}{\rotatebox{90}{\textbf{Poison}}} & \propseRND& 17.33 $\pm$ 4.16& 32.00 $\pm$ 6.00& \underline{38.00 $\pm$ 2.00}& 39.33 $\pm$ 1.15& 38.67 $\pm$ 2.31 & 17.33 $\pm$ 5.03& \underline{26.00 $\pm$ 3.46}& \underline{30.00 $\pm$ 14.00}& \underline{31.33 $\pm$ 6.43}& \underline{34.67 $\pm$ 7.02}\\
            & PR-BCD (NA)& \underline{22.00 $\pm$ 5.29}& \underline{34.67 $\pm$ 3.06}& 36.67 $\pm$ 1.15& \underline{40.67 $\pm$ 5.03}& \underline{39.33 $\pm$ 1.15} & \underline{17.33 $\pm$ 3.06}& 24.00 $\pm$ 2.00& 19.33 $\pm$ 3.06& 26.00 $\pm$ 2.00& 26.00 $\pm$ 2.00\\
            & SGA& \textbf{36.67 $\pm$ 2.31} & \textbf{48.67 $\pm$ 8.08} & \textbf{56.00 $\pm$ 2.00} & \textbf{57.33 $\pm$ 1.15} & \textbf{58.67 $\pm$ 1.15} & \textbf{40.00 $\pm$ 3.46} & \textbf{60.00 $\pm$ 8.00} & \textbf{61.33 $\pm$ 5.77} & \textbf{70.67 $\pm$ 5.03} & \textbf{74.67 $\pm$ 8.08}\\
            
            \bottomrule
        \end{tabular}
    }
    \vspace{-1.5em}
\end{table*}


\begin{figure*}[ht!]
    \begin{center}
    
    \subfigure[High margin]{\includegraphics[width=0.20\textwidth]{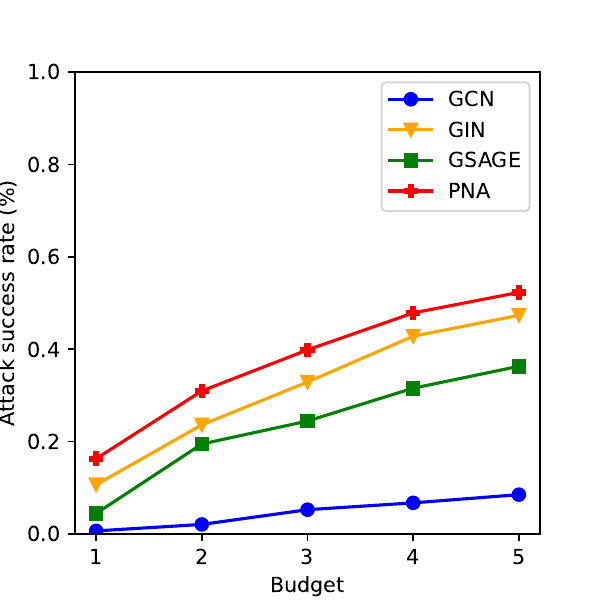}\label{fig:high_margin_poison_best}}\hspace{-7px}
    \subfigure[Low margin]{\includegraphics[width=0.20\textwidth]{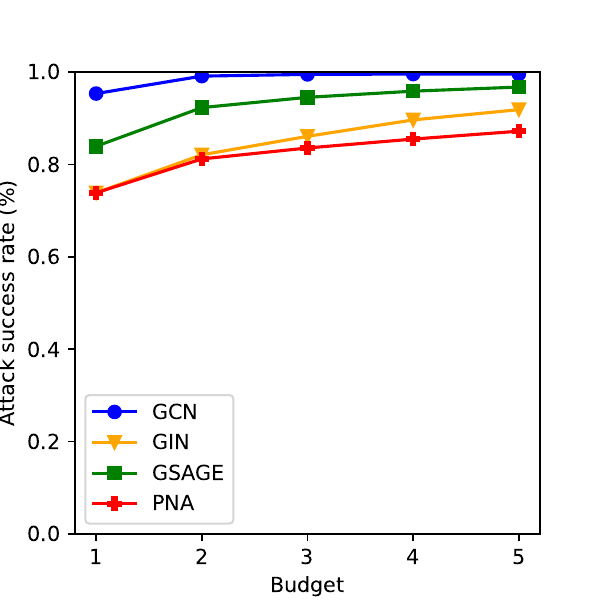}\label{fig:low_margin_poison_best}}\hspace{-7px}
    \subfigure[High degree]{\includegraphics[width=0.20\textwidth]{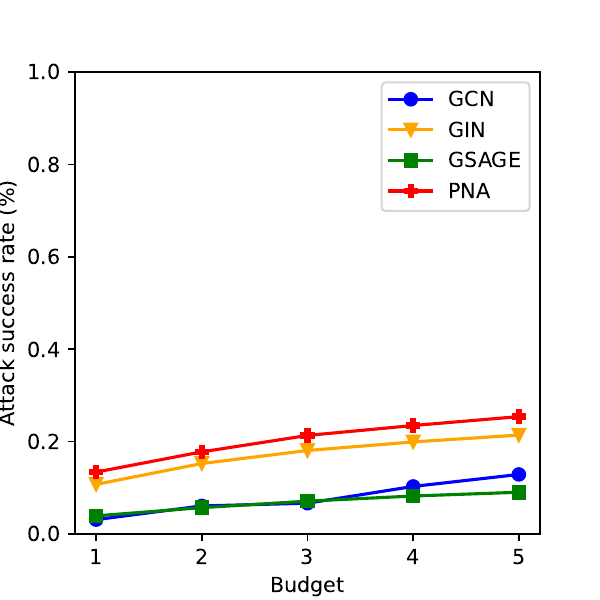}\label{fig:high_degree_poison_best}}\hspace{-7px}
    \subfigure[Low degree]{\includegraphics[width=0.20\textwidth]{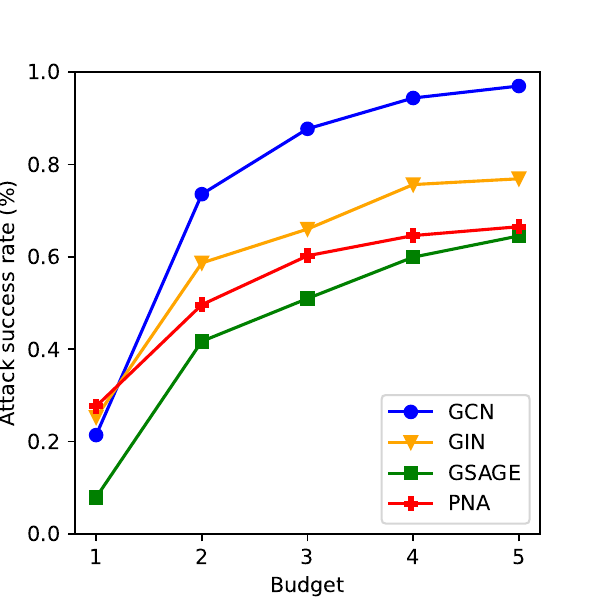}\label{fig:low_degree_poison_best}}\hspace{-7px}
    \subfigure[Random]{\includegraphics[width=0.20\textwidth]{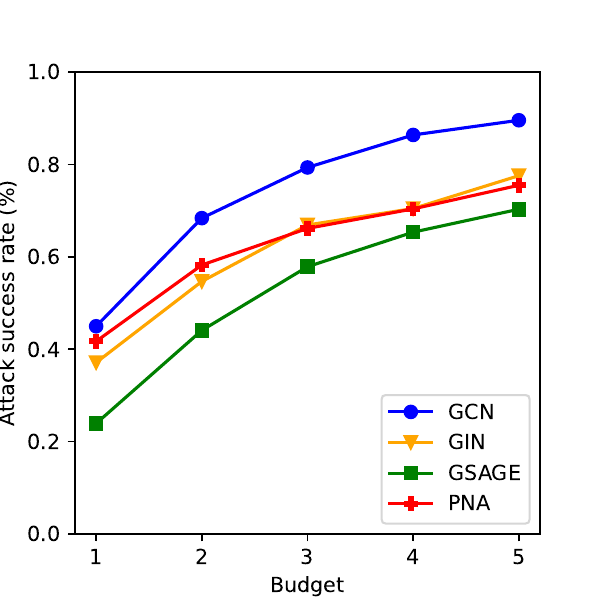}\label{fig:random_poison_best}}
    
    \caption{Average misclassification rate for different node categories of four non-defense models caused by seven adversarial attacks on three homophily datasets in the poison setting.}
    \label{fig:node_cat_poison}
    \end{center}
\vspace{-1.5em}
\end{figure*}


\subsection{Defended Evasion and Poisoning Attacks} \label{sec:eval_attack_defended}

 The results for evasion and poisoning attacks are shown in Appendix Tables~\ref{tab:eval_defense_homo_p1},~\ref{tab:eval_defense_homo_p2} and~\ref{tab:eval_defense_homo_p3} for homophily datasets and Tables~\ref{tab:eval_defense_heter_p1},~\ref{tab:eval_defense_heter_p2} and~\ref{tab:eval_defense_heter_p3} for heterophily datasets. We also show reduced results in Table~\ref{tab:budget1meta_homophily} and Table~\ref{tab:budget1meta_heterophily}, lower rows. 
 
\noindent\textbf{Homophily datasets.} Appendix Tables~\ref{tab:eval_defense_homo_p1},~\ref{tab:eval_defense_homo_p2} and~\ref{tab:eval_defense_homo_p3} demonstrate that defense models substantially reduce misclassification rates, with poisoning attacks being generally easier to defend against than evasion. For example, Nettack's average misclassification rate drops from $49.1\%$ to $32.2\%$ under evasion, and from $53.6\%$ to $23.56\%$ under poisoning when defenses are applied. Nettack still ranks highest in 8 out of 15 defended poisoning scenarios.

Among defense methods, GNNGuard is the most effective: at $\Delta = 1$, it achieves an average misclassification rate (i.e., best defense) of just $5.01\%$ for evasion and $5.92\%$ for poisoning across three datasets and seven attacks. GRAND consistently ranks as the second-best defense. In contrast, the FGA attack model, despite its strong performance in the vanilla setting, performs poorly against all defended models.

The performance of the \propseRND baseline is noteworthy: this simple, naive attack achieves the highest misclassification rate in 19 out of 45 defended evasion settings and 18 out of 45 defended poisoning settings (across three datasets, three victim models, and five budgets). \textbf{Its surprisingly strong performance, despite lacking any optimization or model-specific tuning, calls into question the actual gains offered by several state-of-the-art adversarial attack methods.} We provide factors driving the effectiveness of \propseRND in Appendix~\ref{sec:rndv2}.

To highlight key trends, we focus on the best-performing defense, GNNGuard, and its performance against vanilla attacks on the widely used GCN model, as summarized in Table~\ref{tab:budget1meta_homophily}.

At lower budgets (e.g., $\Delta = 1$), which represent realistic perturbation scenarios such as the addition or removal of a single edge, Nettack does not outperform any other model in evasion attacks. When defenses are applied, Nettack’s evasion effectiveness often falls below that of \prbcdNA. However, in poisoning attacks, Nettack remains strong, with GOttack emerging as the second-best method.
 
Overall, \textbf{Table~\ref{tab:budget1meta_homophily} highlights the robustness of GNNGuard, which substantially reduces the effectiveness of advanced attacks in both evasion and poisoning settings, often lowering their impact to the level of the naive \propseRND baseline.}

\noindent\textbf{Heterophily datasets.} As shown in Tables~\ref{tab:eval_defense_heter_p1},~\ref{tab:eval_defense_heter_p2} and~\ref{tab:eval_defense_heter_p3}, {defenses are more effective on heterophily datasets than on homophily datasets}. Across budgets $\Delta = 1$ to $5$, average misclassification rates on homophily datasets are $[20.11\%, 40.82\%]$, while the corresponding rates on heterophily datasets are substantially lower: $[16.08\%, 23.33\%]$. This discrepancy suggests that \textbf{heterophily graphs are more resistant to adversarial perturbations}, likely due to weaker local homogeneity, which reduces the impact of structural changes. Under defense, attack methods achieve abysmally low rates on heterophily datasets compared to homophily datasets. Notably, FGA becomes the top-performing attack on heterophily datasets (rank $1.92$), while Nettack's performance degrades sharply, dropping from rank $2.08$ to $6.10$, the worst among all methods. The best defence model, RUNG~\citep{HouFDL24}, reduces most attack rates to ~2\%. This underscores that \textbf{novel approaches are needed in heterophily settings}, marking this as an open and underexplored research area. \textbf{Additionally, despite its strong robustness on homophily datasets, GNNGuard becomes notably fragile under heterophily. On the Chameleon dataset, it prunes all existing edges, resulting in uniformly zero misclassification rates (0.00) across all attacks (Table~\ref{tab:eval_defense_heter_p1}).}

 \noindent\textbf{The Impact of Target Node Selection.}  
Target node selection plays a critical role in evaluating adversarial attacks. In the literature, Nettack’s strategy, selecting 10 high-margin nodes, 10 low-margin nodes, and 20 randomly chosen ones (totaling 40), has become a de facto standard. However, this approach ignores structural properties such as node degree, which substantially influence attack success. Appendix Table~\ref{tab:missclass-table-degree-impact} shows that using node degree as a selection criterion results in substantially lower attack success compared to margin-based or random selection, even in vanilla (undefended) settings. Figure~\ref{fig:node_cat_poison} shows that all attack methods perform poorly on these nodes, while low-degree and low-margin nodes remain vulnerable. High-degree nodes exhibit alarmingly low misclassification rates, ranging from only $0.05$ to $0.28$, despite the absence of any defense. Crucially, {because most attacks were developed and benchmarked on small graphs with low average degree (e.g., \cora), this vulnerability remained undetected, largely due to scalability limitations that prevent testing on larger, high-degree networks}. This suggests that \textbf{evaluations ignoring node degree systematically overstate both the effectiveness of attack models and the fragility of GNNs and raises serious concerns about the applicability of current attack models to real-world networks}, where average node degrees are much higher~\citep{networkdegree}. {Our results reveal that low-degree nodes tend to be substantially more vulnerable, whereas high-degree nodes exhibit stronger inherent robustness under adversarial attacks. This suggests that future defense mechanisms may benefit from prioritizing protection for low-degree nodes rather than applying a uniform strategy across the entire graph.}

\noindent\textbf{The Impact of Victim Model Selection.}  
In practical deployments, victim models are selected based on performance over training and validation sets. However, many adversarial GNN studies evaluate attacks on fixed, non-optimized model configurations, ignoring this critical step. Our results in Table~\ref{tab:missclass-table-model-selection-impact} show that incorporating model selection into evaluations can significantly alter attack outcomes. On \cora and \citeseer, victim models chosen via model selection are generally more vulnerable: for example, SGA's misclassification rate on GIN in the poisoning setting differs by an average of $15.27\%$ across budgets. In contrast, on \pubmed, model selection sometimes leads to more robust victim models. This is particularly evident for GraphSAGE, where attacks are less effective on tuned models than on fixed ones.

These findings highlight a critical inconsistency: the perceived effectiveness of \textbf{adversarial attacks depends not only on the attack method but also on whether the victim model is realistically selected}. Evaluations that omit this step may either overstate or understate the vulnerability of GNNs, leading to misleading conclusions about attack strength.

\vspace{-1em}
\section{Conclusion}
\label{Conclusion}
\vspace{-1em}
We have conducted a large-scale evaluation of adversarial attacks and defenses on GNNs, revealing that conclusions from prior work often do not hold under fair and rigorous settings. While Nettack has remained a strong performer, the unexpectedly competitive results of our naive baseline, \propseRND, challenge assumptions about the progress made in adversarial graph learning. PR-BCD and FGA are scalable options. Our analysis has shown that dataset properties, target node selection, and victim model configuration significantly affect attack success, yet have been inconsistently addressed in past evaluations. Our findings highlight the need for standardized, practical benchmarks that reflect real-world constraints and model selection practices. By exposing gaps in current evaluation protocols, we have laid the groundwork for more reliable assessments of adversarial robustness in graph learning. We hope this work prevents the repetition of past methodological pitfalls and encourages more transparent and scalable evaluations moving forward.

\bibliography{references}
\bibliographystyle{abbrvnat} 

\newpage
\appendix

\newpage
\newpage
\appendix
\onecolumn
\centerline{\bf \Large Appendix}
\section{LLM Usage}
We acknowledge the use of LLMs to aid or polish writing. We confirm that research motivation, methodology, idea and content comes from us, the authors, while the LLM’s role was limited to refining clarity, grammar, style, and LaTeX formatting.

\section{Reproducibility Statement}
We release a complete anonymized codebase to ensure full reproducibility at \url{https://github.com/FDataLab/GAB}. All experiments are run with fixed random seeds, and model hyperparameters are obtained from performing model selection on set of all possible hyperparameters provided in Table~\ref{tab:Hyper-parameters-table}. Additional details on compute resources and experimental setup are described in Section~\ref{Experiments}.

\section{Broader Impact \& Limitation}\label{sec:limitation}

This work addresses critical challenges in the evaluation of adversarial attacks and defenses for Graph Neural Networks. By introducing a unified and rigorous benchmarking framework, our study ensures reproducibility, fairness, and robustness in assessing adversarial learning methods. This initiative contributes to a more reliable understanding of GNN vulnerabilities and their mitigations, promoting trustworthy adoption of GNNs in high-stakes domains such as finance, healthcare, and social networks. Furthermore, our discovery of specific node properties, such as the higher resilience of high-degree nodes, opens pathways for new research into graph robustness. While our focus is primarily on methodological evaluation, the broader societal impact includes fostering ethical AI practices through improved transparency and standardization in machine learning research.

{While our benchmark explores a comprehensive range of hyper-parameters, we acknowledge that model weight initialization can also influence robustness outcomes. \citeauthor{sofiane2024if} highlights that different initializations may lead to noticeably different robustness levels. Due to the computational cost, over 446,000 additional experiments, our current benchmark utilized the initialization used by the original papers and does not perform hyper-parameter tuning on initialization. We therefore report the specific initialization scheme used for each model, leaving a full exploration of initialization robustness for future work.}

{Our benchmark focuses on static graphs, which is the predominant setting in the adversarial GNN literature. While attacks on dynamic graphs and continuous-time embeddings are also important, they remain largely unexplored across existing benchmarks. We view these as natural and valuable extensions of our work rather than omissions, and anticipate that our framework can provide the foundation for evaluating such scenarios in the future.}

{
Our benchmark does not include an evaluation of certified robustness. While certification offers valuable theoretical guarantees, most defense methods in our study do not provide certification modules or compatible implementations, making it infeasible to run a fair and standardized certification comparison. We therefore consider certification an important but orthogonal direction, and we leave a systematic benchmark of certified defenses to future work.
}

\section{Hyper-parameters} \label{sec:Hyper-parameters}

Table~\ref{tab:Hyper-parameters-table} presents the hyperparameters for various models. Model selections include the number of convolutional layers, the learning rate, and the early stopping criterion (based on either validation accuracy or validation loss) for all models. In addition, model-specific parameters such as regularization terms, dropout rates, and other configurations were chosen as appropriate for each model.

Moreover, we outline the hyperparameters for the model's specific parameters, as follows.

\textbf{RobustGCN}: The hyperparameters for the RGCN model are configured as follows: $\gamma = 0.5$, $\beta_1 = 1$, and $\beta_2 = 1.5$ across all datasets. The dropout rate $\in$ $\{0.5, 0.6\}$, while the Adam optimizer is used for optimization with a learning rate $\in$ $\{0.001, 0.01, 0.005\}$. The model is trained for a maximum of $1000$ epochs, with early stopping applied based on the performance on the validation set.

\textbf{GARNET}: We set the hyperparameters as follows: the number of eigenpairs used for spectral embedding, $r \in \{50, 500\}$; the number of nearest neighbors used for base graph construction, $k \in \{50, 30, 10\}$; and the edge pruning threshold, $\gamma$, set to $0.0003$. Moreover, we explore GARNET configurations where node features are either included or excluded to understand the role of node attributes. Similarly, for the construction of the nearest-neighbor graph, we assess both weighted and unweighted approaches. We evaluated the effect of normalizing versus not normalizing the adjacency matrix on stabilization and scaling in graph computations.

\textbf{GRAND}: We set the hyperparameters for the GRAND model as follows: the node drop rate is set to $0.5$, the temperature parameter, $\text{T} \in \{0.1, 0.5, 0.3\}$, the regularization coefficient, $\lambda \in \{1.0, 0.5, 0.7\}$, and the order of graph convolutions, $\text{order} \in \{2, 4, 8\}$. Moreover, we explore GRAND configurations by adjusting the number of samples, $\text{sample} \in \{2, 3, 4\}$, to examine the impact of sample size on learning.

Given a model, all possible choices of all hyperparameters form a set of hyperparameter configurations. We perform model selection using Algorithm~\ref{alg:select} to select the configuration having the best performance on the validation dataset, $\mathcal{V}_{valid}$.

\begin{algorithm}[H]\small
   \caption{Model Selection (select)}
   \label{alg:select}
\begin{algorithmic}[1]
   \STATE {\bfseries Input:} Set of configurations $\mathbf{\Theta}$, Dataset $\mathcal{D}$,\\
   Train indexes $\mathcal{V}_{train}$, Validation indexes $\mathcal{V}_{valid}$
   \STATE {\bfseries Output:} $\theta_{best}$: the best hyper-parameter configuration
   \STATE p$_{\theta}$ = $\emptyset$ 
   \FOR{$\mathbf{\theta}$ in $\mathbf{\Theta}$}
   \STATE $f$ = train($\theta$, $\mathcal{V}_{train}$,$\mathcal{V}_{test}$,  $\mathcal{D}$)
   \STATE performance = validate($\theta$, $\mathcal{V}_{valid}$, $\mathcal{D}$)
   \STATE p$_{\theta}$ = p$_{\theta}$ $\cup$ performance
   \ENDFOR

   \STATE $\theta_{best}$ = $\mathop{\arg\max}\limits_{\theta}$ p$_{\theta}$
   \STATE {\bfseries Return:} $\theta_{best}$

\end{algorithmic}
\end{algorithm}

{\textbf{Initializer.} Initialization plays a non-trivial role in determining the stability and robustness of GNN models. Consistent with findings in recent literature, different initializations may lead to different robustness levels, even under identical training settings. While our benchmark uses commonly adopted default initializers for each architecture, we recognize that further investigation into initialization sensitivity may reveal additional insights into model and defense behavior.
}

\section{Datasets}\label{sec:dataset}

We conducted experiments on six widely used node classification datasets, with their statistics provided in Table~\ref{tab:dataset_stats}. Specifically, we use three homophilic graphs: \cora~\citep{yang2016revisiting}, \citeseer~\citep{yang2016revisiting}, and \pubmed~\citep{yang2016revisiting} datasets are citation networks characterized by undirected edges and binary features, where nodes represent publications and edges correspond to citation links. We also include two heterophilic datasets \chameleon~\citep{Rozemberczki2021} and \squirrel~\citep{Rozemberczki2021} which are Wikipedia-based page-to-page networks focused on specific topics. In these datasets, nodes represent web pages, edges indicate mutual hyperlinks between pages, and node features are derived from several informative nouns found within the corresponding Wikipedia content. Moreover, we also consider one large-scale graph dataset \ogb~\citep{weihua2020open}, which is a directed graph representing the citation network among all Computer Science papers on arXiv. In this dataset, each node corresponds to an arXiv paper, and each directed edge indicates that one paper cites another.

\section{Adversarial Attacks Evaluation Pitfalls} \label{sec:adversarial_pitfall}

\textbf{Nettack.} The adversarial impact on victim models was reported with an average of over five random initiations/splits. However, the results do not include standard deviation, and different train/validation/test splits are not publicly available, which makes it difficult to re-produce the evaluation experiments. The evaluation of Nettack is limited to a single victim model, GCN, with fixed hyperparameters; hence, it is unclear if model selection has been performed. In addition, the authors evaluate Nettack using a target node selection strategy that does not encompass all classes of nodes, particularly those with high degrees. This limitation results in an incomplete assessment of Nettack's impact on high-degree target nodes. 

\textbf{FGA.} The authors evaluate FGA on multiple victim models, such as GCN~\citep{thomas2017gcn}, GraRep~\citep{shapshen2015grarep}, GraphCAN~\citep{hongwei2018GraphGAN}, trained and evaluated on at least a random train/validation/test split. However, whether the averaged results are reported from different random splits and whether model selection is performed for each victim model is unclear. The adversarial impact of FGA is assessed on target nodes selected randomly based on their labels. Importantly, the evaluation code has not been released, making it difficult to reproduce the results.

\textbf{SGA.} SGA is evaluated on multiple victim models such as GCN~\citep{thomas2017gcn}, SGC~\citep{felix2019sgc}, GraphSAGE~\citep{william2017gsage}, etc,. However, similarly to FGA and Nettack, it is unclear whether the assessment and model selection are conducted over different random splits of training/validation/test sets. Moreover, the adversarial effect of SGA is evaluated on randomly selected target nodes, which take neither high-margin and low-margin nodes as used in Nettack nor nodes with high degrees.

\textbf{PR-BCD.} The article reports results obtained from the average over three random/splits and follows a target node selection strategy similar to Nettack according to the code. The evaluation process omitted model selection for each split; hence, PR-BCD is evaluated on different scalable victim models with fixed hyper-parameters for all splits. Importantly, the authors did not compare PR-BCD performance with popular baselines such as Nettack and FGA, even on small datasets, as the authors focus on robustness and adversarial performance on large graphs.

\textbf{PGD.} Similar to Nettack, PGD's performance is evaluated and reported with the mean and standard deviation of five different random splits on a single GCN classifier. However, data splits are not publicly released, making it impossible to reproduce the result, and it is not clear if GCN hyper-parameters are fixed in advance for every split or obtained from model selection for each split.
\label{adversarial_attack}

\textbf{GOttack.} GOttack is evaluated on three backbone GNNs (GCN, GIN, and GraphSAGE) and four defense models (RGCN, GCN-Jaccard, GCN-SVD, and MedianGCN), using fixed hyperparameters. It follows Nettack’s target node selection strategy, which does not account for all classes of nodes, especially high-degree nodes. Additionally, the experiments were not conducted over different random splits of training/validation/test sets. This limits the evaluation of GOttack’s effectiveness.

\subsection{Addressing Common Questions on Pitfalls}

\noindent\textbf{Have attacks such as Nettack, FGA, PR-BCD, and GOttack not already been studied extensively?}  

While these attacks have been evaluated in earlier work, prior studies were typically ad-hoc and limited: using fixed hyperparameters, single data splits, or narrow subsets of target nodes. Our benchmark corrects these shortcomings by conducting over 446,000 experiments across homophilic, heterophilic, and large-scale datasets, while incorporating proper model selection and multiple random splits. This provides the first systematic, large-scale, and reproducible comparison of adversarial GNN attacks.  

\noindent\textbf{Were structural node properties, such as margin or degree, not already analyzed before?}  

Some papers have presented anecdotal evidence on node properties, such as degree or margin, often restricted to small datasets like \cora~\citep{zheng2021graph,zhu2022does}. These analyses were not comprehensive or reproducible, as they relied on binning or one-off case studies. In contrast, our benchmark systematically shows that node degree strongly influences attack success rates across multiple datasets. Appendix Table~\ref{tab:missclass-table-degree-impact} further confirms that degree-based target selection drastically reduces attack effectiveness, a pattern overlooked in prior work due to scalability and methodological limitations.

\noindent\textbf{Have victim model configurations and model selection not already been considered?}  

Most previous evaluations fixed hyperparameters and skipped model selection, a limitation also noted in earlier benchmarks~\citep{zheng2021graph}. Our results demonstrate that incorporating realistic victim model selection can shift attack outcomes by more than 15\% depending on the dataset and budget. This shows that omitting model selection produces misleading conclusions about attack strength. Our benchmark closes this gap by embedding model selection systematically into the evaluation.

\section{Detailed Results on Adversarial Attack Evaluation} \label{sec:detail_result_attack}

\textbf{Experiment breakdown.} \label{appendix:numberBreakdown} Each instance involving training a victim model is counted as one experiment. Specifically, a total of 354,501 experiments are conducted for model selection across 14 models, including surrogate, vanilla, and defense models, on homophily, heterophily, and large-scale datasets. In contrast, 98,960 experiments are required to evaluate seven attack methods on all vanilla and defense models across all datasets. The breakdown of experiments for model selection and attack evaluation is presented in Table~\ref{tab:fine_tune_breakup} and Table~\ref{tab:eval_attack_breakup}, respectively. In total, this study involves 453,461 experiments.

\begin{table}[ht]
    \centering
    \caption{Break up of the numbers of experiments required for performing model selection for surrogate models, vanilla and defense models on homophily, heterophily and large-scale datasets. There are three homophily datasets, each of which has five training/validation/test splits. Each heterophily dataset, \chameleon and \squirrel, has also five training/validation/test splits. Due to high time complexity, we only consider one training/validation/testing split for large-scale dataset, \ogb, provided by~\citet{weihua2020ogb}}
    \small
    \renewcommand\arraystretch{1.2}
    \setlength{\tabcolsep}{3pt}
    \label{tab:fine_tune_breakup}

    \resizebox{\linewidth}{!}{ 
        \begin{tabular}{ c| c | c | c | c | c | c }
\toprule \toprule
\textbf{Model} & \textbf{\# combinations} & \textbf{homophily datasets ($3 \times 5$)} & \textbf{\chameleon ($\times 5$)} & \textbf{\squirrel ($\times 5$)} & \textbf{\ogb ($\times 1$)} & \textbf{Total} \\
\midrule
GCN          & 2187  & 32805  & 10935 & 10935 & 2187 &       \\
GIN          & 2187  & 32805  & 10935 & NA    & NA   &       \\
GSAGE        & 2187  & 32805  & 10935 & 10935 & 2187 &       \\
PNA          & 729   & 10935  & NA    & NA    & NA   &       \\
GCN-surrogate & 243   & 3645   & 1215  & 1215  & 243  &       \\
SGC          & 243   & 3645   & 1215  & 1215  & 243  &       \\
GNNGuard     & 48    & 720    & 240   & 240   & NA   &       \\
GRAND        & 4374  & 65610  & 21870 & 21870 & NA   &       \\
ElasticGNN   & 1296  & 19440  & 6480  & 6480  & NA   &       \\
RobustGCN    & 324   & 4860   & 1620  & 1620  & NA   &       \\
GCORN        & 27    & 405    & 135   & 135   & NA   &       \\
RUNG         & 216   & 3240   & 1080  & 1080  & NA   &       \\
GCN-GARNET       & 48    & 720    & 240   & 240   & NA   &       \\
GCN-Jaccard      & 6     & 90     & NA    & NA     & NA   &       \\
NoisyGNN      & 36     & 540     & 180    & 180     & NA   &       \\

\midrule
\textbf{Total} & \textbf{14,151} & \textbf{212,265} & \textbf{67,080} & \textbf{56,145} & \textbf{4,860} & \textbf{354,501} \\
\bottomrule \bottomrule

  \end{tabular}
    }
\end{table}

\begin{table}[ht]
    \centering
    \caption{Break up of the number of experiments required for evaluating attacks on vanilla and defense models on homophily, heterophily and large-scale datasets }
    \small
    \renewcommand\arraystretch{1.2}
    \setlength{\tabcolsep}{3pt}
    \label{tab:eval_attack_breakup}

    \resizebox{0.8\linewidth}{!}{ 
       \begin{tabular}{c|c|c|c|c|c|c}
        \toprule
        \toprule

        \textbf{Table index} & \textbf{\# rows } & \textbf{\# cols } & \textbf{\# cells} & \textbf{Expected experiments} & \textbf{OOR} & \textbf{\# Actual experiments} \\
        \hline
        Table~\ref{tab:ogb_results}  & 6   & 10  & 60  & 180   & 0   & 180   \\
        Table~\ref{tab:eval_attack_non_defense_homo} & 56  & 15  & 840 & 12,600 & 0   & 12,600 \\
        Table~\ref{tab:non_defense_heterophily_p1} & 28  & 5   & 140 & 2,100  & 0   & 2,100  \\
        Table~\ref{tab:non_defense_heterophily_p2} & 42  & 5   & 210 & 3,150  & 0   & 3,150  \\
        Table~\ref{tab:eval_defense_homo_p1} & 56  & 15  & 840 & 12,600 & 0   & 12,600 \\
        Table~\ref{tab:eval_defense_homo_p2} & 56  & 15  & 840 & 12,600 & 70  & 12,530 \\
        Table~\ref{tab:eval_defense_homo_p3} & 14  & 15  & 210 & 3,150 & 0  & 3,150 \\
        Table~\ref{tab:eval_defense_heter_p1} & 28  & 10  & 280 & 4,200  & 0   & 4,200  \\
        Table~\ref{tab:eval_defense_heter_p2} & 56  & 10  & 560 & 8,400  & 0   & 8,400  \\
        Table~\ref{tab:eval_defense_heter_p3} & 14  & 10  & 140 & 2,100  & 0   & 2,100  \\
        Table~\ref{tab:adaptive_homo} & 4   & 15  & 60  & 900   & 0   & 900   \\
        Table~\ref{tab:adaptive_heter} & 4   & 10  & 40  & 600   & 0   & 600   \\
        Table~\ref{tab:missclass-table-degree-impact} & 56  & 15  & 840 & 12,600 & 0   & 12,600 \\
        Table~\ref{tab:missclass-table-model-selection-impact} & 56  & 15  & 840 & 12,600 & 0   & 12,600 \\
        Table~\ref{tab:rnd_vs_rndv2} & 14  & 5   & 70  & 1,050  & 0   & 1,050  \\
        Table~\ref{tab:feat-attack} & 40  & 15   & 600  & 9,000  & 0   & 2,250  \\
        Table~\ref{tab:rnd_ablation} & 16  & 5   & 80  & 1,200  & 0   & 1,200  \\
        \hline
        \multicolumn{6}{r}{\textbf{Total:}} & \textbf{98,960} \\
        \bottomrule \bottomrule

    \end{tabular}
    }
\end{table}

\subsection{Vanilla Evasion and Poisoning Attacks}
\textbf{Homophily datasets.}
In this section, we present the experimental results. Table \ref{tab:eval_attack_non_defense_homo} illustrates the success rates of six adversarial attacks in four vanilla models (GCN, GIN, GSAGE and PNA) in three homophily datasets under evasion and poisoning attack settings.

For the \cora dataset in the evasion attack setting, Nettack achieves the best performance in $12$ of the $20$ tasks, including $4$ out of $5$ tasks for each of the GIN, GSAGE and PNA models. However, in the GCN model, Nettack delivers the second-best results. In contrast, \prbcdNA performs best in $5$ of $5$ tasks on the GCN model, although it only secures $7$ of $20$ tasks overall in the evasion setting. Interestingly, the proposed baseline \propseRND, achieves the highest score on the PNA model for budget $\Delta=1$. Likewise, in the poisoning attack setting for the same dataset, Nettack demonstrates the best performance, securing $14$ out of $20$ tasks. Meanwhile, \prbcdNA, GOttack, SGA, FGA and PGD achieve the best scores on at least one task out of $20$. For a budget $\Delta = 1$, the best-performing model, Nettack, achieves a $22.13\%$ relative increase on the \cora dataset, rising from an average attack success rate of $26.65\%$ in evasion to $32.56\%$ in poisoning across four victim models.

On the \citeseer dataset in the evasion attack setting, Nettack achieves the best performance in $13$ out of $20$ tasks. Similarly to the \cora dataset, Nettack delivers the second-best results on the GCN model and the proposed baseline, \propseRND, achieves the highest score in the PNA model for a budget $\Delta=1$. In contrast, \prbcdNA achieves the best performance in all tasks in the GCN model, although it secures only $5$ of the total $20$ tasks in the evasion setting. Likewise, in the poisoning attack setting, Nettack demonstrates the best performance, securing $14$ out of $20$ tasks. Moreover, \prbcdNA and FGA each score the best on $3$ out of $20$ tasks. However, Nettack achieves a $33.48\%$ relative increase on the \citeseer dataset, rising from $26.06\% $ in evasion to $34.79\%$ in poisoning across four victim models for a budget $\Delta = 1$.

Likewise, on the \pubmed dataset, Nettack achieves the best performance in $11$ out of $20$ tasks in both the evasion and poisoning attack settings. In contrast, GOttack delivers the best performance in $5$ out of $20$ tasks in the evasion setting and $3$ out of $20$ tasks in the poisoning setting. Notably, the proposed baseline, \propseRND, secures the highest score on the PNA model for a budget of $\Delta=1$ in both settings. Additionally, FGA achieves the best scores on $2$ out of $20$ tasks in the evasion setting and $3$ out of $20$ tasks in the poisoning setting. Furthermore, Nettack exhibits a $3.01\%$ relative decrease on the \pubmed dataset, dropping from $31.03\%$ in evasion to $30.09\%$ in poisoning across four victim models for a single budget. On average, the misclassification rate of all attacks on GCN on the \pubmed dataset in both evasion and poison settings at budget $\Delta=1$ drops by $22.59\%$ when high-degree nodes are selected as target nodes. Still, on average, a node degree-based selection of target nodes decreases the misclassification rate of adversarial attacks by $9.8\%$.

\textbf{Heterophily datasets.} This section presents the experimental results. Table~\ref{tab:non_defense_heterophily_p1} and ~\ref{tab:non_defense_heterophily_p2} show the success rates of six adversarial attacks on three vanilla models (GCN, GIN, and GSAGE) across two heterophilic datasets under both evasion and poisoning attack settings.

For the \squirrel dataset under the evasion attack setting, \prbcdNA achieves the highest performance in $5$ out of $10$ tasks, including $4$ out of $5$ tasks for the GCN model. FGA ranks second, performing best in $4$ out of $10$ tasks, with $4$ out of $5$ for the GSAGE model, and also securing the second-best results in $4$ out of $5$ tasks for the GCN model. Under the poisoning attack setting for the same dataset, both \prbcdNA and FGA achieve top performance, each leading in $4$ out of $10$ tasks. In contrast, Nettack, GOttack, SGA, and PGD demonstrate the lowest success rates in both evasion and poisoning scenarios. For a budget $\Delta = 1$, the best-performing model, \prbcdNA , achieves a $4.76\%$ relative increase on the \squirrel dataset, rising from an average attack success rate of $53.07\%$ in evasion to $55.6\%$ in poisoning across three victim models.

Similarly, on the \chameleon dataset, the proposed naive \propseRND achieves the best performance in $7$ out of $15$ tasks under the evasion attack setting. In comparison, FGA and \prbcdNA both demonstrate the second-best performance, each excelling in $4$ out of $15$ tasks. However, under the poisoning attack setting, FGA outperforms all other methods, achieving the best results in $10$ out of $15$ tasks, while \prbcdNA and the proposed \propseRND achieve top results in $3$ and $2$ tasks, respectively. Consistent with observations on other heterophily datasets, Nettack, GOttack, SGA, and PGD show the lowest performance across both evasion and poisoning settings. However, \prbcdNA demonstrates a $37.94\%$ relative improvement on the \chameleon dataset, increasing from $35.37\%$ in the evasion setting to $48.8\%$ in the poisoning setting across three victim models at a budget of $\Delta = 1$.

\textbf{Large-scale datasets.} Table~\ref{tab:ogb_results} presents the success rates of three adversarial attacks on two vanilla models (GCN and GSAGE) using a large-scale dataset under both evasion and poisoning attack settings. On the \ogb dataset, SGA achieves the best performance, outperforming others in all $10$ out of $10$ tasks across both settings. In comparison, \prbcdNA secures the second-best performance in $7$ out of $10$ tasks under the evasion setting. Similarly, under the poisoning setting, both \prbcdNA and the proposed \propseRND model achieve the second-best results in $5$ out of $10$ tasks.

\subsection{Defense Evasion and Poisoning Attacks}
\textbf{Homophily datasets.}
In Tables \ref{tab:eval_defense_homo_p1} and \ref{tab:eval_defense_homo_p2}, we present the results of six adversarial attacks on 8 defense models, across three homophily datasets in both evasion and poisoning attack settings.   

On ElasticGNN, the average performance of adversarial attacks is $23.35\%$ under a budget of 1 and $43.88\%$ under a budget of 5. Among all evaluated attacks, FGA exhibits the highest adversarial impact on ElasticGNN, achieving an average rank of 2.47 across both evasion and poisoning settings, followed closely by Nettack with an average rank of 2.57.

Conversely, while FGA performs best on ElasticGNN, it demonstrates surprisingly poor performance on GCN-GARNET, with an average rank of 5.27, ranking as the second-worst among the seven evaluated attacks. In contrast, \propseRND attains the second-best average rank of 2.47. This result indicates that the relative effectiveness of adversarial attacks varies significantly depending on the target defense model, underscoring the necessity of evaluating attacks across diverse defense strategies. On GCN-GARNET, the average attack performance is $18.89\%$ for budget 1 and $45.15\%$ for budget 5.

For GCN-Jaccard, the average attack performance is $24.76\%$ under budget 1 and $50.89\%$ under budget 5. Similar to ElasticGNN, GCN-Jaccard effectively defense against the \propseRND baseline attack, which consistently ranks lowest (average rank of 7.00 in both evasion and poison scenarios). In contrast, Nettack and GOttack emerge as the most effective attacks, with average ranks of 1.96 and 2.40, respectively.

GNNGuard demonstrates exceptional robustness against all adversarial attacks, with average misclassification rate of $5.46\%$, $9.31\%$, $12.22\%$, $15.19\%$, and $17.61\%$ from budget 1 through 5, respectively. Notably, attack effectiveness varies significantly across datasets. In particular, SGA performs best in 6 out of 10 tasks on the \cora dataset, while Nettack dominates on \citeseer in 9 out of 10 tasks), and \prbcdNA leads on \pubmed in 8 out of 10 tasks. These observations further emphasize the importance of evaluating adversarial methods across varied datasets and defense models.

On GRAND, the average attack performance is $20.33\%$ and $40.92\%$ for budgets 1 and 5, respectively. Our simple baseline attack \propseRND achieves superior performance across all homophily datasets on GRAND, with an average rank of 1.13 and best results in 26 out of 30 tasks, outperforming even gradient-based attack methods and demonstrating significantly higher computational efficiency. This raises concerns regarding the effectiveness on defense models, especially GRAND, of existing attack methods against simple baselines.
RobustGCN exhibits the weakest robustness among all defenses, with average adversarial performances of $28.76\%$ for budget 1 and $55.57\%$ on budget 5. Nettack again proves most effective on this model, with an average rank of 2.16 and top performance in 11 out of 30 tasks.

Among recent defense models, GCORN shows average adversarial performance of $21.10\%$ for budget 1 and $34.88\%$ for budget 5, while RUNG achieves $18.75\%$ and $33.21\%$, respectively, ranking as the second most robust model. On both GCORN and RUNG, Nettack remains the top-performing attack (average rank ~1.90), followed surprisingly by \propseRND.

In summary, Nettack consistently ranks as the most effective adversarial attack across 8 defense models on homophily datasets, with an overall average rank of 2.09. GOttack follows as the second-best method with an average rank of 3.83. Notably, our \propseRND performs on bar or even surpasses gradient-based methods on GCN-GARNET, GCORN, RUNG, and especially GRAND.

\textbf{Heterophily datasets.} Tables~\ref{tab:eval_defense_heter_p1} and~\ref{tab:eval_defense_heter_p2} present the success rates of seven adversarial attacks against six defense models in two heterophilic datasets, in both the evasion and poisoning attack settings. In the setting of evasion attack, the FGA demonstrates the best performance, achieving success in $12$ and $11$ of the $30$ tasks on the \squirrel and \chameleon datasets, respectively. Similarly, in the context of the poisoning attack, FGA continues to perform well, securing $17$ and $14$ of the tasks $30$ in the datasets \squirrel and \chameleon, respectively. However, the best-performing model, FGA, shows a significant performance gain in the \squirrel dataset, with a $62.61\%$ relative increase in the success rate that increases from $18.17\%$ under evasion to $29.55\%$ under poisoning attacks in all defense models. Likewise, for a budget of $\Delta = 1$, FGA achieves a relative improvement $48.16\%$ on the data set \chameleon, increasing its average attack success rate from $17.75\%$ in the evasion setting to $26.04\%$ in the poisoning setting. Interestingly, the proposed naive approach \propseRND achieves the second-best results with $9$ out of $30$ successful attacks on both datasets in the evasion setting. In the poisoning setting, the proposed model \propseRND secures $5$ and $11$ of the $30$ tasks in the \squirrel and \chameleon datasets, respectively. On the other hand, SGA achieves moderate performance, scoring $8$ and $6$ in the evasion setting on the \squirrel dataset, and $7$ and $1$ on the \chameleon dataset in the evasion and poisoning settings, respectively. Notably, similar to their performance on vanilla models, Nettack, GOttack, and PGD consistently exhibit lower success rates across both heterophilic datasets. {It is important to note that, in the \chameleon dataset under the GNNGuard defense, all attack models, including the best performing model, FGA,  produced surprising results. The misclassification rate remained at zero, even as the budget increased, with no observable changes. This occurred because GNNGuard pruned all the edges in this dataset.}

\begin{figure*}[ht!]
    \centering
    \subfigure[\cora]{\includegraphics[width=0.45\textwidth]{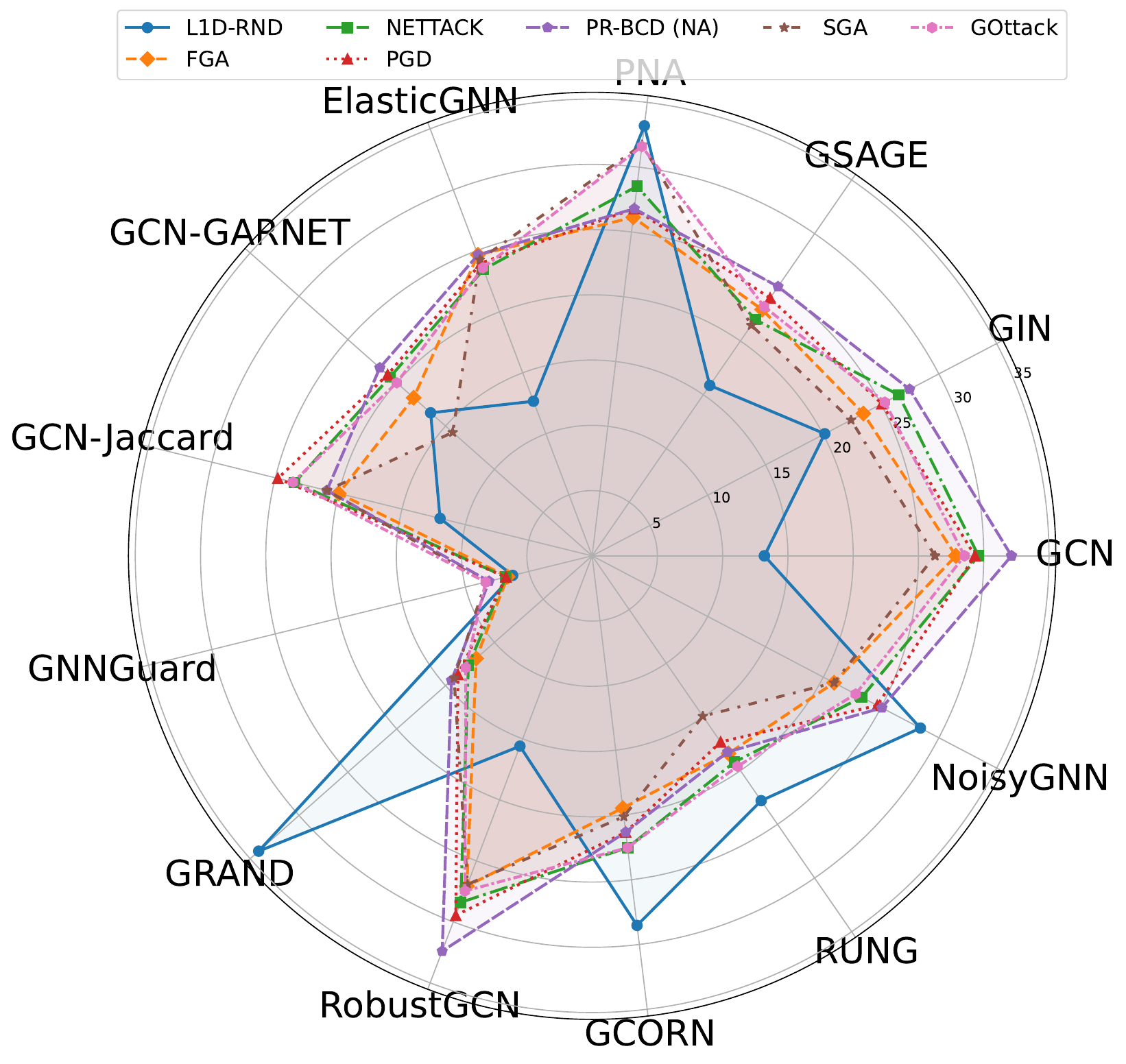}\label{fig:Cora_Evision_Radar_Plot}}
    \subfigure[\cora]{\includegraphics[width=0.45\textwidth]{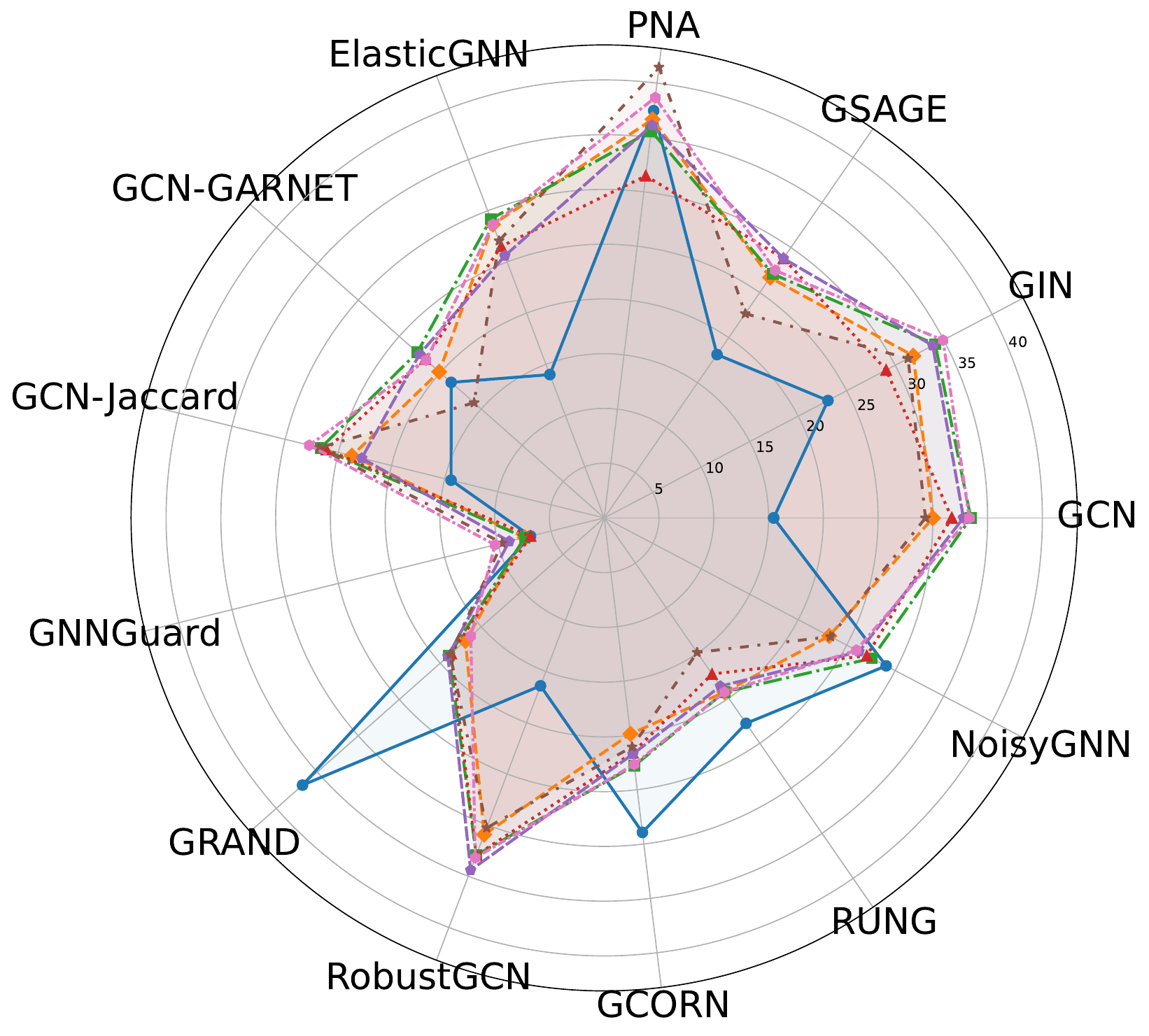}\label{fig:Cora_Poision_Radar_Plot}}
     
    \subfigure[\citeseer]{\includegraphics[width=0.45\textwidth]{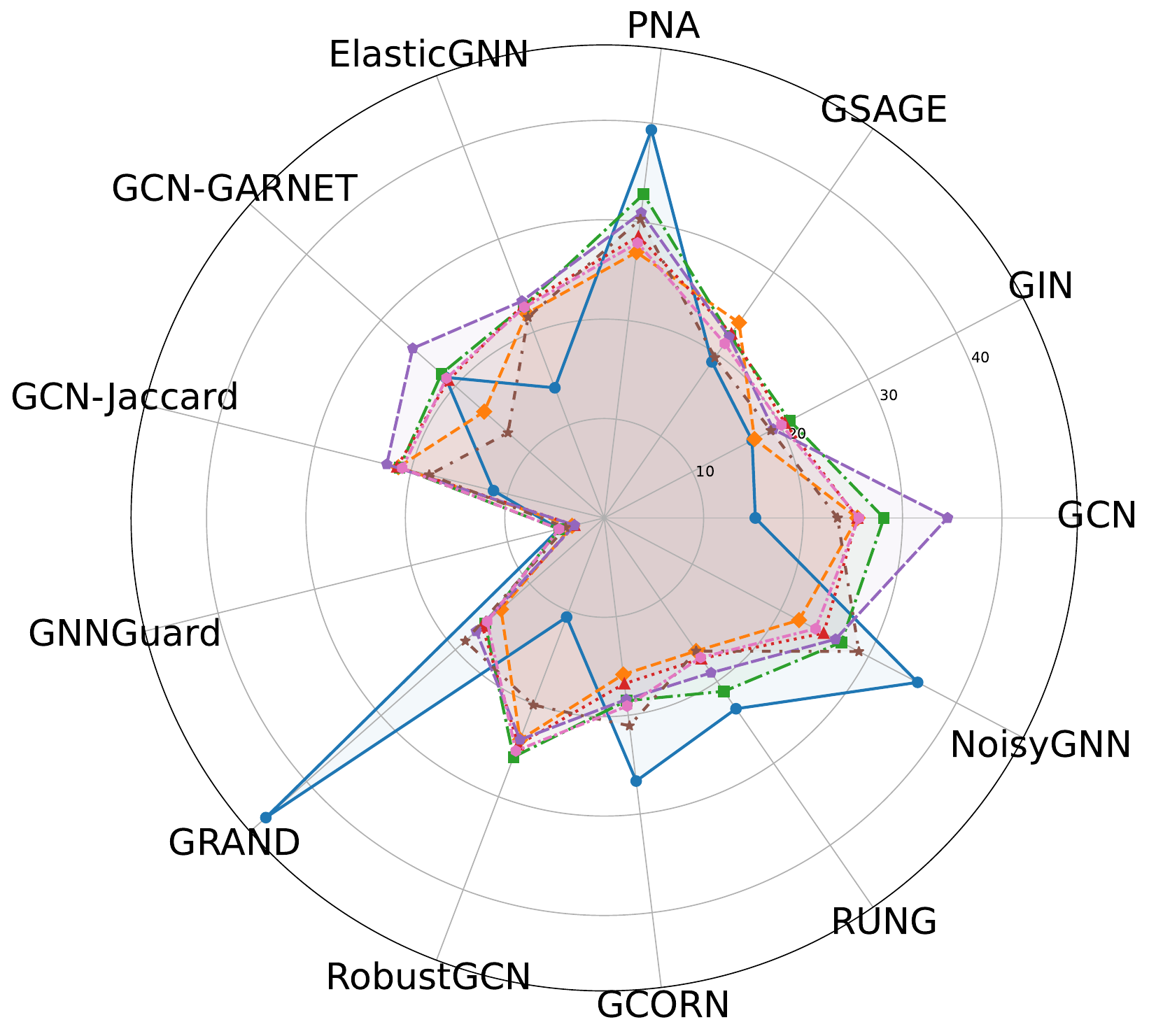}\label{fig:Citeseer_Evision_Radar_Plot}}
    \subfigure[\citeseer]{\includegraphics[width=0.45\textwidth]{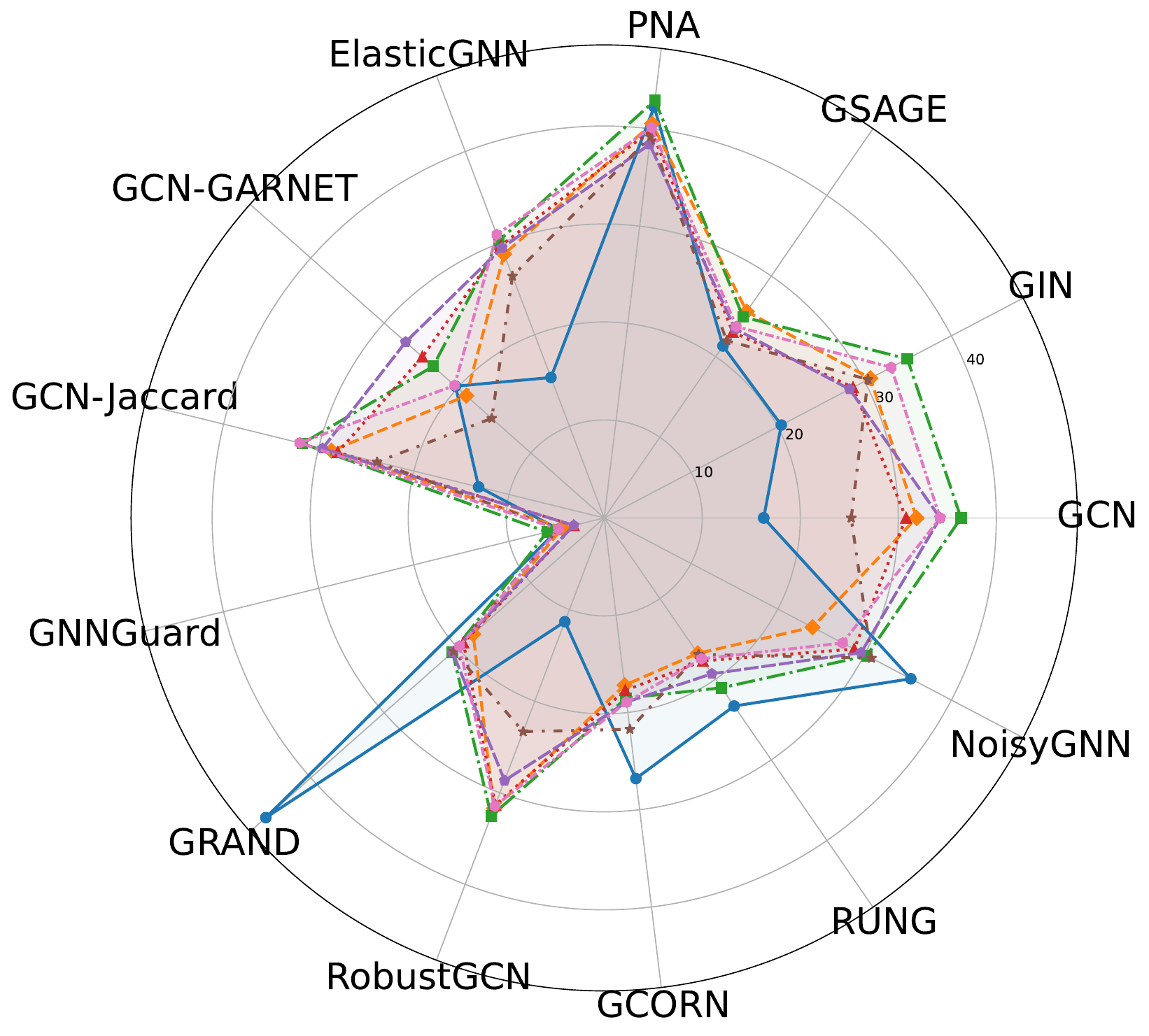}\label{fig:Citeseer_Poision_Radar_Plot}}
    
    \subfigure[\pubmed]{\includegraphics[width=0.45\textwidth]{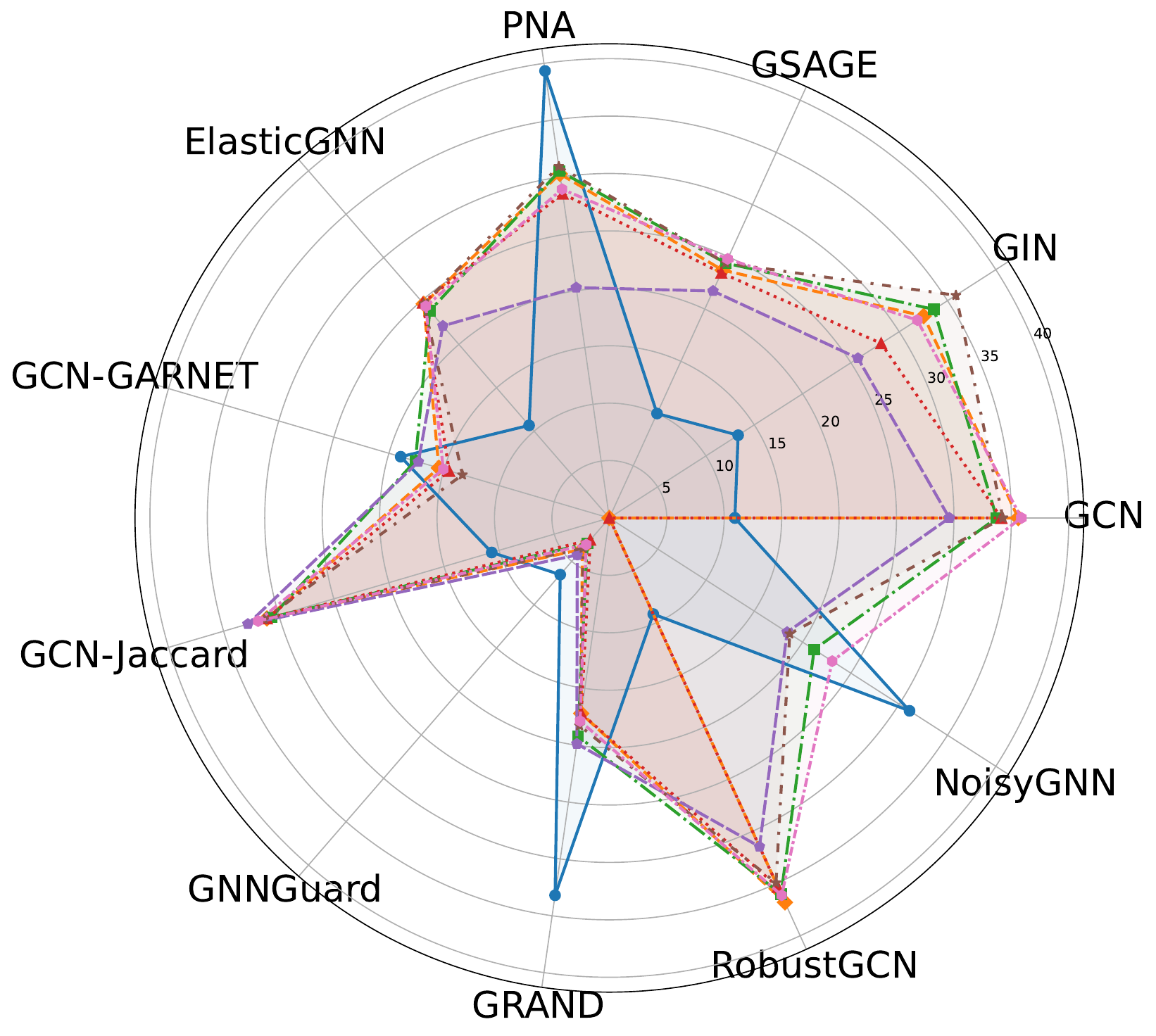}\label{fig:Pubmed_Evision_Radar_Plot}}
    \subfigure[\pubmed]{\includegraphics[width=0.45\textwidth]{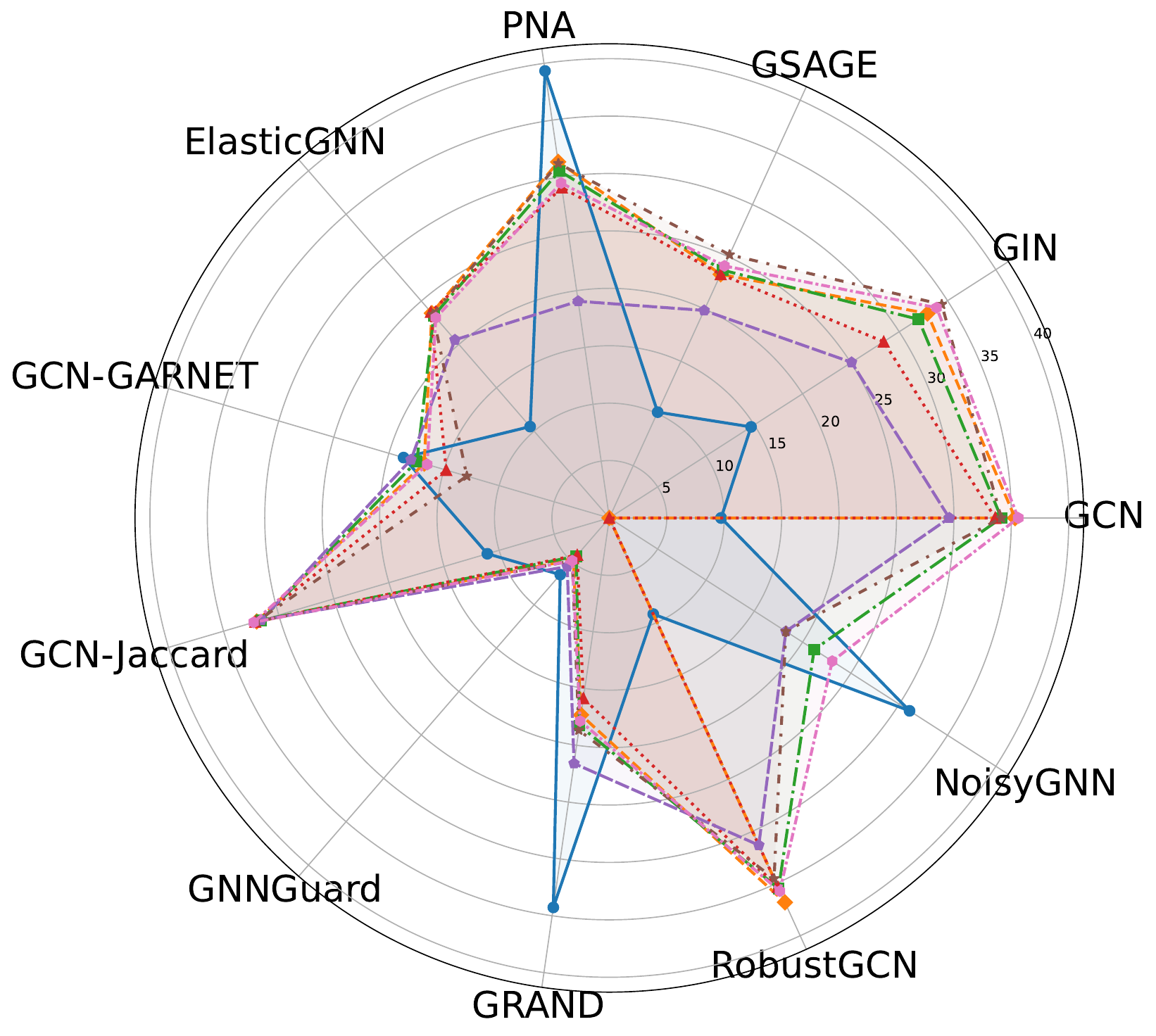}\label{fig:Pubmed_poision_Radar_Plot}}
    
    \caption{{Misclassification rates of defense and non-defense models under different adversarial attacks (budget $1$) across three homophily datasets. Left column: evasion setting; Right column: poison setting.}}
    \label{fig:homophily_radar_plots}
\end{figure*}

\begin{figure*}[ht!]
    \centering
    \subfigure[\squirrel]{\includegraphics[width=0.45\textwidth]{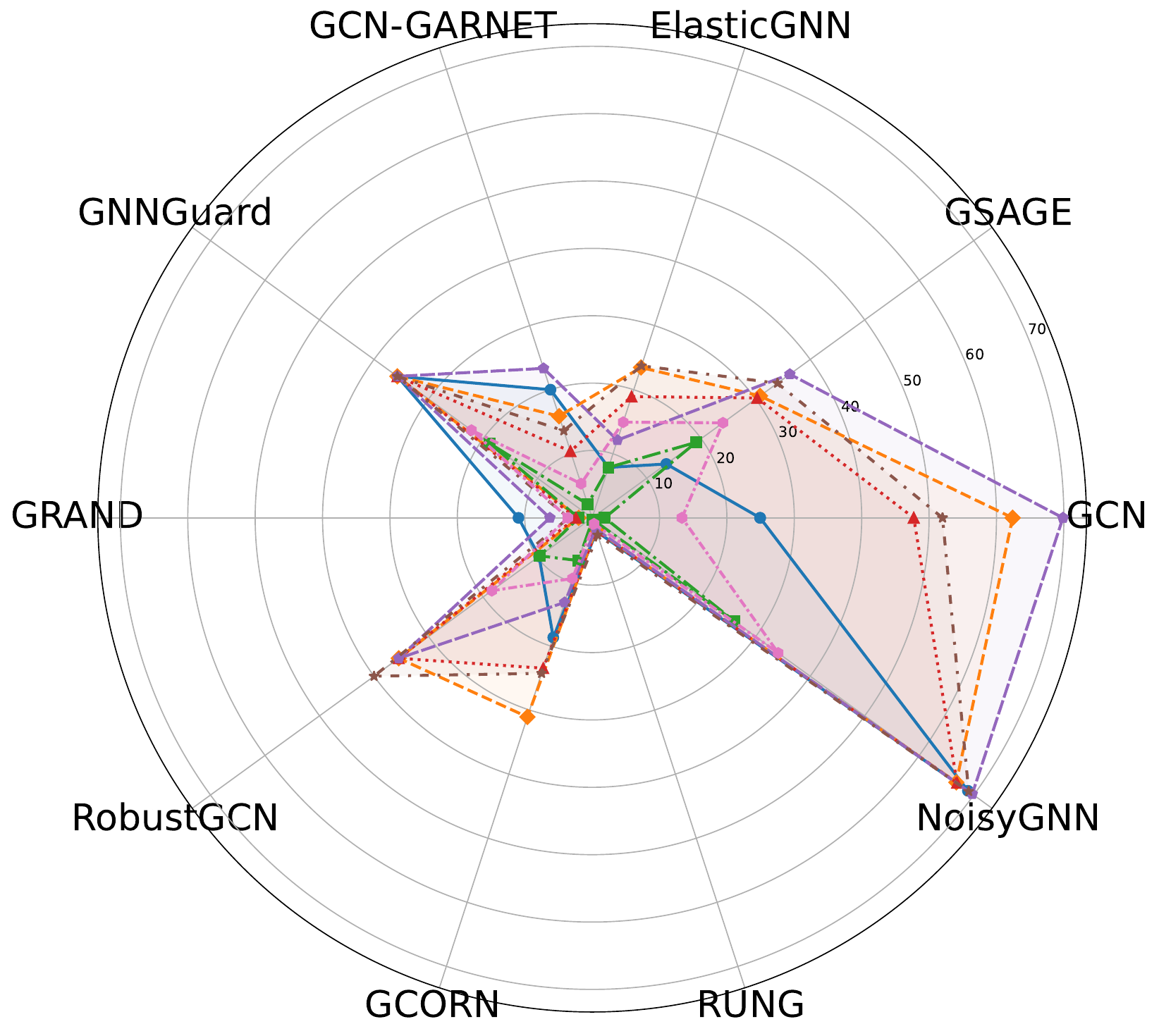}\label{fig:SQUIRREL_Evision_Radar_Plot}}
    \subfigure[\squirrel]{\includegraphics[width=0.45\textwidth]{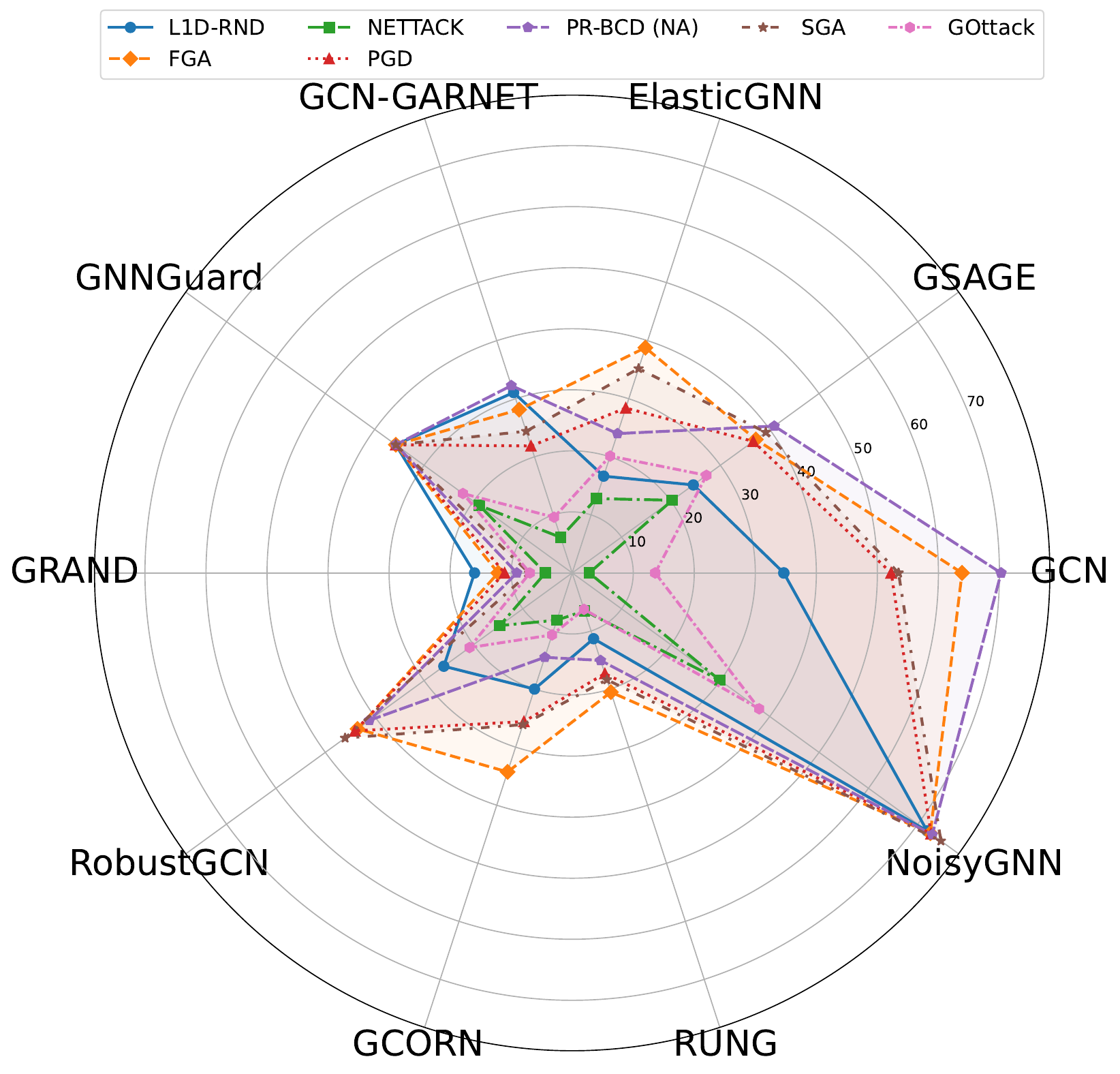}\label{fig:SQUIRREL_Poision_Radar_Plot}}
    
    \subfigure[\chameleon]{\includegraphics[width=0.45\textwidth]{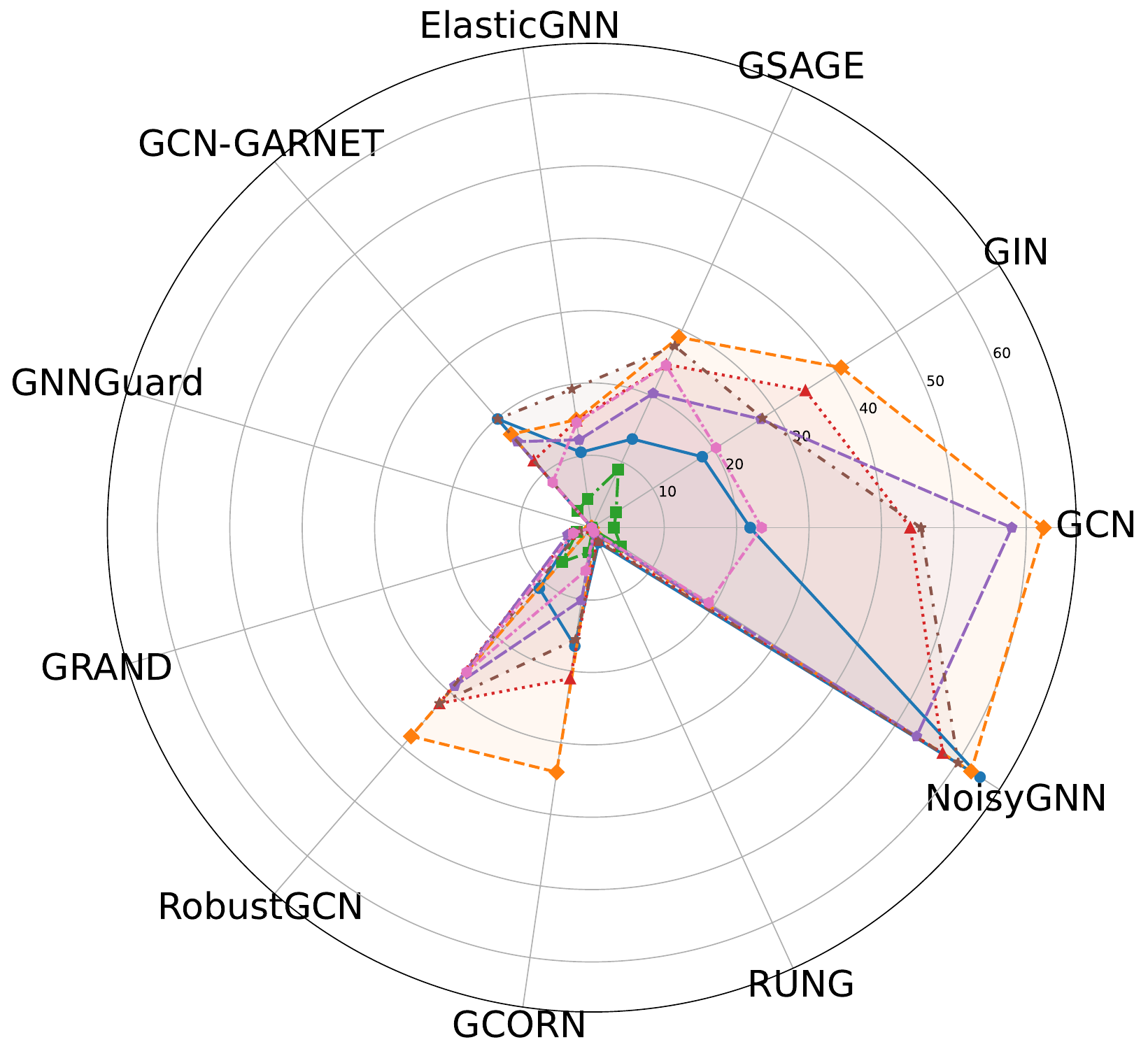}\label{fig:CHAMELEON_Evision_Radar_Plot}}
     \subfigure[\chameleon]{\includegraphics[width=0.45\textwidth]{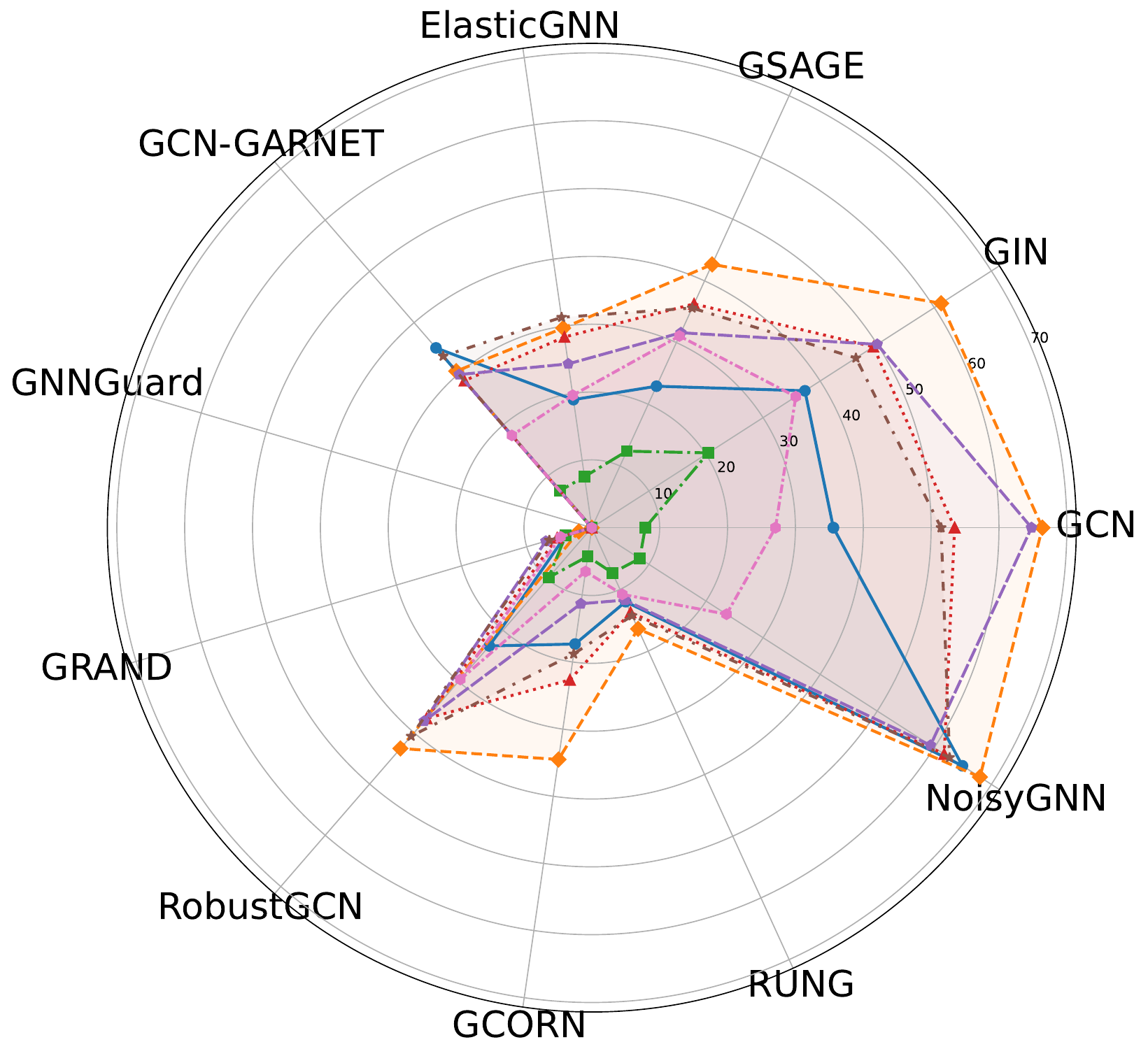}\label{fig:CHAMELEON_Poision_Radar_Plot}}

    \caption{{Misclassification rates of defense and non-defense models under different adversarial attacks (budget $1$) across two heterophily datasets and one large-scale dataset. Left column: evasion setting; Right column: poison setting.}}
    \label{fig:heterophily_radar_plots}
\end{figure*}

{We also present the misclassification rates of defense and vanilla models under different adversarial attacks across three homophily and two heterophily datasets, as shown in Figure~\ref{fig:homophily_radar_plots} and Figure~\ref{fig:heterophily_radar_plots}  for the evasion and poisoning settings, respectively.}

{\textbf{Understanding defense success and failure across graph structures.} Tables~\ref{tab:defense_deeper_evasion_taxonomy} and ~\ref{tab:defense_deeper_poison_taxonomy} provide deeper insight into why certain defenses succeed or fail under different graph structures. Similarity-based pruning methods such as GCN-Jaccard and GNNGuard collapse under heterophily because neighboring nodes are inherently dissimilar. For instance, on \chameleon , all attack models produce a misclassification rate of $0.00$ under GNNGuard as the attack budget increases, since GNNGuard prunes nearly all edges in heterophilous graphs (refer Table~\ref{tab:eval_defense_heter_p1}). Training-based defenses that rely on feature smoothness, such as GRAND, also deteriorate sharply under heterophily. In the evasion setting, GRAND drops from $19.05\%$ on homophily graphs to $3.33\%$  on heterophily graphs, a $471.40\%$ relative decrease. In the poisoning setting, it falls from $21.63\%$ to $7.17\%$, corresponding to a $201.56\%$ relative decrease, revealing its sensitivity to structure–feature mismatch. Defenses like RUNG and GCORN also become unstable when homophily is low. RUNG decreased from $18.63\%$ to $1.47\%$ in evasion ($1169.72\%$ decline) and from $18.88\%$ to $12.98\%$ in poisoning ($45.49\%$ decrease). GCORN also drops from $20.88\%$ to $16.70\%$ in evasion ($25.03\%$ decline) and $21.32\%$ to $18.20\%$ in poisoning ($17.16\%$ decrease), highlighting their struggles to generalize beyond homophilous settings.
In contrast, defenses that make fewer homophily assumptions such as RobustGCN, ElasticGNN, and to some extent GARNET exhibit more stable behavior across diverse datasets. In the poisoning setting, GARNET improves from $19.47\%$ to  $24.02\%$, yielding an $18.96\%$ gain. RobustGCN remains steady, shifting only slightly from $26.71\%$ to  $25.66\%$ in evasion and increasing from $29.52\%$ to  $32.65\%$ in poisoning ($9.59\%$ gain). Meanwhile, ElasticGNN, although experiencing some degradation from $21.99\%$ to $14.27\%$ in evasion and $24.71\%$ to  $24.02\%$ in poisoning ($2.88\%$ decline) still performs more reliably than heavily homophily-dependent defenses. These patterns collectively confirm that defenses grounded in rigid smoothing or homophily assumptions tend to fail on heterophily graphs, whereas structurally flexible defenses remain more resilient.
}

\begin{table*}[ht]
    \centering
    \caption{{Selected defenses and their average success rates (evasion setting) under budget 1 across five datasets. GCN-Jaccard excluded for SQUIRREL and CHAMELEON because of an implementation error. Higher is better. Best performance in \textbf{Bold}, second best is \underline{underline}.}}
    \label{tab:defense_deeper_evasion_taxonomy}
    \vspace{0.5em}
    \resizebox{0.98\linewidth}{!}{
    \begin{tabular}{lll|l|ccccc}
        \toprule \toprule
        \multicolumn{3}{c|}{Taxonomy} 
        & Selected Defense 
        & \cora & \citeseer & \pubmed & \squirrel & \chameleon \\
        \midrule
        
        \multirow{2}{*}{\begin{tabular}[c]{@{}l@{}}Improving\\ graph\end{tabular}}
        & \multicolumn{2}{l|}{Unsupervised} 
        & Jaccard-GCN
        &  20.81 $\pm$ 3.94 & 19.58 $\pm$ 3.53 & \underline{28.49 $\pm$ 7.30} & -- & --  \\
        
        \cmidrule{2-9}
        
        & \multicolumn{2}{l|}{Supervised} 
        & GARNET
        & 18.91 $\pm$ 2.50  & 20.01 $\pm$ 3.87  & 16.03 $\pm$ 1.82  & 12.95 $\pm$ 7.05  & 13.73 $\pm$ 5.80   \\
        
        \midrule

        \multicolumn{3}{l|}{Improving training} 
        & \begin{tabular}[c]{@{}l@{}} 
        GRAND  \end{tabular}
        & 16.26 $\pm$ 7.33 & 20.47 $\pm$ 10.29  & 20.40 $\pm$ 5.30  & 4.38 $\pm$ 3.00  & 2.28 $\pm$ 1.03  \\ 
        \cmidrule{4-9}

        &  & & \begin{tabular}[c]{@{}l@{}} 
        GCORN \end{tabular}
        &  22.26 $\pm$ 2.76 & 19.48 $\pm$ 3.30 & OOR & 18.11 $\pm$ 8.14 & 15.27 $\pm$ 9.58 \\
        \cmidrule{4-9}
         &  & & \begin{tabular}[c]{@{}l@{}} 
         NoisyGNN \end{tabular}
        &  \underline{23.73 $\pm$ 2.42}  &  \textbf{26.97 $\pm$ 4.05}  & 16.05 $\pm$ 10.88   & \textbf{57.38 $\pm$ 17.40}  & \textbf{45.86 $\pm$ 21.98}   \\

        \midrule

        \multirow{5}{*}{\begin{tabular}[c]{@{}l@{}}Improving\\ architecture\end{tabular}}
        & \multirow{5}{*}{\begin{tabular}[c]{@{}l@{}}Adaptively\\ weighting\\ edges\end{tabular}}
        & Rule-based 
        & GNNGuard  
        & 7.358 $\pm$ 0.84 &  3.9 $\pm$ 0.70 & 3.77 $\pm$ 1.23  & \underline{31.36 $\pm$ 6.95}  & 0.00 $\pm$ 0.00 \\
        \cmidrule{3-9}

        & & Probabilistic 
        & RobustGCN  
        & \textbf{26.76 $\pm$ 4.88}  & \underline{21.96 $\pm$ 4.89}  & \textbf{31.40 $\pm$ 9.20}  & 26.30 $\pm$ 12.28  &  \underline{25.00 $\pm$ 10.93} \\
        \cmidrule{3-9}

        & & Robust agg. 
        & \begin{tabular}[c]{@{}l@{}} ElasticGNN \end{tabular}
        &  22.47 $\pm$ 4.02 &  21.33 $\pm$ 3.04  & 22.17 $\pm$ 4.77  & 15.56 $\pm$ 6.22  &  12.97 $\pm$ 4.44 \\
        \cmidrule{4-9}
         & & & \begin{tabular}[c]{@{}l@{}} RUNG \end{tabular}
        &  18.64 $\pm$ 2.21 & 18.61 $\pm$ 2.51  & OOR & 1.65 $\pm$ 0.74 & 1.27 $\pm$ 0.73 \\

        \bottomrule \bottomrule
    \end{tabular}
    }
\end{table*}

\begin{table*}[ht]
    \centering
    \caption{{Selected defenses and their average success rates (poison setting) under budget 1 across five datasets. GCN-Jaccard excluded for SQUIRREL and CHAMELEON because of an implementation error. Higher is better. Best performance in \textbf{Bold}, second best is \underline{underline}.}}
    \label{tab:defense_deeper_poison_taxonomy}
    \vspace{0.5em}
    \resizebox{0.98\linewidth}{!}{
    \begin{tabular}{lll|l|ccccc}
        \toprule \toprule
        \multicolumn{3}{c|}{Taxonomy} 
        & Selected Defense 
        & \cora & \citeseer & \pubmed & \squirrel & \chameleon \\
        \midrule
        
        \multirow{2}{*}{\begin{tabular}[c]{@{}l@{}}Improving\\ graph\end{tabular}}
        & \multicolumn{2}{l|}{Unsupervised} 
        & Jaccard-GCN$^*$
        &  24.03 $\pm$ 4.26 & 26.72 $\pm$ 6.06 & \underline{28.95 $\pm$ 7.30} & -- & --  \\
        
        \cmidrule{2-9}
        
        & \multicolumn{2}{l|}{Supervised} 
        & GARNET
        & 20.51 $\pm$ 2.32  & 21.51 $\pm$ 3.65  & 16.45 $\pm$ 1.83  & 21.92 $\pm$ 9.53  & 26.11 $\pm$ 9.24   \\
        
        \midrule

        \multicolumn{3}{l|}{Improving training} 
        & \begin{tabular}[c]{@{}l@{}} 
        GRAND  \end{tabular}
        & 20.78 $\pm$ 6.61 & 23.54 $\pm$ 9.26  & 20.55 $\pm$ 5.82  & 9.52 $\pm$ 3.59  & 4.81 $\pm$ 1.58  \\ 
        \cmidrule{4-9}

        &  & & \begin{tabular}[c]{@{}l@{}} 
        GCORN \end{tabular}
        &  22.66 $\pm$ 2.71 & 19.97 $\pm$ 3.08 & OOR & 19.90 $\pm$ 8.70 & 16.49 $\pm$ 9.59 \\
        \cmidrule{4-9}
         &  & & \begin{tabular}[c]{@{}l@{}} 
         NoisyGNN \end{tabular}
        &  \underline{26.07 $\pm$ 2.00}  &  \textbf{29.46 $\pm$ 3.18}  & 15.98 $\pm$ 10.86   & \textbf{61.77 $\pm$ 17.79}  & \textbf{49.80 $\pm$ 21.90}   \\

        \midrule

        \multirow{5}{*}{\begin{tabular}[c]{@{}l@{}}Improving\\ architecture\end{tabular}}
        & \multirow{5}{*}{\begin{tabular}[c]{@{}l@{}}Adaptively\\ weighting\\ edges\end{tabular}}
        & Rule-based 
        & GNNGuard  
        & 8.22 $\pm$ 1.25 &  4.45 $\pm$ 0.91 & 5.06 $\pm$ 0.71  & 31.36 $\pm$ 6.95  & 0.00 $\pm$ 0.00 \\
        \cmidrule{3-9}

        & & Probabilistic 
        & RobustGCN  
        & \textbf{30.15 $\pm$ 5.76}  & 27.19 $\pm$ 7.10  & \textbf{31.19 $\pm$ 9.10}  & \underline{33.71 $\pm$ 11.92}  &  \underline{31.58 $\pm$ 10.94} \\
        \cmidrule{3-9}

        & & Robust agg. 
        & \begin{tabular}[c]{@{}l@{}} ElasticGNN \end{tabular}
        &  25.67 $\pm$ 4.91 &  \underline{27.27 $\pm$ 5.05}  & 21.18 $\pm$ 4.46  & 25.14 $\pm$ 8.85  &  22.89 $\pm$ 7.64 \\
        \cmidrule{4-9}
         & & & \begin{tabular}[c]{@{}l@{}} RUNG \end{tabular}
        &  18.81 $\pm$ 2.20 & 18.95 $\pm$ 2.27 & OOR & 13.63 $\pm$ 5.29 & 12.32 $\pm$ 2.67 \\

        \bottomrule \bottomrule
    \end{tabular}
    }
\end{table*}


\subsection{{Computational Costs of Adversarial Attack}}
\label{sec:costs}
Although Nettack delivers the best performance, it is significantly more computationally expensive than GOttack, SGA, and \prbcdNA. Table~\ref{tab:time_record} shows that GOttack is the fastest among gradient-based adversarial attack methods on smaller datasets, such as \cora and \citeseer, while SGA outperforms others on the \pubmed dataset due to its focus on the local information of target nodes. While \prbcdNA is not the fastest method on the datasets studied, its time complexity is independent of both the number of nodes and edges, making it advantageous for large-scale datasets like OGB-arxiv~\citep{weihua2020open}. In any case, the \textbf{reliance on results from only three datasets, as is common among the studied models, presents a significant limitation and raises concerns about the robustness and generalizability of the field's findings}. The authors are not aware of any other machine learning domain where the majority of results rely on such a limited number of datasets.

\begin{table} 
    \centering
    \caption{Average total attack time on 50 target nodes (in Seconds). Smaller is better. Best time in \textbf{Bold}, second best is \underline{underline}.}
    \label{tab:time_record}
    \resizebox{0.95\linewidth}{!}{
    \begin{tabular}{ccc | rrrrr}
        \toprule \toprule
        &&$\Delta \rightarrow$ & 1 & 2 & 3 & 4 & 5  \\
        \midrule
        \multirow{7}{*}{\rotatebox{90}{\textbf{\cora}}} &  \multirow{6}{*}{\textbf{Global}}& \propseRND & \textbf{0.17 $\pm$ 0.08}  & \textbf{0.29 $\pm$ 0.06} & \textbf{0.49 $\pm$ 0.11} & \textbf{0.65 $\pm$ 0.10} & \textbf{0.92 $\pm$ 0.13}\\

        && FGA & \underline{0.36 $\pm$ 0.03} & \underline{0.67 $\pm$ 0.06} & \underline{0.98 $\pm$ 0.10} & \underline{1.30 $\pm$ 0.12} & \underline{1.57 $\pm$ 0.17}\\
        && NETTACK & 19.72 $\pm$ 1.46 & 29.75 $\pm$ 1.00 & 41.01 $\pm$ 2.11 & 50.90 $\pm$ 2.51 & 62.80 $\pm$ 2.06\\
        && PGD (NA) & 110.86 $\pm$ 7.63 & 104.17 $\pm$ 8.61 & 95.58 $\pm$ 10.93 & 95.83 $\pm$ 5.90 & 95.09 $\pm$ 14.17\\
        && \prbcdNA & 159.17 $\pm$ 26.05 & 158.33 $\pm$ 19.49 & 163.96 $\pm$ 20.16 & 164.16 $\pm$ 17.39 & 163.26 $\pm$ 21.55\\
        && GOttack & 8.10 $\pm$ 1.03 & 11.09 $\pm$ 0.28 & 14.57 $\pm$ 0.32 & 17.73 $\pm$ 0.55 & 21.43 $\pm$ 0.83\\
        \cline{2-8}
        &\multirow{1}{*}{\textbf{Local}}& SGA & 16.90 $\pm$ 0.37 & 17.73 $\pm$ 0.43 & 18.44 $\pm$ 0.39 & 18.99 $\pm$ 0.46 & 19.23 $\pm$ 0.62\\

        \hline

        \multirow{7}{*}{\rotatebox{90}{\textbf{\citeseer}}}&  \multirow{6}{*}{\textbf{Global}} &   \propseRND & \textbf{0.15 $\pm$ 0.04}  & \textbf{0.29 $\pm$ 0.12} & \textbf{0.36 $\pm$ 0.09} & \textbf{0.37 $\pm$ 0.09} & \textbf{0.42 $\pm$ 0.07}\\

        && FGA & \underline{0.25 $\pm$ 0.04} & \underline{0.36 $\pm$ 0.08} & \underline{0.50 $\pm$ 0.06} & \underline{0.56 $\pm$ 0.03} & \underline{0.58 $\pm$ 0.04}\\
        && NETTACK & 22.84 $\pm$ 1.37 & 29.27 $\pm$ 0.91 & 35.83 $\pm$ 1.24 & 42.80 $\pm$ 1.27 & 49.48 $\pm$ 1.62\\
        && PGD (NA) & 134.79 $\pm$ 6.95 & 129.90 $\pm$ 8.68 & 126.59 $\pm$ 8.77 & 127.51 $\pm$ 9.37 & 121.21 $\pm$ 10.59\\
        && \prbcdNA & 209.56 $\pm$ 40.58 & 215.76 $\pm$ 45.17 & 224.09 $\pm$ 43.40 & 199.46 $\pm$ 36.82 & 195.49 $\pm$ 34.81\\
        && GOttack & 6.39 $\pm$ 1.08 & 8.18 $\pm$ 0.17 & 10.41 $\pm$ 0.23 & 12.50 $\pm$ 0.34 & 14.66 $\pm$ 0.36\\
        \cline{2-8}

        &\multirow{1}{*}{\textbf{Local}}& SGA & 17.39 $\pm$ 0.56 & 17.97 $\pm$ 0.49 & 19.02 $\pm$ 0.52 & 19.13 $\pm$ 0.60 & 19.46 $\pm$ 0.60\\

        \hline

        \multirow{7}{*}{\rotatebox{90}{\textbf{\pubmed}}}&  \multirow{6}{*}{\textbf{Global}} & \propseRND & \textbf{1.81 $\pm$ 0.54}  & \textbf{1.72 $\pm$ 0.42} & \textbf{1.77 $\pm$ 0.44} & \textbf{1.75 $\pm$ 0.37} & \textbf{1.64 $\pm$ 0.45}\\

        && FGA & \underline{13.68 $\pm$ 0.29} & \underline{27.72 $\pm$ 0.48} & \underline{39.25 $\pm$ 3.37} & \underline{47.43 $\pm$ 0.70} & \underline{59.28 $\pm$ 0.87}\\
        && NETTACK & 423.84 $\pm$ 18.17 & 765.05 $\pm$ 29.30 & 1095.95 $\pm$ 31.13 & 1426.95 $\pm$ 42.03 & 1753.58 $\pm$ 67.42\\
        && PGD (NA) & 4114.65 $\pm$ 182.27 & 3922.62 $\pm$ 159.91 & 3814.07 $\pm$ 112.16 & 3721.93 $\pm$ 108.08 & 3689.32 $\pm$ 103.25\\
        && \prbcdNA & 173.83 $\pm$ 6.62 & 177.46 $\pm$ 7.42 & 173.28 $\pm$ 3.54 & 171.80 $\pm$ 8.22 & 173.10 $\pm$ 7.56\\
        && GOttack & 153.77 $\pm$ 3.32 & 252.04 $\pm$ 6.47 & 355.97 $\pm$ 10.38 & 452.34 $\pm$ 10.73 & 553.61 $\pm$ 13.92\\
        \cline{2-8}

        &\multirow{1}{*}{\textbf{Local}}& SGA & 18.09 $\pm$ 0.40 & 17.71 $\pm$ 0.63 & 19.00 $\pm$ 0.68 & 19.85 $\pm$ 0.93 & 19.66 $\pm$ 0.99\\

        \bottomrule \bottomrule
    \end{tabular}}
\end{table}

{\textbf{Time complexity breakdown.} To better understand the computational cost of different adversarial attack methods, we break attack algorithms into three main stages according to their implementations provided by DeepRobust\cite{li2020deeprobust}: \textbf{Surrogate}, \textbf{Pre-attack} and \textbf{Attacks}. \textbf{Surrogate} stage includes the time required to train and set up the surrogate used in the attack. \textbf{Pre-attack} involves operations: computing logits of target nodes and normalizing the adjacency matrix in NETTACK and GOttack; performing project gradient descent training and projected randomized block coordinate descent in PGD and PR-BCD, respectively; retrieving subgraphs in SGA. Finally, in \textbf{Attacks} phases, attack algorithms find the best edges to flip according to surrogate loss of all potential edges in NETTACK and selected potential edges in GOttack, while PR-BCD and PGD perform a Bernoulli sample to select the optimal edge to flip.}

{
Table~\ref{tab:time_record_breakdown} shows the time complexity breakdown of three main stages of adversarial attacks for a budget of 1 across three datasets (\cora, \pubmed, and \ogb). Across all datasets, \propseRND is the fastest method, achieving close to zero seconds in all three stages. Its efficiency is due to its simple random edge selection strategy, which avoids surrogate training and iterative optimization entirely. Benefiting from fast gradient-based updates, FGA spends significantly less time on \textbf{Pre-attack} and \textbf{Attack} stages on \cora and \pubmed datasets and achieves the second fastest on \cora and \pubmed. However, FGA need to maintain a dense adjacency matrix to compute the gradient, which requires more than 100GiB on \ogb, causing Out-Of-Memory (OOM).
On \cora, \pubmed and \ogb, SGA is established to be one of the most efficient attacks, as it limits the pre-computation to small subgraphs, resulting in minimal in \textbf{Pre-attak} and \textbf{Attack} stages. In contrast, PGD's runtime on \cora and \pubmed is dominated by project gradient descent training in \textbf{Pre-attack} stage, accounting for 96\% of total runtime on \pubmed. In addition, PGD encounters the same memory constraints as FGA on \ogb. Instead of project gradient descent, PR-BCD utilize projected randomized block coordinate descent, which has been shown to reduce the dominant effect of \textbf{Pre-attack} stage to 55\% on \pubmed.
Methods such as NETTACK and GOttack are significantly slower, particularly on larger datasets. \textbf{Pre-attack} and \textbf{Attack} stages dominate the runtime in NETTACK due to performing optimization to select the best edge to flip over all potential edges. GOttack's \textbf{Pre-attack} and \textbf{Attack} stages' runtime has shown improvements over NETTACK due to reducing the search space by wisely filtering potential edges.
}

Table \ref{tab:gpu-memory} reports the GPU memory usage (in MB) of different attack methods under budget 1 across five benchmark datasets. As expected, gradient-based methods such as PGD (NA) and FGA consume substantially more memory, with peak usage above 400 MB on CiteSeer and PubMed. PRBCD (NA) reduces the cost relative to PGD but still requires more than 130 MB across datasets. In contrast, traditional structure-based methods like NETTACK and GOttack are much lighter, staying around 60–66 MB, with GOttack consistently achieving the second lowest footprint. The best efficiency is obtained by L$^1$D-RND, which uses less than 40 MB on all datasets. These results confirm that L$^1$D-RND provides the most memory-efficient attack, while GOttack is a competitive second choice.

\begin{table}[t]
\centering
\caption{GPU memory usage in MB of attacks on budget 1. Smaller is better. Best GPU memory usage is in \textbf{bold}, and the second best is in \underline{underline}.}
\label{tab:gpu-memory}
\begin{tabular}{lrrrrr}
\toprule
Method & CHAMELEON & CORA & CITESEER & SQUIRREL & PUBMED \\
\midrule
SGA & 90.824 & 105.619 & 92.143 & 92.312 & 92.537 \\
NETTACK & 65.747 & 64.909 & 66.041 & 64.258 & 62.898 \\
FGA & 335.235 & 331.022 & 413.564 & 328.964 & 328.773 \\
PGD (NA) & 421.835 & 417.622 & 499.784 & 415.564 & 415.374 \\
PRBCD (NA) & 196.014 & 195.169 & 159.254 & 131.659 & 157.704 \\
L$^1$D-RND & \textbf{39.568} & \textbf{39.553} & \textbf{38.901} & \textbf{37.825} & \textbf{38.703} \\
GOttack & \underline{64.984} & \underline{63.315} & \underline{65.275} & \underline{62.536} & \underline{62.510} \\
\bottomrule
\end{tabular}
\end{table} 

\begin{table}[h!]
\centering
\caption{{Time break down of adversarial attacks on budget 1 (in Seconds). Smaller is better. Best time in \textbf{Bold}, second best is \underline{underline}.}}
\label{tab:time_record_breakdown}
\resizebox{0.8\textwidth}{!}{%
\begin{tabular}{c c c c c}
\hline
&Method  & Surrogate & Pre-attack & Attack \\
\hline
\multirow{7}{*}{\rotatebox{90}{\textbf{\cora}}} &\propseRND & \textbf{0.00000 $\pm$ 0.00000} & \textbf{0.00006 $\pm$ 0.00000} & \textbf{0.00213 $\pm$ 0.00245} \\
&FGA & \underline{0.73066 $\pm$ 0.00001} & \underline{0.00011 $\pm$ 0.00000} & 0.00531 $\pm$ 0.00013 \\

&NETTACK & 1.23075 $\pm$ 0.00011 & 0.16365 $\pm$ 0.06042 & 0.20430 $\pm$ 0.14573 \\
&PGD (NA) & 0.98343 $\pm$ 0.00003 & 3.69364 $\pm$ 0.82624 & 0.04686 $\pm$ 0.00793 \\
&PRBCD (NA) & 1.43114 $\pm$ 0.00000 & 4.31843 $\pm$ 0.16030 & 0.06399 $\pm$ 0.01213 \\
&SGA & 1.62636 $\pm$ 0.00005 & 0.32616 $\pm$ 0.05827 & \underline{0.00323 $\pm$ 0.00016} \\
&GOttack & 1.16423 $\pm$ 0.00007 & 0.08035 $\pm$ 0.05181 & 0.07113 $\pm$ 0.04261 \\

\midrule
\multirow{7}{*}{\rotatebox{90}{\textbf{\pubmed}}} &\propseRND  & \textbf{0.00000 $\pm$ 0.00000} & \textbf{0.00007 $\pm$ 0.00000} & \textbf{0.00387 $\pm$ 0.00429} \\
&FGA  & \underline{1.86278 $\pm$ 0.00002} & \underline{0.00045 $\pm$ 0.00022} & {0.27881 $\pm$ 0.00710} \\
&NETTACK & 4.45833 $\pm$ 0.00026 & 1.43834 $\pm$ 0.04844 & 6.53812 $\pm$ 5.08244 \\
&PGD (NA) & 2.75710 $\pm$ 0.00006 & 155.74790 $\pm$ 32.54714 & 2.74592 $\pm$ 0.33127 \\
&PRBCD (NA) & 2.57097 $\pm$ 0.00000 & 3.29684 $\pm$ 0.42744 & 0.06547 $\pm$ 0.01111 \\
&SGA & 10.47330 $\pm$ 0.00007 & 0.33145 $\pm$ 0.05469 & \underline{0.00329 $\pm$ 0.00013} \\
&GOttack & 1.91390 $\pm$ 0.00005 & 0.91947 $\pm$ 0.05263 & 2.03039 $\pm$ 1.37443 \\

\midrule

\multirow{7}{*}{\rotatebox{90}{\textbf{\ogb}}}  & \propseRND & \textbf{0.00000 $\pm$ 0.00000} & \textbf{0.00010 $\pm$ 0.00001} & \textbf{0.13041 $\pm$ 0.23359} \\
&FGA  & OOM & OOM & OOM \\
&NETTACK & 12.54029 $\pm$ 0.28053 & 362.42469 $\pm$ 39.61532 & 894.00637 $\pm$ 8.04044 \\
&PGD (NA)  & OOM & OOM & OOM \\
&PRBCD (NA) & 7.86793 $\pm$ 0.00000 & 12.93282 $\pm$ 0.14348 & 0.34241 $\pm$ 0.01064 \\
&SGA & \underline{3.85193 $\pm$ 0.00870} & \underline{0.50283 $\pm$ 0.20965} & \underline{0.01772 $\pm$ 0.01850} \\
&GOttack & 12.80845 $\pm$ 0.87578 & 326.28377 $\pm$ 27.37981 & 288.88327 $\pm$ 7.75647 \\

\bottomrule

\end{tabular}
}
\end{table}

\subsection{Computational Costs of Defense}
\label{sec:costs_defense}

\begin{table}[t]
\centering
\caption{{Average total time to train defense models (in Seconds). Smaller is better. Best time in \textbf{Bold}, second best is \underline{underline}.}}
\label{tab:defense_time}
\resizebox{\textwidth}{!}{%
\begin{tabular}{rccccc}
\toprule
\textbf{Defense} & \cora & \citeseer & \pubmed & \squirrel & \chameleon \\
\midrule

ElasticGNN & 5.03 $\pm$ 2.78 & 3.92 $\pm$ 2.03 & 8.57 $\pm$ 5.27 & 10.67 $\pm$ 2.79 & 9.72 $\pm$ 5.05\\
GCN-GARNET & 2.56 $\pm$ 2.51 & 3.87 $\pm$ 3.66 & 25.81 $\pm$ 9.39 & 3.45 $\pm$ 2.14 & 3.79 $\pm$ 1.56\\
GCN-Jaccard & \underline{1.24 $\pm$ 0.68} & 1.58 $\pm$ 0.79& -& -& -\\
GNNGuard & 1.91 $\pm$ 0.47 & 1.62 $\pm$ 0.59 & 4.39 $\pm$ 0.84 & 5.42 $\pm$ 0.63 & 2.37 $\pm$ 0.54\\
GRAND & 6.46 $\pm$ 3.00 & 7.50 $\pm$ 2.39 & 93.48 $\pm$ 40.46 & 28.94 $\pm$ 23.13 & 5.42 $\pm$ 6.92\\
RobustGCN & 3.95 $\pm$ 2.99 & \textbf{0.87 $\pm$ 0.12} & \underline{5.54 $\pm$ 1.67} & \underline{2.24 $\pm$ 0.56} & \textbf{1.09 $\pm$ 0.22}\\
GCORN & 54.99 $\pm$ 8.24 & 29.04 $\pm$ 7.89 & 4605.11 $\pm$ 727.34 & 130.53 $\pm$ 134.07 & 15.40 $\pm$ 5.34\\
RUNG & 130.52 $\pm$ 5.34 & 106.62 $\pm$ 10.01 & 4265.07 $\pm$ 2279.45 & 46.06 $\pm$ 20.67 & 7.69 $\pm$ 2.78\\
NoisyGNN & \textbf{0.77 $\pm$ 0.28} & \underline{1.29 $\pm$ 0.49} & \textbf{2.13 $\pm$ 0.50} & \textbf{1.78 $\pm$ 1.35} & \underline{1.70 $\pm$ 0.68}\\

\bottomrule
\end{tabular}
}
\end{table}

{Table~\ref{tab:defense_time} reports the average total training time (in seconds) for various defense models across five benchmark datasets. Overall, NoisyGNN consistently achieves the fastest training times on most datasets, with particularly low values on \cora, \pubmed, and \squirrel. Its efficiency can be attributed to its lightweight architecture and minimal computational overhead, which allows it to scale well even on larger graphs. The second fastest model is generally RobustGCN, which performs competitively on \citeseer, \pubmed, and \chameleon, likely due to its optimized graph convolution operations that reduce redundant computations. In contrast, RUNG is the slowest model by a wide margin, especially on large datasets such as \pubmed and \citeseer. This significant slowdown is likely caused by its complex training procedure and extensive use of robust aggregation mechanisms, which introduce high computational cost. Other models such as GCORN and GRAND also show high training times on large graphs, highlighting the trade-off between defense sophistication and training efficiency.}

\subsection{Feature attack results} \label{sec:feature-attack}

We conduct feature-attack variants of FGA, NETTACK, PGD, PR-BCD, and SGA on vanilla models across homophily datasets. Overall, the results, Table~\ref{tab:feat-attack} show that NETTACK and FGA consistently yield the highest misclassification rates in both evasion and poisoning settings, while SGA and PGD perform substantially worse. We observe a similar trend to that in structural attacks: PR-BCD outperforms the remaining feature-attack methods on \cora and \citeseer for GCN, while PR-BCD performs worse on \pubmed and other vanilla models; NETTACK remains a dominant attack method, with an average rank of $1.52$ across all settings, whereas FGA remains the second best with the average rank of $2.56$; NETTACK achieves superior performance on PNA on all datasets and in both evasion and poison settings;high-degree nodes remain significantly more robust (11.92\% average success rate) than low-degree nodes (94.85\%).

Feature attacks largely mirror the trends observed in structural attacks, with consistent rankings and performance gaps across methods and settings. Given this strong alignment, our study primarily focuses on structural attacks for evaluation and analysis, as they provide representative insights while reducing redundancy. We leave a more in-depth exploration of feature attacks, building upon our current framework, as future work.

\section{Defense Models}\label{gnn}

\textbf{Selection criteria.} In addition to the vanilla attack scenario, we also consider evaluating adversarial attack methods against defense models (defense attack scenario). However, to avoid the significant number of experiences needed to evaluate each adversarial attack to all defense models existing in the literature, we select defense models used to evaluate adversarial attacks, we first select six defense models, one from each taxonomy described in Table~\ref{tab:categorization} to cover the entire spectrum of defense techniques. In this work, we select defense models according to the following criteria: i) highly cited defenses published at renowned venues; ii) publicly available codes, recent; iii) recently published. Following \citeauthor{felix2022robust}'s work, we exclude defense models from \textquote{Robust training} taxonomy from our study, since they require knowing clean graph $\mathcal{G}$, which is typically not available in the poison attack setting.
\begin{table*}[ht]
    \label{tab:defense_taxanomy}
    \centering
    \caption{Categorization of selected defenses, adopted and and extended from~\citet{gunnemann2022graph,felix2022robust}.}
    \label{tab:categorization}
    \vspace{0.5em}
    \resizebox{0.8\linewidth}{!}{
        \begin{tabular}{lll|ll}
            \toprule \toprule
            \multicolumn{3}{c|}{Taxonomy} & Selected Defenses & Other Defenses \\
            \midrule
            \multirow{2}[2]{*}{\begin{tabular}[c]{@{}l@{}}Improving \\ graph\end{tabular}} & \multicolumn{2}{l|}{Unsupervised} & \begin{tabular}[c]{@{}l@{}}Jaccard-GCN \citep{huijun2019jaccardgcn} 
            \end{tabular} & \begin{tabular}[c]{@{}l@{}}  \citep{negin2020gcnsvd,dongsheng2020aane} \\ \citep{yang2021aLightweight,zhang2019comparing,yingxue2021detection} \end{tabular}  \\
            
            \cmidrule{2-5}
            & \multicolumn{2}{l|}{Supervised} & \begin{tabular}[c]{@{}l@{}}  GARNET \citep{chenhui2022garnet}  
            \end{tabular}& \begin{tabular}[c]{@{}l@{}} \citep{wei2020prognn,hui2021speedup,shuchang2021adversarial} \\ \citep{baolian2022graph,wei2023GTrans} \end{tabular}  \\
            \midrule
            
            \multicolumn{3}{l|}{Improving training} & \begin{tabular}[c]{@{}l@{}}  GRAND~\citep{wenzheng2020grand} \\ GCORN ~\citep{AbbahaddouELVB24} \\ {NoisyGNN}~\citep{sofiane2024a} \end{tabular}  & \begin{tabular}[c]{@{}l@{}} \citep{wenzheng2020grand,pan2020variational,ming2019power} \\ \citep{yuning2020graph,cheng2020robust,jun2022defending} \\ \citep{jinyin2021smoothing,zhijie2023batch,fuli2021graph} \\ \citep{weibo2021robust,jiaron2022unsupervised,kaidi2019pgd} \end{tabular}  \\
            
            \midrule
            \multirow{5}[5]{*}{\begin{tabular}[c]{@{}l@{}}Improving \\ architecture\end{tabular}} & \multirow{5}[5]{*}{\begin{tabular}[c]{@{}l@{}}Adaptively \\ weighting \\ edges\end{tabular}} & Rule-based & GNNGuard \citep{xiang2020gnnguard}  & \citep{wei2021node,xiarui2021graph,li2020afeature}  \\
            \cmidrule{3-5}
            & & Probabilistic & RobustGCN \citep{dingyuan2019robust}  & \begin{tabular}[c]{@{}l@{}}\citep{lingwei2020enhancing,vassilis2019edge,bouyan2021uag} \\ \cite{vassilis2020tensor,dongseng2021learning} \end{tabular} \\
            \cmidrule{3-5}
            & & Robust agg. & \begin{tabular}[c]{@{}l@{}}   ElasticGNN \citep{xiaorui2021elastic} \\ RUNG ~\citep{HouFDL24} \end{tabular} & \citep{liang2021understanding,simon2021robustness} \\

            \bottomrule \bottomrule
        \end{tabular}
    }
\end{table*}

 \textbf{Jaccard-GCN.} Prior to train GCN, adjacency matrix is first preprocessed by pruning edges of pairs of nodes that have Jaccard coefficient of the binarized features $J_{ij} = \frac{\mathbf{X}_i \mathbf{X}_j}{min\{\mathbf{X}_i + \mathbf{X}_j,1\}}$ exceed a threshold $\epsilon$.

 \textbf{GARNET. } Similar to Jaccard-GCN, GARNET is a spectral method that preprocesses data to enhance the robustness of GNN models. GARNET first leverages weighted spectral embedding to construct a base graph that utilizes the top few dominant singular components of $\mathbf{A}$ to restore its important graph spectrum. This is not only robust to adversarial attacks but also preserves important structural information for GNNs training. Secondly, GARNEET refines the base graph by pruning uncritical edges according to a probabilistic graphical model.

 \textbf{GRAND. } GRAND begins by introducing a random propagation strategy to augment the graph data. Following that, GRAND leverages consistency regularization to optimize the prediction consistency of unlabeled nodes across different data augmentations.

 \textbf{GCORN. } A theoretical concept of expected robustness in the context of attributed graphs and an upper bound of the expected robustness are first defined by \citeauthor{AbbahaddouELVB24}. Leveraging the established upper bound on the expected robustness, GCORN utilizes the orthonormalization of weight matrices to control the robustness of GCNs against adversarial attacks.

 \textbf{GNNGuard. } The robustness of GNNGUARD is attributed to two novel components:  the neighbor importance estimation and the layer-wise graph memory. GNNGUARD estimates an importance weight for every edge $e_{uv}$ to quantify how relevant node $u$ is to another node $v$ according to the hypothesis that similar nodes are more likely to interact than dissimilar nodes. Updating edges with low importance weights corresponds to pruning incritical edges, since those edges will likely be ignored by the GNN. However, neighbor importance estimation and edge pruning change the graph structure between adjacent GNN layers, which could potentially destabilize GNN training. To allow for robust estimation of importance weights and smooth evolution of edge pruning, GRAND utilizes layer-wise graph memory at each GNN layer, keeping partial memory of the pruned graph structure from the previous layer.

 \textbf{RobustGCN. } To improve GCN robustness, instead of representing nodes as vectors, RobustGCN utilizes Gaussian distributions as the hidden representations to absorb perturbations. Moreover, it uses a variance-based attention mechanism that assigns weights to the node neighbourhoods according to their variances when performing convolutions.

 \textbf{ElasticGNN. } ElasticGNN utilizes a novel and general message passing scheme that works based on $\mathcal{L}_1$ and $\mathcal{L}_2$-based graph smoothing. This message passing algorithm is not only friendly to back-propagation training but also achieves the desired smoothing properties with a theoretical convergence guarantee. Locally adaptive smoothness makes Elastic GNNs more robust to adversarial attacks on graph structure. This is because the attack tends to connect nodes with different labels, which fuzzes the cluster structure in the graph. But LeasticGNN can tolerate large node differences along these wrong edges, and maintain the smoothness along correct edges.

 \textbf{RUNG.} RUNG improves robustness against adversarial attacks by introducing a robust and unbiased graph signal estimator that mitigates the estimation bias present in traditional $\mathcal{L}_1$-based methods. It employs the Minimax Concave Penalty (MCP) to downweight or prune suspicious edges with large feature differences, reducing the impact of adversarial perturbations. The Quasi-Newton Iteratively Reweighted Least Squares (QN-IRLS) algorithm efficiently solves the non-convex optimization problem, ensuring stable convergence and enabling the model to maintain clean accuracy while defending against attacks.

\section{Importance of node degree as criteria to select target nodes and model selection in adversarial attack evaluation}\label{sec:importance_degree&modelSelection}

\begin{figure}[ht!] 
    \begin{center}
        \subfigure[\cora]{\includegraphics[width=0.3\textwidth]{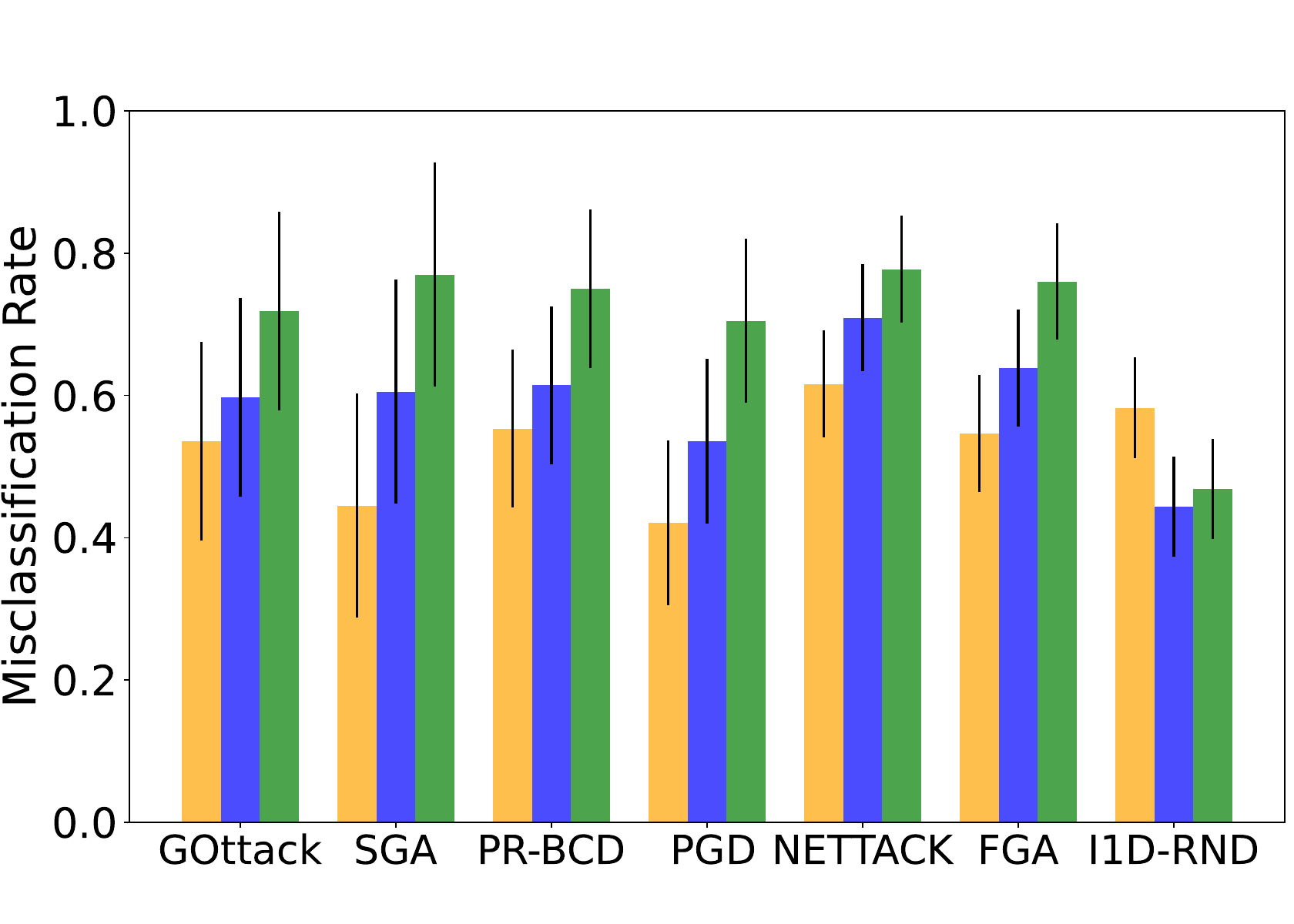}
}
        \subfigure[\citeseer]{\includegraphics[width=0.275\textwidth]{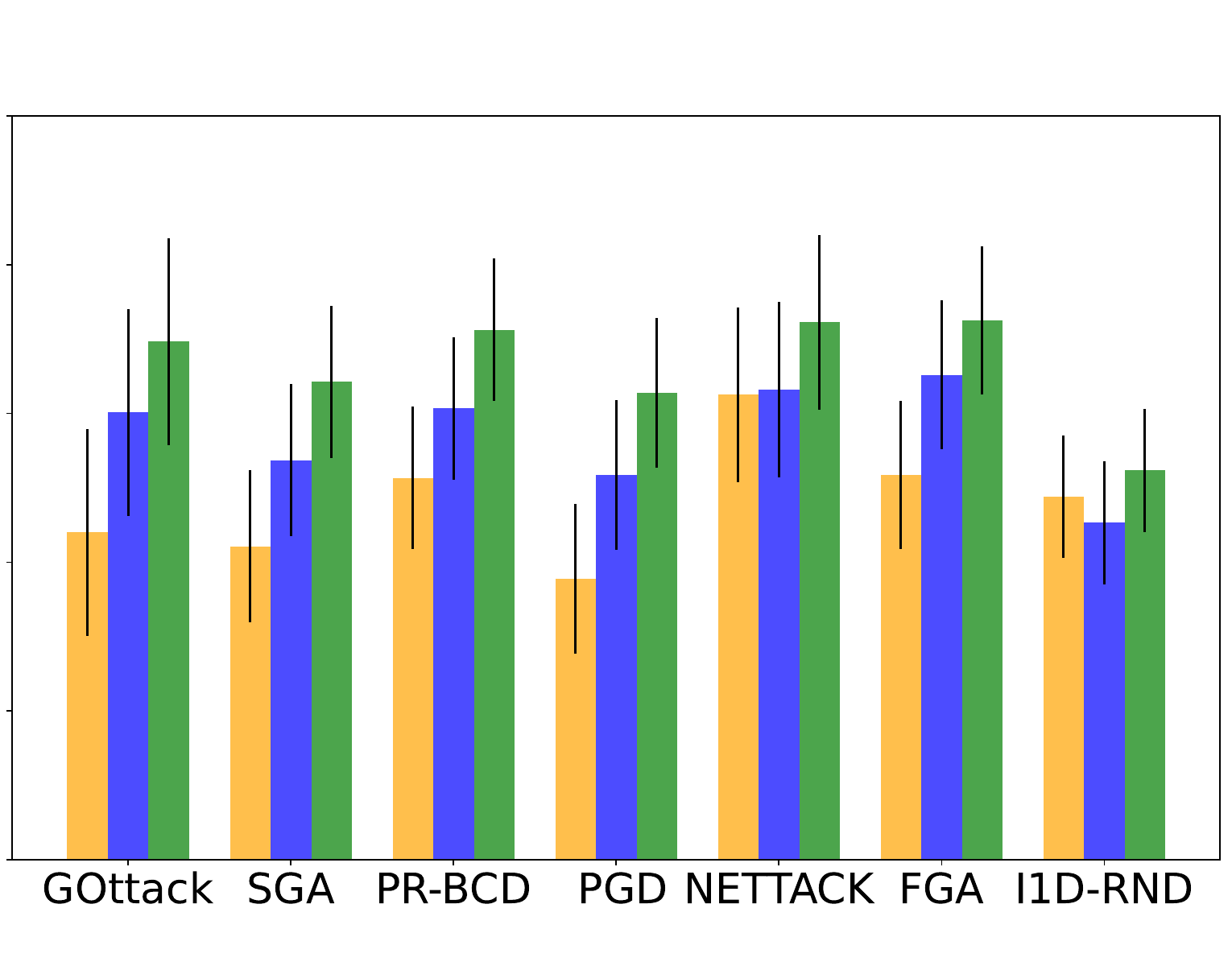}}
        \subfigure[\pubmed]{\includegraphics[width=0.405\textwidth]{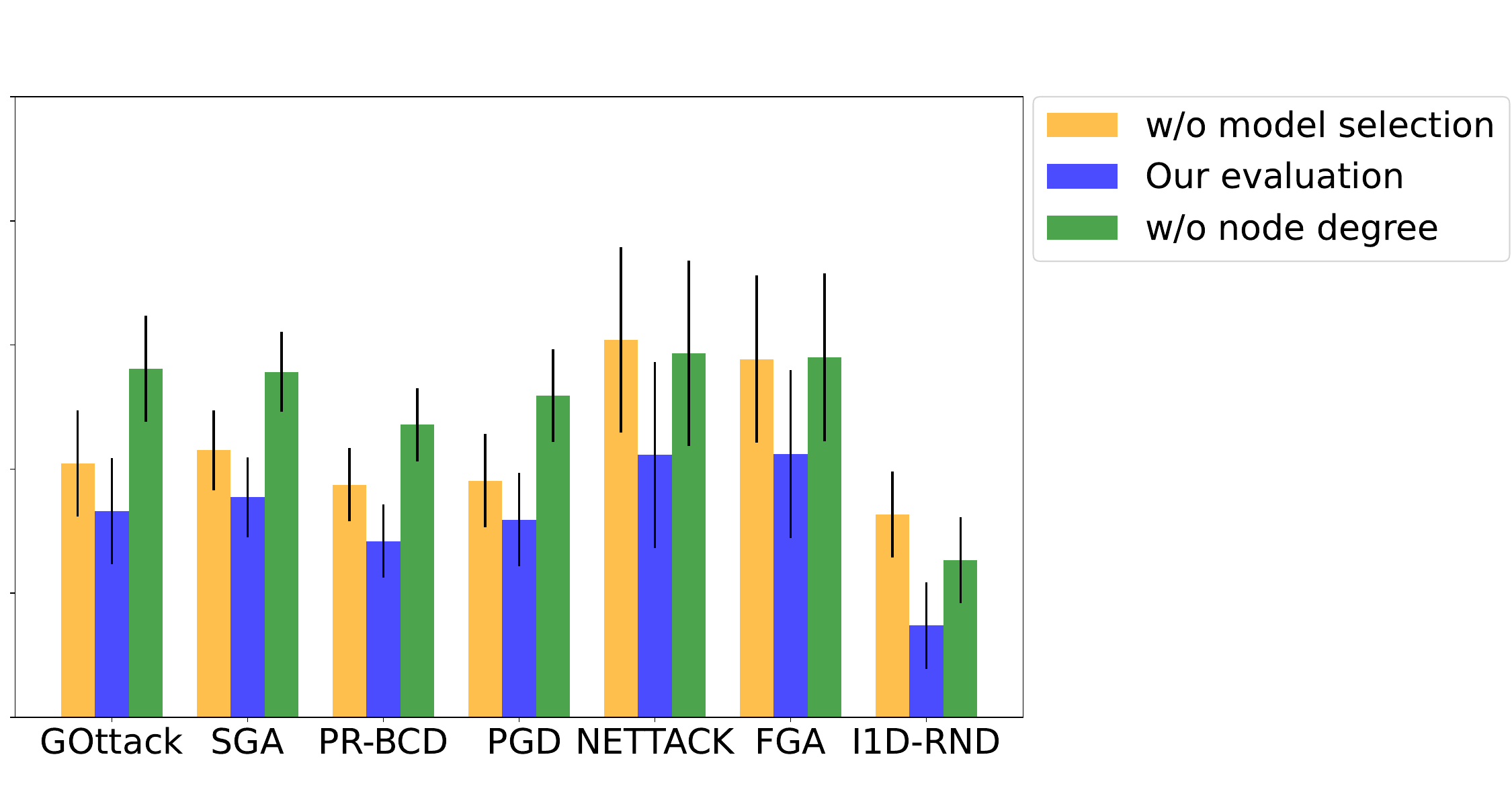}\label{fig:all_difference_pubmed}}

        \caption{Performance of seven adversarial attacks on GSAGE on \cora datasets in poison setting. The figures show the comparison of adversarial attacks' performance obtained from our proposed evaluation procedure, our evaluation without model selection and our evaluation without node degree as a criteria to select target nodes.  }
        \label{fig:all_difference}
    \end{center}
\end{figure}

In this section, we highlight the importance of node degree and model selection in evaluating adversarial attacks. Table~\ref{tab:missclass-table-model-selection-impact} provides a comprehensive comparison of the differences in performance of adversarial attacks when evaluating with and without model selection, while Table~\ref{tab:missclass-table-degree-impact} shows complete differences caused by considering degree as target node selection. A high level summary of Table~\ref{tab:missclass-table-model-selection-impact} and Table~\ref{tab:missclass-table-degree-impact} is provided by Figure~\ref{fig:all_difference}. In general, high-degree nodes are more challenging to attack by existing adversarial methods, resulting in significant increases in the performance of all adversarial attacks when we exclude high-degree nodes in evaluation, especially on the \pubmed dataset (Figure~\ref{fig:all_difference_pubmed}). In addition, Figure~\ref{fig:all_difference} also indicates that evaluating adversarial attacks without performing model selections on victim and surrogate models, which is typically done in practice, results in differences in adversarial attack performance.

Table~\ref{tab:degree-misclassification} highlights a clear dependence of adversarial vulnerability on node degree. For both Cora and Citeseer, nodes with lower degrees exhibit markedly higher average misclassification rates under the top three attacks, confirming that sparsely connected nodes are the most susceptible to perturbations. In Cora, degree-1 nodes reach nearly 0.9 average misclassification, while nodes of degree 14 or 15 show negligible rates. A similar pattern holds for Citeseer, where the rate drops from 0.84 at degree 1 to almost zero by degree 14. Interestingly, intermediate degrees (such as 10–12) show fluctuations rather than a strictly monotonic decline, suggesting that local structural context and model bias interact with degree in nontrivial ways. Overall, these results reinforce the intuition that robustness is strongly linked to connectivity, with low-degree nodes remaining a persistent weak point across models and datasets.

\begin{table}[t]
\centering
\caption{Top 15 node degrees and their average misclassification rates on vanilla models (GCN, GIN, GraphSAGE, PNA) under top-3 adversarial attacks (Nettack, FGA, PRBCD) in the evasion setting.}
\label{tab:degree-misclassification}
\begin{tabular}{rcc}
\toprule
\textbf{Node Degree} & \textbf{Cora (Avg. Misclassification \%)} & \textbf{Citeseer (Avg. Misclassification \%)} \\
\midrule
1  & 0.88$\pm$0.13 & 0.84$\pm$0.13 \\
2  & 0.84$\pm$0.15 & 0.83$\pm$0.09 \\
3  & 0.79$\pm$0.12 & 0.71$\pm$0.14 \\
4  & 0.65$\pm$0.20 & 0.60$\pm$0.08 \\
5  & 0.61$\pm$0.16 & 0.47$\pm$0.15 \\
6  & 0.58$\pm$0.19 & 0.51$\pm$0.11 \\
7  & 0.50$\pm$0.28 & 0.31$\pm$0.14 \\
8  & 0.50$\pm$0.14 & 0.26$\pm$0.18 \\
9  & 0.29$\pm$0.27 & 0.12$\pm$0.14 \\
10 & 0.31$\pm$0.29 & 0.35$\pm$0.30 \\
11 & 0.42$\pm$0.47 & 0.19$\pm$0.30 \\
12 & 0.47$\pm$0.41 & 0.19$\pm$0.21 \\
13 & 0.33$\pm$0.44 & 0.08$\pm$0.29 \\
14 & 0.06$\pm$0.16 & 0.00$\pm$0.00 \\
15 & 0.18$\pm$0.24 & 0.01$\pm$0.04 \\
\bottomrule
\end{tabular}
\end{table}

\textbf{Model Selection.} Model selection is an important step performed prior to deploying GNNs in real-world application. Table~\ref{tab:missclass-table-model-selection-impact} indicates that failing to perform model selection in adversarial attacks could prevent adversarial GNN practitioners from exploring the impact of them in practice. The difference between evaluating adversarial attacks with and without model selections ranges from $-16.40\%$ to $19.07\%$ with $5.19\%$ average absolute difference across adversarial attacks. Performing model selection in evaluation could potentially change the average rank of adversarial attacks. Especially in the setting of GIN and GSAGE, in both evasion and poison settings, the performance of \propseRND attacks significantly drops by $9.15\%$ on average; thus increasing the average rank of \propseRND by $2.74$ on average.

\textbf{Node degree.} Table~\ref{tab:missclass-table-degree-impact} shows that considering node degree as a target node selection criterion instead of randomly selecting results in a significant drop in adversarial attack performance in most of the cases, with $9.80\%$ average decrease. Especially in the setting of GCN on \pubmed dataset on budget 3 in both settings, the difference in \prbcdNA performance between evaluating with and without degree as target node selection criteria is notably significant, $26.3\%$ in evasion and $25.97\%$. Furthermore, the difference of adversarial attack performances in the \pubmed dataset is $-14.969\%$ on average, while that of those in \cora and \citeseer is $-6.93\%$ and $-7.51\%$ on average, respectively.

\section{\propseRND: A simple yet effective baseline for adversarial attacks} \label{sec:rndv2}
This section provides further information about the naive baseline \propseRND. The algorithm and intuition behind \propseRND are discussed in Section~\ref{sec:rndv2_alg}, while Section~\ref{sec:rndv2_vs_rnd} highlights the improvement of \propseRND over random attack (RND).

\textbf{Attack strategy and motivation.}\label{sec:rndv2_alg} \propseRND attack follow the strategy described in Algorithm~\ref{alg:l1d-rnd}. Given a target node $v_t$, the adjacency matrix $\mathbf{A}$, and node features $\mathbf{X}$, \propseRND randomly decides to add or to remove an edge (Line~\ref{alg_line:rand_add_remove}). In the case of adding an edge, \propseRND first randomly sampled a set of vertices $\mathcal{V}_{\text{candidate}}'$ that are not neighbours of the target node $v_t$ (Line~\ref{alg_line:select_add_candidate}). \propseRND then adds an edge between the target node $v_t$ and $u$, where $u$ is the node with the highest degree in $\mathcal{V}_{\text{candidate}}'$ (Line~\ref{alg_line:select_highest_degree}). We select the highest node from the node's sample $\mathcal{V}_{\text{candidate}}'$ to avoid always selecting a node with the highest degree from the whole graph. The adversarial impact of edges added by \propseRND is attributed to two features:  stochasticity and node degree. In particular, sampling a subset of nodes enables stochastic noise to be introduced to the target node $v_t$, and selecting nodes with the highest degree enables the node that can aggregate the most messages from its neighbours. In the case of removing an edge, \propseRND considers removing an edge between target node $v_t$ and its neighbours, excluding new neighbours added by \propseRND (Line~\ref{alg_line:filter_neigbour}). Similar to the case of adding an edge, \propseRND then sampled a set of vertices $\mathcal{V}_{\text{candidate}}'$ from its neighbours. Finally, \propseRND removes an edge between target node $v_t$ and $u$, where $u$ has the highest influence score that is computed in Algorithm~\ref{alg:s_influence}. For each node, $u$, the influence score is defined by computing the $\mathcal{L}_1$ norm of $u$ and its neighbour node features. This influence score provides an approximation of the contribution of messages of node $u$ to the representation of $u$'s neighbour in the message aggregation stage.

\begin{algorithm}[tb]
   \caption{ \propseRND ($v_t$, $\Delta$, $\mathcal{G}$, $r$)}
   \label{alg:l1d-rnd}
\begin{algorithmic}[1]
   \STATE {\bfseries Input:} Target node $v_t$, modified budget $\Delta$, \\
   Graph $\mathcal{G} = (\mathbf{A}, \mathbf{X})$, sample ratio $r$
   \STATE {\bfseries Output:} Modified graph $\mathcal{G'} = (\mathbf{A}', \mathbf{X})$
   \STATE {$\mathbf{A}' = \mathbf{A}$}
   \STATE $\mathcal{N}_{v_t}^{\text{new}} = [ ]$
   \WHILE{$|\mathbf{A} - \mathbf{A}'| < \Delta $}
   \STATE $\mathcal{N}_{v_t} =$ neighbor($v_t$, $\mathbf{A}'$)
   \STATE $x \gets \text{rand()}$ \cmt{Random number between 0 and 1} \label{alg_line:rand_add_remove}
   \IF{$x > 0.5$} 
   \STATE \cmt{Add an edge}
   \STATE $\mathcal{V}_{\text{candidate}}$ = $ [0, 1, 2, \dots, \text{len}(\mathbf{A})] - \mathcal{N}_{v_t}$ \cmt{Exclude neighbours nodes} \label{alg_line:select_add_candidate}
   \STATE $k = \lfloor r \times |\mathcal{V}_{\text{candidate}}| \rfloor$
   \STATE $\mathcal{V}_{\text{candidate}}' \sim \text{Uniform}(\mathcal{V}_{\text{candidate}}, k)$ \label{alg_line:sample_add_candidate}
   \STATE $e^* = (v_t,u) \gets \mathop{\arg\max}\limits_{u \in \mathcal{V}_{\text{candidate}}'} \text{degree}(u, \mathbf{A'})$ \label{alg_line:select_highest_degree}
   \STATE $\mathbf{A'} = \mathbf{A'} + e^*$
   \STATE $\mathcal{N}_{v_t}^{\text{new}}$.append($u$)
   \ELSE 
   \STATE \cmt{Remove an edge}
   \STATE $\mathcal{N}_{v_t}^- = \mathcal{N}_{v_t} - \mathcal{N}_{v_t}^{\text{new}}$ \label{alg_line:filter_neigbour}
   \STATE $k = \lfloor r \times |\mathcal{N}_{v_t}^-| \rfloor$
   \STATE $\mathcal{V}_{\text{candidate}} \sim \text{Uniform}(\mathcal{N}_{v_t}^-, k)$ \label{alg_line:sample_remove_candidate}
   \STATE $e^* = (v_t,u) \gets \mathop{\arg\max}\limits_{u \in \mathcal{V}_{\text{candidate}}} s_{\text{influence}}(u, \mathbf{A'}, \mathbf{X})$ \label{alg_line:select_highest_l1}
   \STATE $\mathbf{A'} = \mathbf{A'} - e^*$   
   
   \ENDIF
  
   \ENDWHILE

\end{algorithmic}
\end{algorithm}

\begin{algorithm}[tb]
   \caption{ $s_{\text{influence}}$ ($u$, $\mathbf{A}$, $\mathbf{X}$)}
   \label{alg:s_influence}
\begin{algorithmic}[1]
   \STATE {\bfseries Input:} Node $u$, Adjacency matrix $\mathbf{A}$, Node feature $\mathbf{X}$
   \STATE {\bfseries Output:} Influence score $s_{\text{influence}}$
   \STATE $\mathcal{N}_u = \text{neighbor}(u, \mathbf{A}) + u$
   
   \STATE $s_{\text{influence}} = L_1 \text{norm} (\mathbf{X}[\mathcal{N}_u,:])$

\end{algorithmic}
\end{algorithm}

\textbf{Improvement over random attack (RND).} \label{sec:rndv2_vs_rnd} This section highlights the improvement of $\mathcal{L}_1$ norm and node degree in \propseRND over random attack (RND). Table~\ref{tab:rnd_vs_rndv2} shows that \propseRND outperforms RND by an average of $5.15\%$  in $84\%$ settings. Especially, \propseRND performs significantly better by an average of $13.4\%$ than RND in all settings on GRAND. 

While many benchmark datasets use sparse, bag-of-words-like node features (e.g., Cora, Citeseer, Pubmed), we test the generality of \propseRND by evaluating it on OGB-Arxiv, a large-scale graph with continuous node embeddings derived from raw text. Even in this setting, \propseRND outperforms PR-BCD (NA) in 8 out of 10 cases on GraphSAGE and remains competitive on GCN (Tables 16–19, Support results).
These results suggest that even basic perturbation strategies can be effective under practical constraints, especially when stronger attacks lose efficacy due to defense adaptations or poor transferability. We advocate including simple baselines like \propseRND in future evaluations to ensure that observed robustness is not overstated by comparisons limited only to complex attacks.

\textbf{Time cost.} Table~\ref{tab:time_record} has shown superior speed of our simple \propseRND attacks compared to iterative or gradient-based methods. 1RND demonstrates remarkable efficiency in adversarial attacks, as evidenced by its consistently low and stable execution times across budgets. With execution times ranging from just $16.90 \pm 0.37$ seconds to $19.23 \pm 0.62$ seconds, \propseRND orders of magnitude faster than remaining gradient-based attacks, especially PGD.

{\textbf{Factors Driving the Effectiveness of \propseRND.} Our hypothesis is that the success of \propseRND attack, especially against defended models, arises from two main factors.}

{First is bias toward high-degree nodes. Even though \propseRND is simple, its edge selection is not purely random. On the addition side, it connects the target node to a high-degree node (line 13 in Alg.3). On the removal side, it prunes based on an $L_1$-norm-based influence measure (Alg.4), which correlates with local feature aggregation. This introduces a structure-aware perturbation mechanism, despite the absence of gradient-based optimization. Our ablation shows that both components, when used independently, are sufficient to create harmful structural noise.}

{The second factor is the mismatch between attack complexity and defense generalization. Existing defenses often assume strong adversarial priors (e.g., feature gradients, link importance). These may fail to generalize against structurally simple, diverse attacks. \propseRND (because it doesn’t follow conventional optimization) creates perturbations that may fall outside the expected adversarial space, allowing it to bypass some defenses.}

{The ablation table supports this: \propseRND-Add Only and  \propseRND-Remove Only consistently outperform the fully randomized baseline and sometimes even the full \propseRND. In particular, Add Only is dominant in many settings, suggesting that the influence of hub-attachment alone can severely distort message-passing dynamics.}

\begin{figure*}[ht!]
    \begin{center}
    
    \subfigure[PNA]{\includegraphics[width=0.39\textwidth]{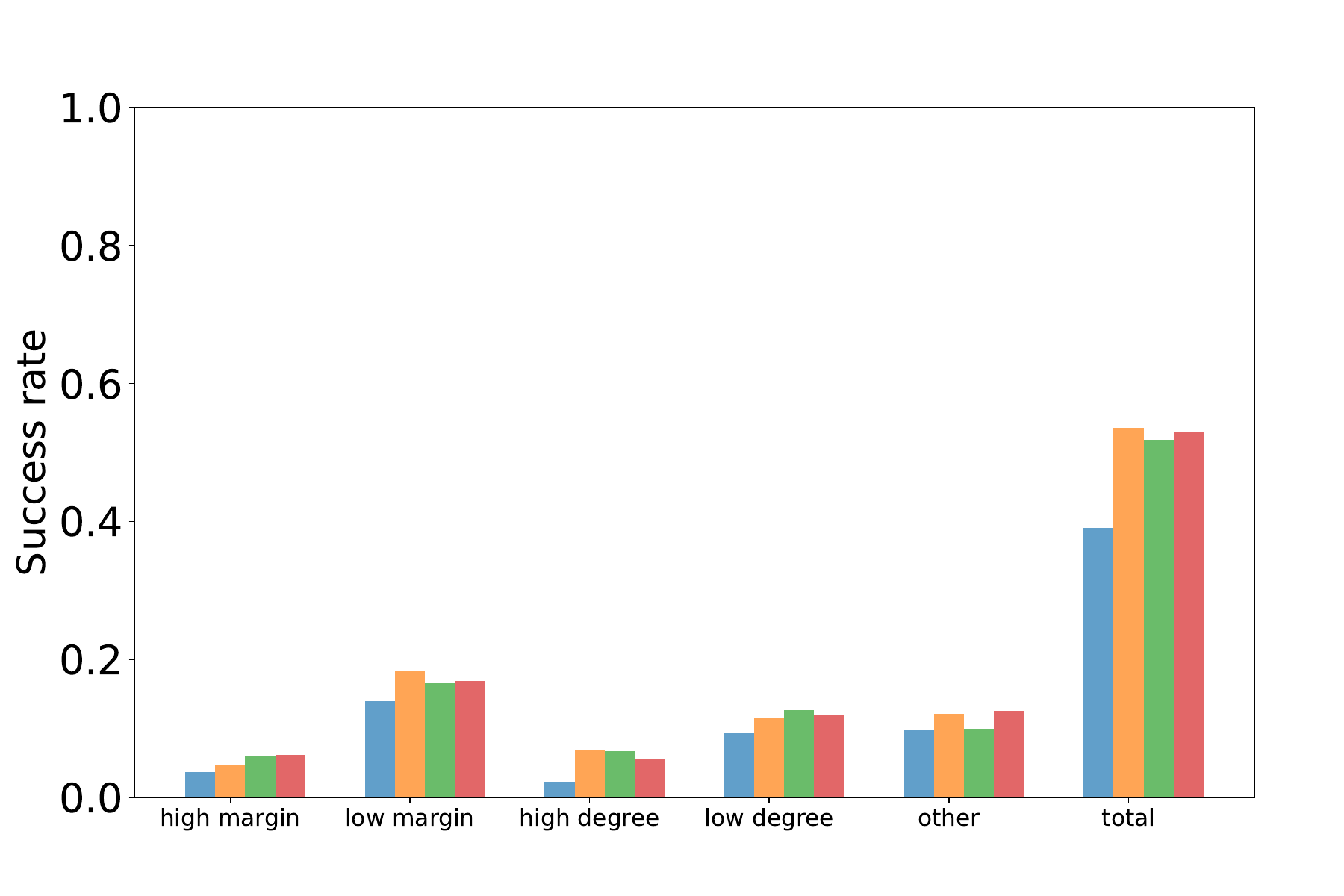}\label{fig:ablation_rnd_cora_PNA_evasion_5}}
    \subfigure[GRAND]{\includegraphics[width=0.55\textwidth]{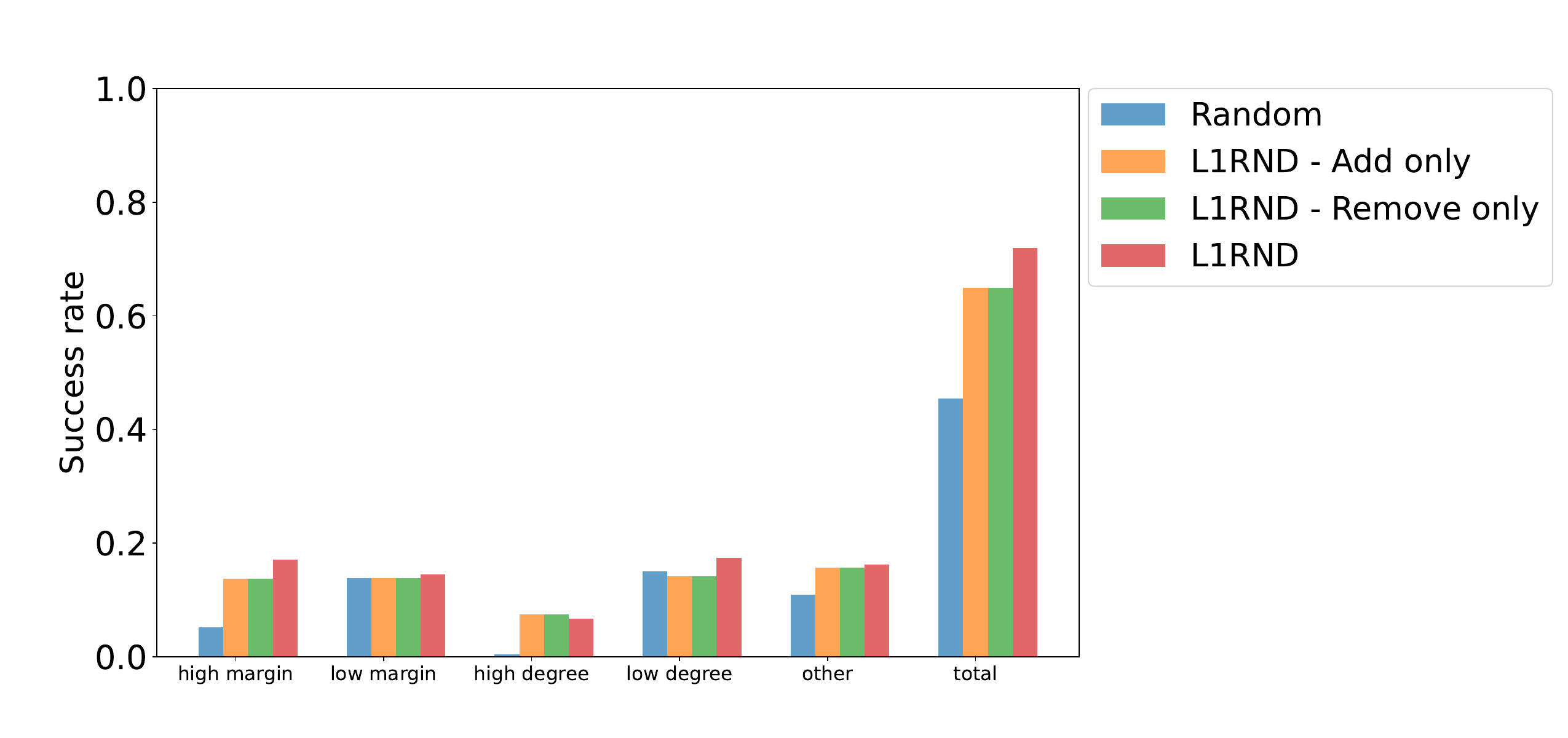}\label{fig:ablation_rnd_cora_GRAND_evasion_5}}
    
    \caption{{Ablation study on \propseRND. The effect of other two variants of \propseRND: \propseRND that add edge only (degree affect) and \propseRND that remove edge only ($L^1$ norm affect) on different class of target nodes: High/low margin, High/low degree and random on budget $5$ of evasion setting.}}
    \end{center}
\vspace{-1.5em}
\end{figure*}

\section{Adversarial Attack Methods} \label{sec:summary_attacks}

In this section, we summarize the state-of-the-art adversarial attack techniques considered in our benchmark. These attacks vary in their use of surrogate models, optimization strategies, and access assumptions, offering a diverse testbed for evaluating GNN robustness.

\begin{table}[H]
\caption{Adversarial attacks and victim models used to evaluate the attack method in the corresponding paper.}
\label{tab:atta_victim}
\vskip 0.15in
\begin{center}
\begin{small}
\begin{sc}
\begin{tabular}{@{}l@{\hspace{8pt}}p{10cm}@{}}
\toprule \toprule
Adversarial attacks & Victim models \\
\midrule
NETTACK     & GCN, DeepWalk, CLN \\ 

FGA         & GCN, GraRep, GraphGAN, DeepWalk, Node2vec, LINE \\ 

PGD         & GCN \\ 
PR-BCD      & GCN, Soft Median GDC, Soft Median PPRGo, Soft Medoid GDC, GDC, PPRGo, RGCN, GCN-Jaccard, GCN-SVD \\ 
SGA         & GCN, SGC, GAT, GraphSAGE, Cluster-GCN \\ 
GOttack     & GCN, GIN, GraphSAGE, RGCN, GCN-Jaccard, GCN-SVD, MedianGCN \\ \hline
\bottomrule
\end{tabular}
\end{sc}
\end{small}
\end{center}
\vskip -0.1in
\end{table}

\textbf{Nettack.} One of the earliest comprehensive attack frameworks on graph data. Nettack perturbs both graph structure and node features using surrogate gradient information from a trained GCN. Specifically, it applies discrete combinatorial search to structure perturbations, while using gradient-based optimization for feature perturbations. Nettack operates in a targeted, transfer-based setting and has been widely adopted as a baseline for local attacks.

\textbf{FGA.} The Fast Gradient Attack generates adversarial examples by computing gradients with respect to the adjacency matrix and feature matrix of a GCN surrogate. Although originally proposed for unsupervised embeddings (e.g., DeepWalk, LINE), it is also used in GNN contexts. FGA operates in a white-box setting with access to model gradients.

\textbf{PGD.} Projected Gradient Descent (PGD) attacks optimize graph perturbations using first-order methods and have been proposed both for attack and for adversarial training. In our benchmark, we use PGD in a transfer setting via a GCN surrogate. However, it is important to note that PGD can also be deployed adaptively in white-box settings, where its full strength is realized by directly attacking the defense model. Adaptive variants are left for future expansion of our benchmark.

\textbf{PR-BCD.} This scalable attack framework employs a block coordinate descent (BCD) strategy to optimize perturbations efficiently. It supports both targeted and global settings and introduces custom surrogate losses. In contrast to prior work that evaluated PR-BCD in a transfer setting, we evaluate it both as a transfer-based attack using a fixed GCN surrogate, and in an adaptive, white-box setting where it directly attacks the defended model. This inclusion addresses recent calls for adaptive robustness evaluation.

\begin{wraptable}{r}{0.4\textwidth}
\caption{Adversarial attacks and corresponding surrogate models.}
\label{tab:atta_surro}
\vskip 0.15in
\begin{center}
\begin{small}
\begin{sc}
\resizebox{0.35\textwidth}{!}{\begin{tabular}{@{}l@{\hspace{8pt}}c@{}}
\toprule \toprule
Adversarial attacks & Surrogate model \\
\midrule
NETTACK   & GCN     \\
FGA       & GCN     \\
PGD       & GCN     \\
PR-BCD    & GCN   \\
SGA       & SGC     \\
GOttack   & GCN     \\
\bottomrule \bottomrule
\end{tabular}
}
\end{sc}
\end{small}
\end{center}
\vskip -0.1in
\end{wraptable}

\textbf{SGA.} The Simplified Gradient-based Attack reduces computational overhead by restricting optimization to smaller subgraphs. It uses SGC (a simplified GCN variant) as its surrogate and targets structural vulnerabilities efficiently. While not adaptive to defenses, SGA offers a lightweight attack option for benchmarking.

\textbf{GOttack.} GOttack introduces a topological approach to adversarial attacks using graph orbits. Instead of relying on gradients, it learns structural patterns to perturb, making it model-agnostic and efficient. GOttack demonstrates strong performance in transfer-based settings, especially on graph datasets where symmetry-based features are predictive.

Table \ref{tab:atta_victim} summarizes the original victim models used in each attack’s evaluation. Table \ref{tab:atta_surro} lists the surrogate models employed during attack generation. While many attacks rely on GCNs as surrogates, we highlight the distinction between transfer-based attacks (e.g., Nettack, FGA, GOttack) and our inclusion of adaptive PR-BCD as a white-box method directly targeting defended models. This allows our benchmark to more realistically assess model robustness under both practical and worst-case threat assumptions.

\newpage

\begin{landscape}
\begin{table}[ht]
    \centering
    \caption{Hyper-parameters used for model selection.}
    \small
    \renewcommand\arraystretch{1}
    \setlength{\tabcolsep}{13pt}

    \label{tab:Hyper-parameters-table}
        \resizebox{\linewidth}{!}{ 

    }
    \end{table}
\end{landscape}

\begin{landscape}
    \begin{table}[ht]
    \centering
    \caption{\textbf{Vanilla Homophily Results.} Evaluating six adversarial attacks on four vanilla attack victim models - Miss-classification rate (\%). Higher is better.  Best performance in \textbf{bold}, second best \underline{underlined}.}
    \small
    \renewcommand\arraystretch{1.2}
    \setlength{\tabcolsep}{3pt}
    \label{tab:eval_attack_non_defense_homo}
        \resizebox{\linewidth}{!}{
                \begin{tabular}{ c | cc | ccccc | ccccc | ccccc | c }
                \toprule \toprule
                \multicolumn{3}{c}{Dataset} & \multicolumn{5}{c}{\cora} & \multicolumn{5}{c}{\citeseer} & \multicolumn{5}{c}{\pubmed} \\
                \hline
                && $\Delta \rightarrow$ & 1 & 2 & 3 & 4 & 5 & 1 & 2 & 3 & 4 & 5 & 1 & 2 & 3 & 4 & 5 & Avg. Rank\\
                \cline{2-19}

                        \multirow{28}{*}{\rotatebox{90}{\textbf{Evasion}}}&\multirow{7}{*}{\rotatebox{90}{\textbf{GCN}}}& \propseRND & 13.20 $\pm$ 4.33 & 22.53 $\pm$ 6.95 & 29.20 $\pm$ 5.60 & 32.93 $\pm$ 6.67 & 32.13 $\pm$ 5.04 & 15.20 $\pm$ 4.20 & 28.67 $\pm$ 12.39 & 36.53 $\pm$ 9.09 & 37.73 $\pm$ 9.44 & 42.13 $\pm$ 7.11 & 10.93 $\pm$ 3.01 & 15.47 $\pm$ 4.87 & 17.20 $\pm$ 6.58 & 18.80 $\pm$ 5.28 & 23.60 $\pm$ 9.63& 7.00\\
                        && FGA & 27.87 $\pm$ 3.74 & 46.13 $\pm$ 7.03 & 55.73 $\pm$ 4.33 & 60.93 $\pm$ 5.06 &\underline{64.67 $\pm$ 4.76} & 25.47 $\pm$ 4.10 & 36.00 $\pm$ 8.18 & 50.27 $\pm$ 5.90 & 56.53 $\pm$ 2.97 & 58.27 $\pm$ 4.27 &\underline{35.60 $\pm$ 3.14} & 47.60 $\pm$ 5.77 & 55.33 $\pm$ 4.12 &\underline{58.40 $\pm$ 2.29} & 58.00 $\pm$ 3.02& 3.40\\
                        && NETTACK &\underline{29.60 $\pm$ 5.19} &\underline{49.60 $\pm$ 5.77} &\underline{56.80 $\pm$ 4.52} &\underline{62.13 $\pm$ 4.17} & 63.73 $\pm$ 3.99 &\underline{28.13 $\pm$ 7.15} &\underline{45.33 $\pm$ 9.06} & 49.87 $\pm$ 9.61 &\underline{58.93 $\pm$ 5.85} &\underline{59.73 $\pm$ 3.92} & 33.73 $\pm$ 3.37 &\textbf{55.73 $\pm$ 4.33} &\textbf{57.47 $\pm$ 2.20} &\textbf{58.67 $\pm$ 2.09} &\underline{58.67 $\pm$ 2.23}& 2.20\\
                        &&  PGD & 29.33 $\pm$ 4.12 & 48.00 $\pm$ 5.24 & 54.93 $\pm$ 3.28 & 59.60 $\pm$ 4.36 & 59.20 $\pm$ 5.49 & 25.47 $\pm$ 4.87 & 40.27 $\pm$ 7.55 & 48.13 $\pm$ 7.23 & 52.27 $\pm$ 7.28 & 55.47 $\pm$ 5.15 & 34.13 $\pm$ 3.66 & 40.67 $\pm$ 3.98 & 40.93 $\pm$ 5.18 & 44.80 $\pm$ 5.75 & 44.93 $\pm$ 6.27& 4.93\\
\bottomrule \bottomrule
    \end{tabular}
    }
    \end{table}

\end{landscape}

 \begin{table}[ht]
\begin{minipage}{0.48\textwidth}  
\centering
      \caption{\textbf{Vanilla Heterophily Results - 1/2.} Evaluating six adversarial attacks on four vanilla attack victim models on \squirrel - Miss-classification rate (\%). Higher is better.  Best performance in \textbf{bold}, second best \underline{underlined}. GIN and PNA are excluded from this table due to OOR.}
    \label{tab:non_defense_heterophily_p1}
  \setlength{\tabcolsep}{3pt}
  \renewcommand{\arraystretch}{1.5}
   \resizebox{\linewidth}{!}{ 
                \begin{tabular}{ c | cc | ccccc | c }
                \toprule
                \multicolumn{3}{c}{Dataset} & \multicolumn{5}{c}{\squirrel} \\
                \hline
                && $\Delta \rightarrow$ & 1 & 2 & 3 & 4 & 5 & Avg. Rank\\
                \cline{2-8}

                         \multirow{14}{*}{\rotatebox{90}{\textbf{Evasion}}}&\multirow{7}{*}{\rotatebox{90}{\textbf{GCN}}}& \propseRND & 24.93 $\pm$ 33.09 & 64.53 $\pm$ 6.91 &\textbf{72.80 $\pm$ 7.36} &\underline{71.60 $\pm$ 5.67} & 71.47 $\pm$ 4.56& 2.80\\
                        && FGA &\underline{62.40 $\pm$ 14.64} &\underline{65.87 $\pm$ 15.45} &\underline{71.07 $\pm$ 7.40} & 71.07 $\pm$ 5.80 &\underline{71.60 $\pm$ 4.36}& 2.20\\
                        && NETTACK & 1.87 $\pm$ 3.34 & 1.47 $\pm$ 2.77 & 1.33 $\pm$ 1.63 & 1.60 $\pm$ 2.29 & 2.13 $\pm$ 2.20& 7.00\\
                        &&  PGD & 47.73 $\pm$ 10.25 & 58.93 $\pm$ 12.28 & 65.87 $\pm$ 10.04 & 62.80 $\pm$ 11.73 & 64.80 $\pm$ 13.54& 4.60\\
                        && \prbcdNA &\textbf{69.87 $\pm$ 10.76} &\textbf{69.73 $\pm$ 6.41} & 69.60 $\pm$ 6.56 &\textbf{71.87 $\pm$ 4.63} &\textbf{73.07 $\pm$ 5.12}& 1.40\\
                        && SGA & 52.00 $\pm$ 6.19 & 59.20 $\pm$ 4.46 & 60.53 $\pm$ 8.83 & 65.07 $\pm$ 8.24 & 66.80 $\pm$ 7.04& 4.00\\
\hline
    \bottomrule
    \end{tabular}
    }
\end{minipage}
\hfill
\begin{minipage}{0.48\textwidth}
  \centering
    \caption{\textbf{Vanilla Heterophily Results - 2/2.} Evaluating six adversarial attacks on four vanilla attack victim models on \chameleon - Miss-classification rate (\%). Higher is better.  Best performance in \textbf{bold}, second best \underline{underlined}. PNA is excluded from this table due to OOR.}
    \label{tab:non_defense_heterophily_p2}
  \setlength{\tabcolsep}{3pt}
  \renewcommand{\arraystretch}{1.5}
   \resizebox{\linewidth}{!}{ 
                \begin{tabular}{ c | cc | ccccc | c }
                \toprule
                \multicolumn{3}{c}{Dataset} & \multicolumn{5}{c}{\chameleon} \\
                \hline
                && $\Delta \rightarrow$ & 1 & 2 & 3 & 4 & 5 & Avg. Rank\\
                \cline{2-8}

                        \multirow{21}{*}{\rotatebox{90}{\textbf{Evasion}}}&\multirow{7}{*}{\rotatebox{90}{\textbf{GCN}}}& \propseRND & 21.87 $\pm$ 28.89 & 50.93 $\pm$ 9.71 &\underline{62.00 $\pm$ 6.05} &\underline{66.67 $\pm$ 7.12} &\underline{66.53 $\pm$ 7.07}& 3.00\\
                        && FGA &\textbf{62.40 $\pm$ 7.72} &\underline{58.00 $\pm$ 12.72} & 59.87 $\pm$ 15.97 & 56.40 $\pm$ 12.54 & 61.07 $\pm$ 12.30& 3.00\\
                        && NETTACK & 3.07 $\pm$ 2.60 & 5.47 $\pm$ 3.96 & 4.93 $\pm$ 4.33 & 6.27 $\pm$ 4.20 & 6.27 $\pm$ 4.83& 7.00\\
                        &&  PGD & 44.00 $\pm$ 20.95 & 50.53 $\pm$ 18.10 & 61.07 $\pm$ 14.64 & 57.73 $\pm$ 15.47 & 59.47 $\pm$ 14.05& 3.60\\
                        && \prbcdNA &\underline{58.00 $\pm$ 18.53} &\textbf{67.87 $\pm$ 20.46} &\textbf{70.27 $\pm$ 8.88} &\textbf{67.07 $\pm$ 10.66} &\textbf{73.47 $\pm$ 8.77}& 1.20\\
                        && SGA & 45.47 $\pm$ 10.38 & 49.07 $\pm$ 9.82 & 55.60 $\pm$ 7.64 & 57.33 $\pm$ 9.37 & 55.33 $\pm$ 6.70& 4.40\\
\bottomrule \bottomrule
    \end{tabular}
    }
\end{minipage}

\end{table}

 \begin{landscape}

\begin{table}[ht]
    \centering
    \caption{\textbf{Defended Homophily Results - {1/3}.} Evaluating six adversarial attacks on four defense victim models (GNNGuard, GCN-Jaccard, GCN-GARNET ElasticGNN) - Miss-classification rate (\%). Higher is better.  Best performance in \textbf{bold}, second best \underline{underlined}.}
    \small
    \renewcommand\arraystretch{1.2}
    \setlength{\tabcolsep}{3pt}
    \label{tab:eval_defense_homo_p1}
        \resizebox{\linewidth}{!}{
                \begin{tabular}{ c | cc | ccccc | ccccc | ccccc | c }
                \toprule \toprule
                \multicolumn{3}{c}{Dataset} & \multicolumn{5}{c}{\cora} & \multicolumn{5}{c}{\citeseer} & \multicolumn{5}{c}{\pubmed} \\
                \hline
                && $\Delta \rightarrow$ & 1 & 2 & 3 & 4 & 5 & 1 & 2 & 3 & 4 & 5 & 1 & 2 & 3 & 4 & 5 & Avg. Rank\\
                \cline{2-19}

                        \multirow{28}{*}{\rotatebox{90}{\textbf{Evasion}}}&\multirow{7}{*}{\rotatebox{90}{\textbf{ElasticGNN}}}& \propseRND & 12.67 $\pm$ 2.47 & 14.80 $\pm$ 4.13 & 19.33 $\pm$ 3.83 & 20.93 $\pm$ 5.55 & 24.40 $\pm$ 5.36 & 14.00 $\pm$ 5.40 & 19.07 $\pm$ 5.50 & 24.53 $\pm$ 8.80 & 27.33 $\pm$ 9.22 & 32.27 $\pm$ 10.14 & 10.67 $\pm$ 3.75 & 10.67 $\pm$ 4.32 & 12.27 $\pm$ 4.71 & 13.87 $\pm$ 3.96 & 13.73 $\pm$ 4.33& 7.00\\
                        && FGA &\textbf{24.67 $\pm$ 3.83} &\textbf{32.40 $\pm$ 4.91} & 37.73 $\pm$ 6.96 & 44.00 $\pm$ 8.55 &\underline{50.93 $\pm$ 6.63} & 22.00 $\pm$ 3.30 & 30.13 $\pm$ 6.16 &\underline{41.20 $\pm$ 9.70} &\textbf{46.80 $\pm$ 12.60} &\underline{51.07 $\pm$ 12.02} &\underline{24.67 $\pm$ 6.87} & 26.27 $\pm$ 5.44 &\textbf{30.80 $\pm$ 7.85} &\underline{33.47 $\pm$ 8.50} &\textbf{35.73 $\pm$ 9.13}& 2.60\\      
                        && NETTACK & 23.47 $\pm$ 3.16 & 30.27 $\pm$ 5.99 &\textbf{43.20 $\pm$ 5.60} &\textbf{49.87 $\pm$ 6.07} &\textbf{54.80 $\pm$ 5.17} & 22.80 $\pm$ 4.26 &\textbf{33.87 $\pm$ 9.46} &\textbf{42.80 $\pm$ 11.85} &\underline{46.67 $\pm$ 12.71} & 49.87 $\pm$ 11.82 & 23.87 $\pm$ 5.63 & 26.00 $\pm$ 6.28 & 29.87 $\pm$ 7.84 &\textbf{33.73 $\pm$ 9.94} &\underline{34.67 $\pm$ 10.13}& 2.87\\
                        &&  PGD & 24.00 $\pm$ 4.34 & 30.40 $\pm$ 2.75 & 38.67 $\pm$ 6.44 & 44.00 $\pm$ 7.56 & 47.33 $\pm$ 8.09 &\underline{22.93 $\pm$ 4.65} & 30.27 $\pm$ 7.09 & 37.60 $\pm$ 9.01 & 42.53 $\pm$ 9.46 & 45.60 $\pm$ 9.75 &\textbf{24.80 $\pm$ 5.89} &\underline{28.00 $\pm$ 5.95} & 30.00 $\pm$ 5.71 & 30.53 $\pm$ 5.26 & 32.27 $\pm$ 5.18& 4.07\\
\bottomrule \bottomrule
    \end{tabular}
    }
    \end{table}

    \begin{table}[ht]
    \centering
    \caption{\textbf{Defended Homophily Results - {2/3}.} Evaluating six adversarial attacks on four defense victim models (GRAND, RobustGCN, GCORN, RUNG) - Miss-classification rate (\%). Higher is better.  Best performance in \textbf{bold}, second best \underline{underlined}. OOR entries are omitted in average rank calculation. }
    \small
    \renewcommand\arraystretch{1.2}
    \setlength{\tabcolsep}{3pt}
    \label{tab:eval_defense_homo_p2}
        \resizebox{\linewidth}{!}{
                \begin{tabular}{ c | cc | ccccc | ccccc | ccccc | c }
                \toprule \toprule
                \multicolumn{3}{c}{Dataset} & \multicolumn{5}{c}{\cora} & \multicolumn{5}{c}{\citeseer} & \multicolumn{5}{c}{\pubmed} \\
                \hline
                && $\Delta \rightarrow$ & 1 & 2 & 3 & 4 & 5 & 1 & 2 & 3 & 4 & 5 & 1 & 2 & 3 & 4 & 5 & Avg. Rank\\
                \cline{2-19}

                        \multirow{28}{*}{\rotatebox{90}{\textbf{Evasion}}}&\multirow{7}{*}{\rotatebox{90}{\textbf{GRAND}}}& \propseRND &\textbf{34.13 $\pm$ 14.99} &\textbf{63.47 $\pm$ 7.54} &\textbf{71.07 $\pm$ 6.54} &\textbf{71.07 $\pm$ 7.70} &\textbf{72.00 $\pm$ 8.72} &\textbf{45.47 $\pm$ 11.80} &\textbf{60.67 $\pm$ 8.64} &\textbf{63.33 $\pm$ 9.43} &\textbf{66.00 $\pm$ 8.49} &\textbf{66.40 $\pm$ 8.72} &\textbf{33.20 $\pm$ 15.62} &\textbf{39.07 $\pm$ 17.07} &\textbf{44.53 $\pm$ 16.41} &\underline{44.93 $\pm$ 16.78} &\underline{44.67 $\pm$ 17.22}& 1.13\\
                        && FGA & 11.87 $\pm$ 4.87 & 14.40 $\pm$ 4.85 & 16.67 $\pm$ 5.79 & 30.40 $\pm$ 8.92 & 36.80 $\pm$ 5.17 & 13.87 $\pm$ 4.93 & 18.80 $\pm$ 4.39 & 24.27 $\pm$ 5.95 & 34.93 $\pm$ 10.44 & 45.73 $\pm$ 8.03 & 17.20 $\pm$ 5.94 & 20.13 $\pm$ 3.25 & 20.40 $\pm$ 3.56 & 30.13 $\pm$ 8.40 & 39.33 $\pm$ 8.67& 5.13\\
                        && NETTACK & 12.67 $\pm$ 5.64 &\underline{30.00 $\pm$ 6.14} &\underline{39.87 $\pm$ 5.10} &\underline{48.13 $\pm$ 5.15} &\underline{51.73 $\pm$ 4.53} & 16.00 $\pm$ 4.21 &\underline{34.13 $\pm$ 7.46} &\underline{47.33 $\pm$ 9.70} &\underline{54.53 $\pm$ 9.81} &\underline{57.73 $\pm$ 8.84} & 19.20 $\pm$ 5.65 &\underline{24.00 $\pm$ 8.59} &\underline{44.00 $\pm$ 5.76} &\textbf{47.87 $\pm$ 7.69} &\textbf{49.87 $\pm$ 9.15}& 2.40\\
\bottomrule \bottomrule
    \end{tabular}
    }
    \end{table}

    \begin{table}[ht]
    \centering
    \caption{{\textbf{Defended Homophily Results - 3/3.} Evaluating six adversarial attacks on four defense NoisyGNN - Miss-classification rate (\%). Higher is better.  Best performance in \textbf{bold}, second best \underline{underlined}.}}
    \small
    \renewcommand\arraystretch{1.2}
    \setlength{\tabcolsep}{3pt}
    \label{tab:eval_defense_homo_p3}
        \resizebox{\linewidth}{!}{
                \begin{tabular}{ c | cc | ccccc | ccccc | ccccc | c }
                \toprule \toprule
                              \multicolumn{3}{c}{Dataset} & \multicolumn{5}{c}{\cora} & \multicolumn{5}{c}{\citeseer} & \multicolumn{5}{c}{\pubmed} \\
                \hline
                && $\Delta \rightarrow$ & 1 & 2 & 3 & 4 & 5 & 1 & 2 & 3 & 4 & 5 & 1 & 2 & 3 & 4 & 5 & Avg. Rank\\
                \cline{2-19}

                        \multirow{7}{*}{\rotatebox{90}{\textbf{Evasion}}}&\multirow{7}{*}{\rotatebox{90}{\textbf{NoisyGNN}}}& \propseRND &\textbf{28.40 $\pm$ 10.56} & 31.60 $\pm$ 12.05 & 38.80 $\pm$ 7.88 & 40.80 $\pm$ 8.61 & 41.20 $\pm$ 9.06 &\textbf{35.60 $\pm$ 11.96} & 36.00 $\pm$ 11.81 & 40.27 $\pm$ 8.58 & 42.93 $\pm$ 6.36 & 45.20 $\pm$ 5.33 &\textbf{31.07 $\pm$ 15.43} & 29.20 $\pm$ 11.13 & 28.80 $\pm$ 8.13 & 25.73 $\pm$ 6.36 & 26.40 $\pm$ 6.38& 3.73\\
                        && FGA & 20.93 $\pm$ 4.06 & 25.07 $\pm$ 6.84 & 32.00 $\pm$ 5.01 & 38.67 $\pm$ 6.03 & 42.00 $\pm$ 8.42 & 22.13 $\pm$ 3.42 & 25.87 $\pm$ 5.78 & 34.00 $\pm$ 9.23 & 42.80 $\pm$ 11.75 & 46.93 $\pm$ 11.63 & 0.00 $\pm$ 0.00 & 0.00 $\pm$ 0.00 & 0.00 $\pm$ 0.00 & 0.00 $\pm$ 0.00 & 0.00 $\pm$ 0.00& 5.93\\
                        && NETTACK & 23.33 $\pm$ 5.00 &\underline{35.87 $\pm$ 8.40} &\textbf{50.27 $\pm$ 8.55} &\textbf{60.27 $\pm$ 9.59} &\textbf{69.07 $\pm$ 9.88} & 26.93 $\pm$ 6.96 &\textbf{40.67 $\pm$ 11.78} &\textbf{49.87 $\pm$ 10.13} &\underline{53.20 $\pm$ 13.79} &\textbf{60.67 $\pm$ 11.82} & 21.20 $\pm$ 5.12 &\textbf{34.53 $\pm$ 4.75} &\textbf{46.27 $\pm$ 7.28} &\textbf{52.53 $\pm$ 7.42} &\textbf{60.53 $\pm$ 3.07}& 1.60\\     
                        &&  PGD & 24.67 $\pm$ 4.88 & 31.47 $\pm$ 4.75 & 40.67 $\pm$ 8.57 & 44.93 $\pm$ 8.94 & 49.47 $\pm$ 8.23 & 24.93 $\pm$ 4.06 & 34.00 $\pm$ 7.48 & 40.67 $\pm$ 8.09 & 44.80 $\pm$ 9.19 & 48.93 $\pm$ 8.58 & 0.00 $\pm$ 0.00 & 0.00 $\pm$ 0.00 & 0.00 $\pm$ 0.00 & 0.00 $\pm$ 0.00 & 0.00 $\pm$ 0.00& 5.13\\
\bottomrule \bottomrule
    \end{tabular}
    }
    \end{table}

\end{landscape}

\begin{landscape}
     \begin{table}[ht]
    \centering

    \caption{\textbf{Defended Heterophily Results - {1/3}.} Evaluating six adversarial attacks on two defense victim models (GCN-GARNET, ElasticGNN, GNNGuard) - Miss-classification rate (\%). Higher is better.  Best performance in \textbf{bold}, second best \underline{underlined}. GCN-Jaccard is excluded from this table due to an error caused by GCN-Jaccard's implementation.}
    \small
    \renewcommand\arraystretch{1.2}
    \setlength{\tabcolsep}{15pt}
    \label{tab:eval_defense_heter_p1}
        \resizebox{\linewidth}{!}{
                \begin{tabular}{ c | cc | ccccc | ccccc | c }
                \toprule \toprule
                \multicolumn{3}{c}{Dataset} & \multicolumn{5}{c}{\squirrel} & \multicolumn{5}{c}{\chameleon} \\
                \hline
                && $\Delta \rightarrow$ & 1 & 2 & 3 & 4 & 5 & 1 & 2 & 3 & 4 & 5 & Avg. Rank\\
                \cline{2-14}

                        \multirow{14}{*}{\rotatebox{90}{\textbf{Evasion}}}&\multirow{7}{*}{\rotatebox{90}{\textbf{ElasticGNN}}}& \propseRND & 7.87 $\pm$ 4.56 & 9.33 $\pm$ 4.32 & 15.33 $\pm$ 4.39 & 18.40 $\pm$ 4.01 & 19.87 $\pm$ 4.37 & 10.53 $\pm$ 9.78 & 16.67 $\pm$ 8.97 & 22.00 $\pm$ 11.11 & 25.60 $\pm$ 12.52 &\underline{31.73 $\pm$ 10.77}& 5.40\\
                        && FGA &\underline{23.47 $\pm$ 11.45} &\textbf{32.93 $\pm$ 8.75} &\textbf{35.73 $\pm$ 9.47} &\textbf{40.53 $\pm$ 8.53} &\textbf{42.67 $\pm$ 10.41} &\underline{15.07 $\pm$ 10.02} &\underline{21.20 $\pm$ 9.70} &\underline{25.33 $\pm$ 9.31} &\underline{28.13 $\pm$ 7.61} & 30.13 $\pm$ 7.58& 1.70\\
                        && NETTACK & 7.87 $\pm$ 3.96 & 12.00 $\pm$ 4.72 & 16.53 $\pm$ 5.37 & 20.40 $\pm$ 4.73 & 22.80 $\pm$ 5.06 & 4.00 $\pm$ 3.55 & 5.33 $\pm$ 3.44 & 7.33 $\pm$ 3.35 & 8.13 $\pm$ 3.58 & 8.67 $\pm$ 4.25& 6.40\\
                        &&  PGD & 18.93 $\pm$ 9.88 & 25.07 $\pm$ 10.74 & 29.60 $\pm$ 9.48 & 28.93 $\pm$ 10.71 & 31.60 $\pm$ 11.72 & 14.93 $\pm$ 8.03 & 19.07 $\pm$ 7.48 & 22.13 $\pm$ 7.27 & 27.20 $\pm$ 5.12 & 27.87 $\pm$ 6.70& 3.10\\
                        && \prbcdNA & 12.13 $\pm$ 3.81 & 15.73 $\pm$ 5.70 & 18.13 $\pm$ 5.83 & 19.33 $\pm$ 6.13 & 20.13 $\pm$ 6.21 & 12.27 $\pm$ 7.59 & 14.40 $\pm$ 8.85 & 16.67 $\pm$ 8.80 & 17.20 $\pm$ 7.55 & 18.40 $\pm$ 8.32& 5.40\\
                        && SGA &\textbf{23.73 $\pm$ 4.53} &\underline{29.33 $\pm$ 8.13} &\underline{35.20 $\pm$ 6.36} &\underline{39.20 $\pm$ 7.66} &\underline{41.07 $\pm$ 8.28} &\textbf{19.33 $\pm$ 6.91} &\textbf{25.60 $\pm$ 9.26} &\textbf{28.40 $\pm$ 10.03} &\textbf{32.67 $\pm$ 10.05} &\textbf{34.13 $\pm$ 9.33}& 1.40\\
                        && GOttack & 14.93 $\pm$ 5.12 & 17.60 $\pm$ 7.79 & 24.40 $\pm$ 6.81 & 24.93 $\pm$ 7.63 & 30.67 $\pm$ 8.64 & 14.67 $\pm$ 9.15 & 13.60 $\pm$ 8.72 & 18.40 $\pm$ 8.92 & 16.13 $\pm$ 8.33 & 20.80 $\pm$ 7.40& 4.60\\
                        \cline{2-14}
                    &\multirow{7}{*}{\rotatebox{90}{\textbf{GCN-GARNET}}}& \propseRND &\underline{20.00 $\pm$ 12.47} &\textbf{30.40 $\pm$ 13.57} &\textbf{35.47 $\pm$ 16.71} &\textbf{39.87 $\pm$ 14.21} &\textbf{36.67 $\pm$ 13.73} &\textbf{19.87 $\pm$ 9.61} &\underline{23.20 $\pm$ 10.33} &\textbf{27.07 $\pm$ 10.87} &\underline{25.07 $\pm$ 11.76} &\textbf{35.60 $\pm$ 10.15}& 1.30\\
                        && FGA & 15.87 $\pm$ 12.73 & 17.87 $\pm$ 11.27 & 22.13 $\pm$ 11.20 & 22.27 $\pm$ 10.87 & 24.53 $\pm$ 10.49 & 17.07 $\pm$ 10.58 & 16.80 $\pm$ 11.88 & 22.00 $\pm$ 10.20 &\textbf{26.40 $\pm$ 12.05} & 21.87 $\pm$ 9.72& 3.20\\
                        && NETTACK & 2.13 $\pm$ 1.41 & 1.73 $\pm$ 1.49 & 2.67 $\pm$ 2.09 & 2.27 $\pm$ 2.60 & 1.73 $\pm$ 1.49 & 3.07 $\pm$ 2.71 & 3.60 $\pm$ 2.75 & 3.33 $\pm$ 2.47 & 4.13 $\pm$ 2.33 & 4.40 $\pm$ 2.64& 7.00\\
                        &&  PGD & 10.40 $\pm$ 5.25 & 16.00 $\pm$ 7.86 & 19.47 $\pm$ 7.80 & 20.00 $\pm$ 8.38 & 21.47 $\pm$ 9.98 & 12.27 $\pm$ 7.28 & 15.33 $\pm$ 8.84 & 22.53 $\pm$ 8.96 & 22.93 $\pm$ 6.63 & 22.00 $\pm$ 8.45& 4.50\\
                        && \prbcdNA &\textbf{23.33 $\pm$ 10.81} &\underline{27.07 $\pm$ 9.62} &\underline{27.47 $\pm$ 13.26} &\underline{30.80 $\pm$ 13.39} &\underline{29.47 $\pm$ 11.72} & 15.73 $\pm$ 9.32 &\textbf{24.40 $\pm$ 8.49} &\underline{25.20 $\pm$ 5.49} & 22.40 $\pm$ 5.08 &\underline{24.00 $\pm$ 8.65}& 2.20\\
                        && SGA & 13.60 $\pm$ 7.75 & 17.20 $\pm$ 7.44 & 19.73 $\pm$ 10.00 & 21.47 $\pm$ 9.02 & 23.20 $\pm$ 7.92 &\underline{19.87 $\pm$ 9.75} & 21.20 $\pm$ 9.53 & 19.73 $\pm$ 8.61 & 21.33 $\pm$ 10.68 & 22.80 $\pm$ 10.60& 3.80\\
                        && GOttack & 5.33 $\pm$ 4.64 & 3.73 $\pm$ 2.71 & 4.27 $\pm$ 3.01 & 5.33 $\pm$ 2.79 & 5.60 $\pm$ 3.04 & 8.27 $\pm$ 4.27 & 6.67 $\pm$ 3.35 & 10.27 $\pm$ 4.83 & 12.00 $\pm$ 8.18 & 9.33 $\pm$ 4.39& 6.00\\
                        \cline{2-14}
                
                        \cline{2-14}
                    &\multirow{7}{*}{\rotatebox{90}{\textbf{GNNGuard}}}& \propseRND &\textbf{35.73 $\pm$ 28.31} &\textbf{35.73 $\pm$ 28.31} &\textbf{35.73 $\pm$ 28.31} &\textbf{35.73 $\pm$ 28.31} &\textbf{35.73 $\pm$ 28.31} &\textbf{0.00 $\pm$ 0.00} &\textbf{0.00 $\pm$ 0.00} &\textbf{0.00 $\pm$ 0.00} &\textbf{0.00 $\pm$ 0.00} &\textbf{0.00 $\pm$ 0.00}& 1.00\\
                        && FGA &\underline{35.73 $\pm$ 28.31} &\underline{35.73 $\pm$ 28.31} &\underline{35.73 $\pm$ 28.31} & 35.47 $\pm$ 28.04 & 34.80 $\pm$ 27.21 &\underline{0.00 $\pm$ 0.00} &\underline{0.00 $\pm$ 0.00} &\underline{0.00 $\pm$ 0.00} &\underline{0.00 $\pm$ 0.00} &\underline{0.00 $\pm$ 0.00}& 2.60\\
                        && NETTACK & 18.80 $\pm$ 15.47 & 18.80 $\pm$ 15.47 & 18.80 $\pm$ 15.47 & 18.80 $\pm$ 15.47 & 18.80 $\pm$ 15.47 & 0.00 $\pm$ 0.00 & 0.00 $\pm$ 0.00 & 0.00 $\pm$ 0.00 & 0.00 $\pm$ 0.00 & 0.00 $\pm$ 0.00& 5.00\\
                        &&  PGD & 35.73 $\pm$ 28.31 & 35.73 $\pm$ 28.31 & 35.73 $\pm$ 28.31 &\underline{35.73 $\pm$ 28.31} &\underline{35.73 $\pm$ 28.31} & 0.00 $\pm$ 0.00 & 0.00 $\pm$ 0.00 & 0.00 $\pm$ 0.00 & 0.00 $\pm$ 0.00 & 0.00 $\pm$ 0.00& 3.30\\
                        && \prbcdNA & 35.73 $\pm$ 28.31 & 35.73 $\pm$ 28.31 & 35.73 $\pm$ 28.31 & 35.73 $\pm$ 28.31 & 35.73 $\pm$ 28.31 & 0.00 $\pm$ 0.00 & 0.00 $\pm$ 0.00 & 0.00 $\pm$ 0.00 & 0.00 $\pm$ 0.00 & 0.00 $\pm$ 0.00& 4.30\\
                        && SGA & 35.73 $\pm$ 28.31 & 35.73 $\pm$ 28.31 & 35.73 $\pm$ 28.31 & 35.73 $\pm$ 28.31 & 35.73 $\pm$ 28.31 & 0.00 $\pm$ 0.00 & 0.00 $\pm$ 0.00 & 0.00 $\pm$ 0.00 & 0.00 $\pm$ 0.00 & 0.00 $\pm$ 0.00& 5.30\\
                        && GOttack & 22.13 $\pm$ 17.54 & 22.13 $\pm$ 17.54 & 22.13 $\pm$ 17.54 & 22.13 $\pm$ 17.54 & 22.13 $\pm$ 17.54 & 0.00 $\pm$ 0.00 & 0.00 $\pm$ 0.00 & 0.00 $\pm$ 0.00 & 0.00 $\pm$ 0.00 & 0.00 $\pm$ 0.00& 6.50\\
                        \cline{2-14}
                    \hline
                        \multirow{14}{*}{\rotatebox{90}{\textbf{Poison}}}&\multirow{7}{*}{\rotatebox{90}{\textbf{ElasticGNN}}}& \propseRND & 16.67 $\pm$ 7.92 & 21.07 $\pm$ 5.23 & 24.67 $\pm$ 7.12 & 26.67 $\pm$ 6.62 & 27.33 $\pm$ 8.57 & 19.07 $\pm$ 9.07 & 25.60 $\pm$ 7.75 & 31.73 $\pm$ 9.85 & 34.80 $\pm$ 10.63 & 42.67 $\pm$ 9.52& 5.50\\
                        && FGA &\textbf{38.80 $\pm$ 13.15} &\textbf{47.73 $\pm$ 11.41} &\underline{47.73 $\pm$ 11.18} &\underline{53.33 $\pm$ 11.70} &\underline{54.53 $\pm$ 13.80} &\underline{29.73 $\pm$ 11.66} &\textbf{40.27 $\pm$ 8.97} &\textbf{44.13 $\pm$ 11.43} &\textbf{49.60 $\pm$ 12.36} &\textbf{49.60 $\pm$ 13.46}& 1.40\\
                        && NETTACK & 12.80 $\pm$ 4.95 & 16.80 $\pm$ 6.45 & 19.33 $\pm$ 4.76 & 22.13 $\pm$ 4.44 & 24.00 $\pm$ 4.90 & 7.60 $\pm$ 4.48 & 8.40 $\pm$ 3.04 & 10.13 $\pm$ 5.10 & 12.27 $\pm$ 5.18 & 11.60 $\pm$ 5.46& 7.00\\
                        &&  PGD & 28.40 $\pm$ 10.93 & 35.33 $\pm$ 14.40 & 39.87 $\pm$ 8.96 & 42.67 $\pm$ 13.17 & 46.80 $\pm$ 13.64 & 28.40 $\pm$ 12.31 & 35.87 $\pm$ 10.73 & 40.13 $\pm$ 14.43 & 44.00 $\pm$ 10.11 & 45.73 $\pm$ 12.49& 3.00\\
                        && \prbcdNA & 24.00 $\pm$ 8.42 & 24.27 $\pm$ 7.92 & 27.47 $\pm$ 9.46 & 29.87 $\pm$ 8.77 & 32.40 $\pm$ 9.69 & 24.40 $\pm$ 7.57 & 30.67 $\pm$ 9.55 & 32.00 $\pm$ 11.34 & 37.33 $\pm$ 12.11 & 37.60 $\pm$ 9.45& 4.40\\
                        && SGA &\underline{35.20 $\pm$ 9.97} &\underline{45.60 $\pm$ 12.74} &\textbf{49.60 $\pm$ 11.42} &\textbf{54.80 $\pm$ 10.74} &\textbf{56.67 $\pm$ 11.73} &\textbf{31.33 $\pm$ 6.40} &\underline{38.53 $\pm$ 10.07} &\underline{42.40 $\pm$ 8.53} &\underline{48.67 $\pm$ 10.73} &\underline{48.67 $\pm$ 10.38}& 1.60\\
                        && GOttack & 20.13 $\pm$ 7.03 & 23.20 $\pm$ 7.04 & 28.93 $\pm$ 6.41 & 30.00 $\pm$ 9.13 & 33.60 $\pm$ 8.39 & 19.73 $\pm$ 9.22 & 17.47 $\pm$ 7.07 & 23.73 $\pm$ 8.55 & 21.07 $\pm$ 8.65 & 27.73 $\pm$ 6.54& 5.10\\
                        \cline{2-14}
                    &\multirow{7}{*}{\rotatebox{90}{\textbf{GCN-GARNET}}}& \propseRND &\underline{31.07 $\pm$ 8.78} &\textbf{37.60 $\pm$ 11.39} &\textbf{39.20 $\pm$ 13.33} &\textbf{44.13 $\pm$ 12.64} &\textbf{42.00 $\pm$ 11.54} &\textbf{35.07 $\pm$ 10.17} & 30.80 $\pm$ 9.99 &\textbf{38.67 $\pm$ 10.08} &\textbf{39.73 $\pm$ 15.65} &\textbf{44.40 $\pm$ 12.47}& 1.30\\
                        && FGA & 28.13 $\pm$ 8.16 & 28.67 $\pm$ 8.57 & 31.47 $\pm$ 8.37 & 30.00 $\pm$ 8.25 & 33.87 $\pm$ 11.53 & 30.53 $\pm$ 7.46 & 30.13 $\pm$ 11.55 & 34.13 $\pm$ 10.10 & 36.53 $\pm$ 9.64 & 31.07 $\pm$ 9.74& 3.60\\
                        && NETTACK & 6.13 $\pm$ 3.81 & 3.73 $\pm$ 2.49 & 5.07 $\pm$ 2.60 & 6.00 $\pm$ 3.46 & 4.53 $\pm$ 2.45 & 7.20 $\pm$ 5.94 & 9.20 $\pm$ 5.12 & 7.20 $\pm$ 4.71 & 11.47 $\pm$ 7.39 & 9.33 $\pm$ 6.13& 7.00\\
                        &&  PGD & 21.87 $\pm$ 6.61 & 26.93 $\pm$ 5.70 & 30.53 $\pm$ 7.95 & 30.00 $\pm$ 9.23 & 30.13 $\pm$ 7.80 & 28.67 $\pm$ 9.22 & 30.40 $\pm$ 7.18 & 36.53 $\pm$ 9.52 & 36.13 $\pm$ 7.61 &\underline{36.00 $\pm$ 9.32}& 4.10\\
                        && \prbcdNA &\textbf{32.27 $\pm$ 9.85} &\underline{34.93 $\pm$ 10.00} &\underline{37.60 $\pm$ 13.44} &\underline{36.67 $\pm$ 12.09} &\underline{36.27 $\pm$ 9.85} & 29.87 $\pm$ 10.18 &\textbf{37.07 $\pm$ 8.61} &\underline{37.20 $\pm$ 7.55} &\underline{39.20 $\pm$ 10.14} & 35.20 $\pm$ 8.61& 2.10\\
                        && SGA & 24.40 $\pm$ 5.82 & 26.13 $\pm$ 5.37 & 29.20 $\pm$ 8.03 & 30.67 $\pm$ 8.54 & 32.80 $\pm$ 4.95 &\underline{33.47 $\pm$ 10.68} &\underline{33.87 $\pm$ 6.48} & 32.40 $\pm$ 9.63 & 34.67 $\pm$ 13.87 & 34.93 $\pm$ 8.28& 3.90\\
                        && GOttack & 9.60 $\pm$ 4.01 & 7.73 $\pm$ 4.89 & 8.27 $\pm$ 2.81 & 8.93 $\pm$ 3.53 & 10.27 $\pm$ 3.10 & 18.00 $\pm$ 6.59 & 16.40 $\pm$ 6.77 & 22.00 $\pm$ 8.55 & 22.00 $\pm$ 11.46 & 20.93 $\pm$ 7.96& 6.00\\
                    \cline{2-14}

                    &\multirow{7}{*}{\rotatebox{90}{\textbf{GNNGuard}}}& \propseRND &\textbf{35.73 $\pm$ 28.31} &\textbf{35.73 $\pm$ 28.31} &\textbf{35.73 $\pm$ 28.31} &\textbf{35.73 $\pm$ 28.31} &\textbf{35.73 $\pm$ 28.31} &\textbf{0.00 $\pm$ 0.00} &\textbf{0.00 $\pm$ 0.00} &\textbf{0.00 $\pm$ 0.00} &\textbf{0.00 $\pm$ 0.00} &\textbf{0.00 $\pm$ 0.00}& 1.00\\
                        && FGA &\underline{35.73 $\pm$ 28.31} &\underline{35.73 $\pm$ 28.31} &\underline{35.73 $\pm$ 28.31} & 35.47 $\pm$ 28.04 & 34.80 $\pm$ 27.21 &\underline{0.00 $\pm$ 0.00} &\underline{0.00 $\pm$ 0.00} &\underline{0.00 $\pm$ 0.00} &\underline{0.00 $\pm$ 0.00} &\underline{0.00 $\pm$ 0.00}& 2.60\\
                        && NETTACK & 18.80 $\pm$ 15.47 & 18.80 $\pm$ 15.47 & 18.80 $\pm$ 15.47 & 18.80 $\pm$ 15.47 & 18.80 $\pm$ 15.47 & 0.00 $\pm$ 0.00 & 0.00 $\pm$ 0.00 & 0.00 $\pm$ 0.00 & 0.00 $\pm$ 0.00 & 0.00 $\pm$ 0.00& 5.00\\
                        &&  PGD & 35.73 $\pm$ 28.31 & 35.73 $\pm$ 28.31 & 35.73 $\pm$ 28.31 &\underline{35.73 $\pm$ 28.31} &\underline{35.73 $\pm$ 28.31} & 0.00 $\pm$ 0.00 & 0.00 $\pm$ 0.00 & 0.00 $\pm$ 0.00 & 0.00 $\pm$ 0.00 & 0.00 $\pm$ 0.00& 3.30\\
                        && \prbcdNA & 35.73 $\pm$ 28.31 & 35.73 $\pm$ 28.31 & 35.73 $\pm$ 28.31 & 35.73 $\pm$ 28.31 & 35.73 $\pm$ 28.31 & 0.00 $\pm$ 0.00 & 0.00 $\pm$ 0.00 & 0.00 $\pm$ 0.00 & 0.00 $\pm$ 0.00 & 0.00 $\pm$ 0.00& 4.30\\
                        && SGA & 35.73 $\pm$ 28.31 & 35.73 $\pm$ 28.31 & 35.73 $\pm$ 28.31 & 35.73 $\pm$ 28.31 & 35.73 $\pm$ 28.31 & 0.00 $\pm$ 0.00 & 0.00 $\pm$ 0.00 & 0.00 $\pm$ 0.00 & 0.00 $\pm$ 0.00 & 0.00 $\pm$ 0.00& 5.30\\
                        && GOttack & 22.13 $\pm$ 17.54 & 22.13 $\pm$ 17.54 & 22.13 $\pm$ 17.54 & 22.13 $\pm$ 17.54 & 22.13 $\pm$ 17.54 & 0.00 $\pm$ 0.00 & 0.00 $\pm$ 0.00 & 0.00 $\pm$ 0.00 & 0.00 $\pm$ 0.00 & 0.00 $\pm$ 0.00& 6.50\\
    \bottomrule \bottomrule
    \end{tabular}
    }
    \end{table}

     \begin{table}[ht]
    \centering
    \caption{\textbf{Defended Heterophily Results - {2/3}.} Evaluating six adversarial attacks on four vanilla attack victim models (GRAND, RobustGCN, GCORN RUNG)  - Miss-classification rate (\%). Higher is better.  Best performance in \textbf{bold}, second best \underline{underlined}.}
    \small
    \renewcommand\arraystretch{1.2}
    \setlength{\tabcolsep}{15pt}
    \label{tab:eval_defense_heter_p2}
        \resizebox{\linewidth}{!}{
                \begin{tabular}{ c | cc | ccccc | ccccc | c }
                \toprule \toprule
                \multicolumn{3}{c}{Dataset} & \multicolumn{5}{c}{\squirrel} & \multicolumn{5}{c}{\chameleon} \\
                \hline
                && $\Delta \rightarrow$ & 1 & 2 & 3 & 4 & 5 & 1 & 2 & 3 & 4 & 5 & Avg. Rank\\
                \cline{2-14}

                       \multirow{28}{*}{\rotatebox{90}{\textbf{Evasion}}}&\multirow{7}{*}{\rotatebox{90}{\textbf{GRAND}}}& \propseRND &\textbf{10.93 $\pm$ 10.74} &\textbf{8.13 $\pm$ 9.78} &\textbf{8.80 $\pm$ 10.58} &\textbf{9.33 $\pm$ 12.34} &\textbf{8.93 $\pm$ 11.26} & 2.67 $\pm$ 5.49 &\underline{6.00 $\pm$ 10.58} &\textbf{6.93 $\pm$ 12.28} &\textbf{7.73 $\pm$ 14.36} &\textbf{7.60 $\pm$ 14.21}& 1.30\\
                        && FGA & 2.00 $\pm$ 3.46 & 2.00 $\pm$ 2.83 & 2.53 $\pm$ 3.74 & 5.20 $\pm$ 7.51 & 3.60 $\pm$ 4.55 & 0.13 $\pm$ 0.52 & 1.20 $\pm$ 2.24 & 1.73 $\pm$ 4.13 & 2.53 $\pm$ 5.58 & 2.93 $\pm$ 6.71& 6.10\\
                        && NETTACK & 2.00 $\pm$ 2.73 & 2.53 $\pm$ 3.16 & 2.53 $\pm$ 3.16 & 3.07 $\pm$ 3.69 & 3.07 $\pm$ 3.69 & 2.13 $\pm$ 4.17 & 2.67 $\pm$ 5.59 & 2.27 $\pm$ 4.53 & 2.27 $\pm$ 4.53 & 2.27 $\pm$ 4.53& 6.00\\
                        &&  PGD & 2.40 $\pm$ 3.79 & 2.53 $\pm$ 3.16 & 3.73 $\pm$ 4.06 & 3.73 $\pm$ 4.27 & 3.60 $\pm$ 3.79 &\underline{3.20 $\pm$ 7.55} & 2.27 $\pm$ 4.13 & 5.47 $\pm$ 9.98 &\underline{7.20 $\pm$ 13.91} & 6.27 $\pm$ 11.21& 4.20\\
                        && \prbcdNA &\underline{6.27 $\pm$ 7.36} &\underline{6.80 $\pm$ 8.03} &\underline{6.40 $\pm$ 7.18} &\underline{6.27 $\pm$ 7.13} &\underline{6.53 $\pm$ 7.54} &\textbf{3.47 $\pm$ 6.16} &\textbf{6.53 $\pm$ 13.10} &\underline{6.80 $\pm$ 13.56} & 6.80 $\pm$ 13.02 &\underline{7.07 $\pm$ 13.56}& 1.90\\
                        && SGA & 3.47 $\pm$ 3.34 & 4.53 $\pm$ 4.17 & 4.67 $\pm$ 4.32 & 4.80 $\pm$ 4.77 & 5.33 $\pm$ 5.16 & 1.73 $\pm$ 3.53 & 2.40 $\pm$ 4.61 & 4.80 $\pm$ 8.17 & 2.80 $\pm$ 5.23 & 4.27 $\pm$ 8.94& 4.10\\
                        && GOttack & 3.60 $\pm$ 4.79 & 2.67 $\pm$ 2.99 & 4.13 $\pm$ 4.81 & 2.93 $\pm$ 3.28 & 4.13 $\pm$ 4.44 & 2.67 $\pm$ 5.27 & 2.67 $\pm$ 6.31 & 4.13 $\pm$ 9.36 & 3.47 $\pm$ 9.27 & 4.27 $\pm$ 9.82& 4.40\\
                        \cline{2-14}
                    &\multirow{7}{*}{\rotatebox{90}{\textbf{RobustGCN}}}& \propseRND & 9.73 $\pm$ 14.38 & 25.33 $\pm$ 9.09 & 40.40 $\pm$ 5.03 & 41.73 $\pm$ 6.58 & 46.40 $\pm$ 7.02 & 11.07 $\pm$ 13.43 & 31.47 $\pm$ 7.39 &\underline{41.60 $\pm$ 9.36} & 47.60 $\pm$ 9.36 & 46.67 $\pm$ 8.16& 4.40\\
                        && FGA &\underline{35.47 $\pm$ 12.52} &\textbf{46.27 $\pm$ 12.33} &\textbf{50.27 $\pm$ 9.85} &\textbf{49.33 $\pm$ 12.06} &\textbf{52.00 $\pm$ 11.44} &\textbf{38.13 $\pm$ 9.52} &\textbf{45.87 $\pm$ 10.18} &\textbf{45.33 $\pm$ 11.65} &\textbf{51.73 $\pm$ 11.66} &\textbf{49.73 $\pm$ 11.63}& 1.10\\
                        && NETTACK & 9.60 $\pm$ 4.48 & 13.33 $\pm$ 4.70 & 15.60 $\pm$ 6.47 & 17.47 $\pm$ 6.78 & 20.27 $\pm$ 6.80 & 6.27 $\pm$ 3.53 & 7.20 $\pm$ 3.10 & 7.07 $\pm$ 4.40 & 8.40 $\pm$ 5.57 & 10.13 $\pm$ 6.30& 7.00\\
                        &&  PGD & 35.47 $\pm$ 11.67 & 41.20 $\pm$ 13.02 & 41.07 $\pm$ 13.37 & 44.80 $\pm$ 16.90 & 42.93 $\pm$ 12.46 &\underline{32.13 $\pm$ 7.07} &\underline{40.67 $\pm$ 10.52} & 41.20 $\pm$ 10.63 &\underline{48.27 $\pm$ 11.08} &\underline{48.40 $\pm$ 12.40}& 3.00\\
                        && \prbcdNA & 35.47 $\pm$ 7.50 & 41.47 $\pm$ 7.95 &\underline{45.20 $\pm$ 6.36} & 44.80 $\pm$ 5.54 &\underline{49.07 $\pm$ 7.28} & 28.93 $\pm$ 7.09 & 33.33 $\pm$ 9.70 & 40.13 $\pm$ 10.60 & 39.33 $\pm$ 10.27 & 38.67 $\pm$ 10.89& 3.70\\
                        && SGA &\textbf{40.00 $\pm$ 5.86} &\underline{44.80 $\pm$ 8.65} & 44.67 $\pm$ 8.09 &\underline{48.13 $\pm$ 7.73} & 48.67 $\pm$ 7.43 & 32.13 $\pm$ 9.18 & 36.53 $\pm$ 9.98 & 37.60 $\pm$ 10.40 & 43.60 $\pm$ 12.19 & 43.73 $\pm$ 11.23& 3.00\\
                        && GOttack & 18.40 $\pm$ 6.98 & 17.20 $\pm$ 8.10 & 21.47 $\pm$ 7.54 & 22.13 $\pm$ 9.84 & 27.60 $\pm$ 8.08 & 26.40 $\pm$ 7.90 & 26.00 $\pm$ 8.94 & 30.93 $\pm$ 10.02 & 24.53 $\pm$ 9.36 & 30.27 $\pm$ 6.71& 5.80\\
                        \cline{2-14}
                    &\multirow{7}{*}{\rotatebox{90}{\textbf{GCORN}}}& \propseRND & 18.67 $\pm$ 15.72 & 26.13 $\pm$ 17.00 & 29.20 $\pm$ 18.15 & 32.27 $\pm$ 16.97 & 34.53 $\pm$ 17.39 & 16.53 $\pm$ 15.94 &\underline{32.53 $\pm$ 12.34} &\underline{38.13 $\pm$ 14.65} &\underline{41.47 $\pm$ 18.60} &\underline{45.73 $\pm$ 20.60}& 3.10\\
                        && FGA &\textbf{31.07 $\pm$ 21.02} &\textbf{33.47 $\pm$ 16.54} &\textbf{36.27 $\pm$ 18.79} &\textbf{36.67 $\pm$ 19.00} &\underline{37.60 $\pm$ 19.14} &\textbf{34.13 $\pm$ 19.23} &\textbf{36.53 $\pm$ 15.99} &\textbf{48.67 $\pm$ 17.03} &\textbf{48.67 $\pm$ 18.40} &\textbf{48.67 $\pm$ 19.03}& 1.10\\
                        && NETTACK & 6.67 $\pm$ 5.38 & 8.53 $\pm$ 5.73 & 9.60 $\pm$ 5.14 & 10.27 $\pm$ 5.85 & 11.07 $\pm$ 6.50 & 3.47 $\pm$ 3.58 & 5.33 $\pm$ 3.75 & 6.67 $\pm$ 4.39 & 8.13 $\pm$ 3.74 & 9.20 $\pm$ 4.06& 6.90\\
                        &&  PGD & 23.47 $\pm$ 10.57 & 31.20 $\pm$ 10.50 & 31.60 $\pm$ 11.93 & 34.00 $\pm$ 13.33 & 35.60 $\pm$ 13.12 &\underline{21.07 $\pm$ 13.69} & 32.13 $\pm$ 16.89 & 31.73 $\pm$ 17.19 & 36.00 $\pm$ 17.44 & 37.20 $\pm$ 18.25& 2.90\\
                        && \prbcdNA & 13.20 $\pm$ 13.18 & 17.07 $\pm$ 11.18 & 17.33 $\pm$ 9.93 & 19.20 $\pm$ 10.55 & 22.67 $\pm$ 10.02 & 10.13 $\pm$ 7.87 & 15.20 $\pm$ 9.47 & 17.07 $\pm$ 11.05 & 20.27 $\pm$ 12.46 & 22.93 $\pm$ 14.91& 5.00\\
                        && SGA &\underline{24.27 $\pm$ 12.14} &\underline{31.87 $\pm$ 13.10} &\underline{35.73 $\pm$ 13.31} &\underline{34.67 $\pm$ 10.49} &\textbf{37.73 $\pm$ 11.36} & 15.60 $\pm$ 7.86 & 21.33 $\pm$ 8.57 & 22.93 $\pm$ 6.84 & 27.87 $\pm$ 10.38 & 29.33 $\pm$ 9.61& 2.90\\
                        && GOttack & 9.47 $\pm$ 6.74 & 10.40 $\pm$ 7.75 & 12.93 $\pm$ 5.99 & 12.93 $\pm$ 6.13 & 14.27 $\pm$ 6.80 & 6.00 $\pm$ 4.14 & 5.07 $\pm$ 3.92 & 9.73 $\pm$ 5.70 & 10.13 $\pm$ 5.88 & 10.27 $\pm$ 5.18& 6.10\\
                        \cline{2-14}
                    &\multirow{7}{*}{\rotatebox{90}{\textbf{RUNG}}}& \propseRND &\underline{2.13 $\pm$ 1.77} & 2.13 $\pm$ 2.77 & 2.13 $\pm$ 2.20 & 2.67 $\pm$ 2.58 & 2.80 $\pm$ 2.48 &\textbf{2.27 $\pm$ 3.28} &\underline{2.27 $\pm$ 5.18} &\textbf{3.47 $\pm$ 6.61} &\underline{4.53 $\pm$ 6.65} &\textbf{5.07 $\pm$ 7.17}& 2.70\\
                        && FGA & 1.87 $\pm$ 2.88 & 1.87 $\pm$ 2.20 & 3.20 $\pm$ 3.36 & 2.93 $\pm$ 2.25 & 3.60 $\pm$ 2.53 & 0.93 $\pm$ 1.83 & 2.27 $\pm$ 3.69 & 2.53 $\pm$ 3.58 & 3.33 $\pm$ 5.22 & 4.13 $\pm$ 5.37& 3.40\\
                        && NETTACK & 0.27 $\pm$ 1.03 & 0.80 $\pm$ 1.47 & 1.33 $\pm$ 1.63 & 1.47 $\pm$ 1.77 & 1.60 $\pm$ 1.72 & 0.27 $\pm$ 0.70 & 0.67 $\pm$ 1.23 & 0.53 $\pm$ 1.19 & 0.93 $\pm$ 1.83 & 1.33 $\pm$ 2.23& 6.90\\
                        &&  PGD & 2.00 $\pm$ 1.85 &\underline{2.67 $\pm$ 2.89} &\underline{4.40 $\pm$ 2.85} &\underline{4.67 $\pm$ 2.99} &\underline{4.93 $\pm$ 3.10} & 1.87 $\pm$ 4.69 & 2.00 $\pm$ 3.02 & 1.60 $\pm$ 2.03 & 3.20 $\pm$ 4.71 & 3.47 $\pm$ 5.93& 3.10\\
                        && \prbcdNA & 1.73 $\pm$ 2.49 & 2.67 $\pm$ 2.47 & 3.20 $\pm$ 2.60 & 2.27 $\pm$ 2.12 & 3.33 $\pm$ 3.09 & 0.80 $\pm$ 2.24 & 1.87 $\pm$ 3.58 & 1.87 $\pm$ 3.96 & 2.40 $\pm$ 3.87 & 3.20 $\pm$ 5.65& 4.60\\
                        && SGA &\textbf{2.67 $\pm$ 3.68} &\textbf{3.47 $\pm$ 4.17} &\textbf{5.07 $\pm$ 4.77} &\textbf{5.07 $\pm$ 4.59} &\textbf{5.60 $\pm$ 4.15} &\underline{2.13 $\pm$ 4.31} &\textbf{2.93 $\pm$ 4.95} &\underline{3.20 $\pm$ 3.61} &\textbf{5.60 $\pm$ 7.38} &\underline{4.40 $\pm$ 6.47}& 1.30\\
                        && GOttack & 0.93 $\pm$ 2.25 & 0.93 $\pm$ 1.83 & 1.73 $\pm$ 2.71 & 1.47 $\pm$ 2.07 & 2.00 $\pm$ 2.83 & 0.67 $\pm$ 1.23 & 1.33 $\pm$ 2.09 & 1.60 $\pm$ 2.16 & 1.87 $\pm$ 3.25 & 3.47 $\pm$ 5.04& 6.00\\
                        \cline{2-14}
                    \hline
                        \multirow{28}{*}{\rotatebox{90}{\textbf{Poison}}}&\multirow{7}{*}{\rotatebox{90}{\textbf{GRAND}}}& \propseRND &\textbf{16.00 $\pm$ 13.56} &\underline{14.00 $\pm$ 13.16} & 16.80 $\pm$ 15.89 & 19.07 $\pm$ 17.81 &\underline{19.07 $\pm$ 17.63} & 4.00 $\pm$ 5.95 &\underline{9.33 $\pm$ 13.75} &\textbf{12.67 $\pm$ 19.50} &\textbf{12.40 $\pm$ 20.38} &\textbf{13.07 $\pm$ 21.93}& 2.10\\
                        && FGA &\underline{12.13 $\pm$ 21.71} &\textbf{16.53 $\pm$ 26.57} &\textbf{19.33 $\pm$ 26.08} &\textbf{21.33 $\pm$ 26.15} & 17.07 $\pm$ 17.56 & 2.00 $\pm$ 3.12 & 5.87 $\pm$ 9.66 & 5.07 $\pm$ 8.00 & 11.20 $\pm$ 24.69 & 8.93 $\pm$ 15.87& 3.30\\
                        && NETTACK & 4.40 $\pm$ 7.53 & 5.47 $\pm$ 6.91 & 6.27 $\pm$ 8.34 & 8.13 $\pm$ 10.32 & 7.47 $\pm$ 10.27 & 4.00 $\pm$ 7.09 & 5.20 $\pm$ 7.92 & 4.93 $\pm$ 7.89 & 5.20 $\pm$ 7.55 & 4.00 $\pm$ 5.55& 6.60\\
                        &&  PGD & 11.07 $\pm$ 11.66 & 12.93 $\pm$ 11.08 &\underline{18.53 $\pm$ 16.15} &\underline{21.20 $\pm$ 18.03} &\textbf{21.07 $\pm$ 17.55} & 5.33 $\pm$ 8.02 & 8.13 $\pm$ 11.80 &\underline{12.40 $\pm$ 18.64} &\underline{11.33 $\pm$ 17.13} &\underline{12.13 $\pm$ 16.59}& 2.30\\
                        && \prbcdNA & 9.07 $\pm$ 9.47 & 12.13 $\pm$ 12.39 & 11.07 $\pm$ 10.69 & 12.13 $\pm$ 12.39 & 11.33 $\pm$ 12.87 &\textbf{7.07 $\pm$ 9.07} &\textbf{9.47 $\pm$ 14.17} & 11.07 $\pm$ 16.18 & 11.07 $\pm$ 15.47 & 11.60 $\pm$ 16.87& 3.40\\
                        && SGA & 7.07 $\pm$ 6.92 & 11.87 $\pm$ 11.75 & 10.40 $\pm$ 8.22 & 14.93 $\pm$ 14.83 & 14.40 $\pm$ 13.14 &\underline{6.53 $\pm$ 11.04} & 6.00 $\pm$ 9.13 & 7.87 $\pm$ 10.89 & 6.80 $\pm$ 9.73 & 8.27 $\pm$ 11.61& 4.30\\
                        && GOttack & 6.93 $\pm$ 9.44 & 6.13 $\pm$ 10.99 & 6.80 $\pm$ 6.96 & 5.20 $\pm$ 5.28 & 6.13 $\pm$ 6.21 & 4.80 $\pm$ 7.16 & 4.00 $\pm$ 6.85 & 5.47 $\pm$ 9.61 & 6.40 $\pm$ 12.29 & 6.67 $\pm$ 10.81& 6.00\\
                        \cline{2-14}
                    &\multirow{7}{*}{\rotatebox{90}{\textbf{RobustGCN}}}& \propseRND & 26.00 $\pm$ 14.89 & 33.33 $\pm$ 10.13 & 45.60 $\pm$ 6.98 & 45.33 $\pm$ 8.13 & 48.93 $\pm$ 8.10 & 23.07 $\pm$ 11.83 & 36.13 $\pm$ 9.69 & 44.27 $\pm$ 10.17 & 49.60 $\pm$ 9.98 & 48.80 $\pm$ 8.48& 4.80\\
                        && FGA & 43.47 $\pm$ 13.64 &\textbf{51.73 $\pm$ 12.33} &\textbf{56.67 $\pm$ 11.05} &\textbf{54.93 $\pm$ 12.09} &\textbf{57.33 $\pm$ 12.69} &\textbf{43.07 $\pm$ 11.08} &\textbf{53.07 $\pm$ 11.46} &\textbf{52.67 $\pm$ 13.58} &\textbf{56.80 $\pm$ 12.78} &\underline{53.33 $\pm$ 11.99}& 1.30\\
                        && NETTACK & 14.67 $\pm$ 7.58 & 18.67 $\pm$ 8.51 & 21.60 $\pm$ 9.30 & 21.60 $\pm$ 7.72 & 24.53 $\pm$ 8.33 & 9.73 $\pm$ 5.28 & 9.73 $\pm$ 3.69 & 10.27 $\pm$ 5.50 & 12.93 $\pm$ 6.92 & 12.27 $\pm$ 8.00& 7.00\\
                        &&  PGD &\underline{44.00 $\pm$ 14.44} & 46.93 $\pm$ 14.66 & 46.53 $\pm$ 16.20 & 50.13 $\pm$ 18.51 & 50.27 $\pm$ 14.28 & 37.20 $\pm$ 8.44 &\underline{46.40 $\pm$ 8.76} &\underline{48.13 $\pm$ 11.10} &\underline{54.13 $\pm$ 9.66} &\textbf{54.27 $\pm$ 12.33}& 2.80\\
                        && \prbcdNA & 41.07 $\pm$ 7.89 & 48.00 $\pm$ 11.56 &\underline{52.27 $\pm$ 7.17} &\underline{51.33 $\pm$ 8.13} & 49.73 $\pm$ 9.28 & 37.73 $\pm$ 8.10 & 41.33 $\pm$ 11.82 & 44.53 $\pm$ 10.21 & 48.40 $\pm$ 9.51 & 44.40 $\pm$ 10.53& 3.50\\
                        && SGA &\textbf{46.00 $\pm$ 8.94} &\underline{48.80 $\pm$ 8.81} & 50.40 $\pm$ 10.48 & 50.67 $\pm$ 8.80 &\underline{53.07 $\pm$ 9.28} &\underline{40.67 $\pm$ 10.05} & 46.27 $\pm$ 10.47 & 44.13 $\pm$ 9.81 & 51.07 $\pm$ 12.65 & 50.53 $\pm$ 12.01& 2.70\\
                        && GOttack & 20.80 $\pm$ 6.92 & 24.13 $\pm$ 11.48 & 25.07 $\pm$ 8.94 & 27.33 $\pm$ 11.58 & 31.87 $\pm$ 10.78 & 29.60 $\pm$ 5.96 & 28.80 $\pm$ 9.50 & 36.80 $\pm$ 7.24 & 31.87 $\pm$ 10.32 & 35.33 $\pm$ 6.26& 5.90\\
                        \cline{2-14}
                    &\multirow{7}{*}{\rotatebox{90}{\textbf{GCORN}}}& \propseRND & 20.00 $\pm$ 16.30 & 26.67 $\pm$ 17.35 & 30.93 $\pm$ 19.71 & 33.60 $\pm$ 17.88 & 36.13 $\pm$ 17.38 & 17.33 $\pm$ 16.74 & 33.47 $\pm$ 13.06 &\underline{38.80 $\pm$ 15.08} &\underline{42.67 $\pm$ 18.07} &\underline{46.27 $\pm$ 20.41}& 3.30\\
                        && FGA &\textbf{34.27 $\pm$ 22.68} &\textbf{34.40 $\pm$ 16.67} &\underline{37.33 $\pm$ 16.26} &\textbf{40.80 $\pm$ 18.80} &\underline{37.73 $\pm$ 17.43} &\textbf{34.53 $\pm$ 17.93} &\textbf{36.80 $\pm$ 15.25} &\textbf{48.13 $\pm$ 17.06} &\textbf{49.87 $\pm$ 18.29} &\textbf{48.40 $\pm$ 18.66}& 1.20\\
                        && NETTACK & 8.13 $\pm$ 6.61 & 9.20 $\pm$ 7.48 & 10.40 $\pm$ 5.96 & 12.13 $\pm$ 6.21 & 13.07 $\pm$ 6.96 & 4.27 $\pm$ 3.45 & 5.87 $\pm$ 4.44 & 7.60 $\pm$ 4.22 & 9.47 $\pm$ 3.96 & 11.47 $\pm$ 5.68& 7.00\\
                        &&  PGD & 25.60 $\pm$ 12.03 & 33.47 $\pm$ 11.07 & 34.53 $\pm$ 11.92 & 37.33 $\pm$ 13.04 & 37.60 $\pm$ 11.69 &\underline{22.67 $\pm$ 13.81} &\underline{34.00 $\pm$ 16.82} & 33.33 $\pm$ 16.08 & 37.87 $\pm$ 16.54 & 37.73 $\pm$ 17.32& 2.80\\
                        && \prbcdNA & 14.53 $\pm$ 13.68 & 18.13 $\pm$ 10.81 & 19.07 $\pm$ 9.94 & 21.33 $\pm$ 7.84 & 24.40 $\pm$ 9.80 & 11.33 $\pm$ 7.24 & 16.13 $\pm$ 8.67 & 18.13 $\pm$ 10.70 & 21.60 $\pm$ 11.96 & 25.20 $\pm$ 14.48& 5.00\\
                        && SGA &\underline{26.13 $\pm$ 13.89} &\underline{34.40 $\pm$ 13.38} &\textbf{37.60 $\pm$ 13.80} &\underline{37.60 $\pm$ 10.96} &\textbf{39.33 $\pm$ 9.67} & 18.80 $\pm$ 8.20 & 24.80 $\pm$ 8.87 & 25.73 $\pm$ 7.25 & 29.07 $\pm$ 10.55 & 32.53 $\pm$ 10.32& 2.70\\
                        && GOttack & 10.67 $\pm$ 8.61 & 11.47 $\pm$ 9.52 & 14.67 $\pm$ 7.24 & 14.53 $\pm$ 7.58 & 15.73 $\pm$ 8.48 & 6.53 $\pm$ 3.66 & 6.40 $\pm$ 4.42 & 11.47 $\pm$ 5.42 & 12.00 $\pm$ 5.55 & 12.00 $\pm$ 5.50& 6.00\\
                        \cline{2-14}
                    &\multirow{7}{*}{\rotatebox{90}{\textbf{RUNG}}}& \propseRND & 11.33 $\pm$ 8.64 & 16.13 $\pm$ 8.26 & 17.07 $\pm$ 8.94 & 18.67 $\pm$ 8.02 & 19.33 $\pm$ 8.09 & 12.00 $\pm$ 8.88 &\textbf{21.33 $\pm$ 12.30} &\textbf{24.00 $\pm$ 10.61} &\textbf{28.00 $\pm$ 13.96} &\textbf{32.40 $\pm$ 14.35}& 3.30\\
                        && FGA &\textbf{20.53 $\pm$ 9.69} &\textbf{24.67 $\pm$ 10.27} &\textbf{29.47 $\pm$ 7.23} &\textbf{28.53 $\pm$ 9.21} &\textbf{28.67 $\pm$ 10.63} &\textbf{16.40 $\pm$ 7.53} & 17.47 $\pm$ 12.06 & 17.20 $\pm$ 7.70 & 18.80 $\pm$ 10.77 & 19.20 $\pm$ 11.18& 2.00\\
                        && NETTACK & 6.53 $\pm$ 4.44 & 6.93 $\pm$ 5.18 & 6.53 $\pm$ 4.81 & 6.93 $\pm$ 4.83 & 6.80 $\pm$ 4.26 & 7.33 $\pm$ 2.89 & 6.67 $\pm$ 2.89 & 7.07 $\pm$ 3.61 & 6.93 $\pm$ 4.06 & 7.47 $\pm$ 3.34& 6.80\\
                        &&  PGD & 17.33 $\pm$ 7.81 &\underline{23.20 $\pm$ 9.91} &\underline{26.40 $\pm$ 9.05} &\underline{27.07 $\pm$ 6.23} &\underline{27.87 $\pm$ 8.63} & 13.73 $\pm$ 6.41 &\underline{18.40 $\pm$ 7.64} & 19.20 $\pm$ 8.94 &\underline{20.00 $\pm$ 9.80} &\underline{19.60 $\pm$ 11.76}& 2.30\\
                        && \prbcdNA & 15.07 $\pm$ 6.18 & 19.47 $\pm$ 7.31 & 19.20 $\pm$ 6.84 & 20.67 $\pm$ 7.73 & 20.53 $\pm$ 6.99 & 11.73 $\pm$ 5.90 & 14.40 $\pm$ 6.15 & 13.87 $\pm$ 7.87 & 14.53 $\pm$ 5.42 & 15.73 $\pm$ 6.45& 4.50\\
                        && SGA &\underline{18.40 $\pm$ 9.33} & 22.27 $\pm$ 8.78 & 26.27 $\pm$ 8.61 & 26.93 $\pm$ 10.63 & 27.73 $\pm$ 10.11 &\underline{14.27 $\pm$ 6.76} & 15.07 $\pm$ 5.23 &\underline{20.40 $\pm$ 7.41} & 19.60 $\pm$ 7.97 & 18.00 $\pm$ 6.46& 2.90\\
                        && GOttack & 6.27 $\pm$ 5.18 & 6.93 $\pm$ 6.23 & 8.00 $\pm$ 5.66 & 7.60 $\pm$ 5.62 & 9.20 $\pm$ 6.58 & 10.80 $\pm$ 4.89 & 8.27 $\pm$ 5.80 & 9.20 $\pm$ 3.69 & 10.53 $\pm$ 3.42 & 12.53 $\pm$ 6.35& 6.20\\
    \bottomrule \bottomrule
    \end{tabular}
    }
    \end{table}

     \begin{table}[ht]
    \centering
    \caption{{\textbf{Defended Heterophily Results - 3/3.} Evaluating six adversarial attacks on four vanilla attack victim models (NoisyGNN)  - Miss-classification rate (\%). Higher is better.  Best performance in \textbf{bold}, second best \underline{underlined}.}}
    \small
    \renewcommand\arraystretch{1.2}
    \setlength{\tabcolsep}{15pt}
    \label{tab:eval_defense_heter_p3}
        \resizebox{\linewidth}{!}{
                \begin{tabular}{ c | cc | ccccc | ccccc | c }
                \toprule \toprule
                \multicolumn{3}{c}{Dataset} & \multicolumn{5}{c}{\squirrel} & \multicolumn{5}{c}{\chameleon} \\
                \hline
                && $\Delta \rightarrow$ & 1 & 2 & 3 & 4 & 5 & 1 & 2 & 3 & 4 & 5 & Avg. Rank\\
                \cline{2-14}

                        \multirow{7}{*}{\rotatebox{90}{\textbf{Evasion}}}&\multirow{7}{*}{\rotatebox{90}{\textbf{NoisyGNN}}}& \propseRND & 68.93 $\pm$ 7.05 & 67.47 $\pm$ 7.87 & 67.47 $\pm$ 7.87 & 67.47 $\pm$ 7.87 & 67.60 $\pm$ 7.83 &\textbf{63.73 $\pm$ 5.99} &\underline{64.67 $\pm$ 7.95} &\underline{64.93 $\pm$ 6.67} &\textbf{63.73 $\pm$ 6.13} & 65.87 $\pm$ 5.32& 2.70\\
                        && FGA & 66.80 $\pm$ 11.13 & 66.93 $\pm$ 10.82 & 66.53 $\pm$ 12.70 & 67.07 $\pm$ 13.33 & 64.93 $\pm$ 17.09 &\underline{62.27 $\pm$ 11.49} &\textbf{64.93 $\pm$ 14.24} &\textbf{70.00 $\pm$ 13.31} & 62.67 $\pm$ 15.50 &\underline{67.20 $\pm$ 14.54}& 3.40\\
                        && NETTACK & 26.13 $\pm$ 4.93 & 26.00 $\pm$ 5.01 & 26.00 $\pm$ 5.01 & 26.00 $\pm$ 5.01 & 26.00 $\pm$ 5.01 & 4.80 $\pm$ 8.51 & 4.80 $\pm$ 8.55 & 5.47 $\pm$ 8.26 & 6.13 $\pm$ 8.02 & 5.73 $\pm$ 8.17& 7.00\\
                        &&  PGD & 66.93 $\pm$ 8.58 & 68.40 $\pm$ 7.90 & 67.33 $\pm$ 9.00 & 67.73 $\pm$ 9.47 & 69.47 $\pm$ 8.83 & 57.60 $\pm$ 8.08 & 62.53 $\pm$ 8.70 & 63.47 $\pm$ 6.95 &\underline{63.60 $\pm$ 7.68} &\textbf{68.53 $\pm$ 6.25}& 3.00\\
                        && PR-BCD &\textbf{69.73 $\pm$ 6.54} &\textbf{69.87 $\pm$ 6.44} &\underline{69.47 $\pm$ 7.07} &\underline{69.20 $\pm$ 7.78} &\underline{70.40 $\pm$ 6.38} & 53.33 $\pm$ 7.70 & 54.80 $\pm$ 7.59 & 53.87 $\pm$ 7.73 & 53.20 $\pm$ 8.20 & 53.47 $\pm$ 8.90& 3.30\\
                        && SGA &\underline{69.07 $\pm$ 6.88} &\underline{69.73 $\pm$ 7.48} &\textbf{69.73 $\pm$ 7.74} &\textbf{70.53 $\pm$ 7.35} &\textbf{71.47 $\pm$ 6.95} & 60.13 $\pm$ 7.69 & 60.13 $\pm$ 6.52 & 60.80 $\pm$ 7.55 & 60.00 $\pm$ 7.13 & 61.60 $\pm$ 8.72& 2.60\\
                        && GOttack & 34.13 $\pm$ 4.87 & 34.27 $\pm$ 4.89 & 34.40 $\pm$ 5.03 & 34.00 $\pm$ 4.72 & 34.40 $\pm$ 4.67 & 19.20 $\pm$ 8.68 & 21.47 $\pm$ 9.87 & 20.27 $\pm$ 8.55 & 21.20 $\pm$ 9.25 & 20.80 $\pm$ 9.47& 6.00\\
                        \cline{2-14}
                    \hline
                        \multirow{7}{*}{\rotatebox{90}{\textbf{Poison}}}&\multirow{7}{*}{\rotatebox{90}{\textbf{NoisyGNN}}}& \propseRND & 71.87 $\pm$ 6.25 & 71.87 $\pm$ 6.25 & 72.27 $\pm$ 6.50 & 71.60 $\pm$ 6.38 & 72.27 $\pm$ 6.04 &\underline{64.93 $\pm$ 5.80} & 66.40 $\pm$ 7.14 & 65.87 $\pm$ 8.50 & 65.73 $\pm$ 4.71 & 67.73 $\pm$ 6.09& 4.10\\
                        && FGA & 72.53 $\pm$ 9.81 & 74.13 $\pm$ 7.31 & 70.40 $\pm$ 9.05 & 71.87 $\pm$ 8.86 & 72.67 $\pm$ 10.79 &\textbf{68.00 $\pm$ 9.41} &\textbf{74.53 $\pm$ 13.78} &\textbf{76.67 $\pm$ 13.06} &\textbf{74.00 $\pm$ 19.61} &\textbf{80.40 $\pm$ 16.65}& 2.50\\
                        && NETTACK & 29.87 $\pm$ 4.37 & 29.60 $\pm$ 5.14 & 30.53 $\pm$ 5.10 & 31.33 $\pm$ 4.82 & 31.47 $\pm$ 3.89 & 8.40 $\pm$ 9.83 & 9.47 $\pm$ 9.15 & 10.13 $\pm$ 10.32 & 9.47 $\pm$ 9.36 & 10.13 $\pm$ 8.60& 7.00\\
                        &&  PGD & 72.67 $\pm$ 7.39 &\underline{74.40 $\pm$ 7.14} &\underline{74.27 $\pm$ 8.14} & 74.40 $\pm$ 8.18 &\textbf{76.40 $\pm$ 8.82} & 61.73 $\pm$ 7.70 &\underline{68.27 $\pm$ 9.38} &\underline{71.87 $\pm$ 5.48} &\underline{73.47 $\pm$ 7.58} &\underline{77.73 $\pm$ 6.76}& 2.30\\
                        && PR-BCD &\underline{72.93 $\pm$ 5.28} & 73.47 $\pm$ 5.04 & 74.27 $\pm$ 5.01 &\underline{74.53 $\pm$ 4.87} & 73.87 $\pm$ 6.12 & 59.33 $\pm$ 7.12 & 60.53 $\pm$ 6.70 & 60.93 $\pm$ 6.50 & 62.00 $\pm$ 7.25 & 60.53 $\pm$ 6.99& 3.90\\
                        && SGA &\textbf{74.67 $\pm$ 6.62} &\textbf{74.80 $\pm$ 5.85} &\textbf{74.93 $\pm$ 6.45} &\textbf{75.47 $\pm$ 5.83} &\underline{75.07 $\pm$ 5.60} & 62.67 $\pm$ 7.88 & 65.47 $\pm$ 6.99 & 69.47 $\pm$ 7.07 & 68.40 $\pm$ 8.32 & 69.60 $\pm$ 9.39& 2.20\\
                        && GOttack & 37.87 $\pm$ 4.75 & 38.13 $\pm$ 5.58 & 39.87 $\pm$ 6.07 & 38.27 $\pm$ 6.32 & 40.00 $\pm$ 5.76 & 23.60 $\pm$ 8.29 & 26.13 $\pm$ 9.15 & 24.80 $\pm$ 10.63 & 27.20 $\pm$ 8.48 & 26.27 $\pm$ 9.68& 6.00\\
                        \cline{2-14}
                    \hline
    \end{tabular}
    }
    \end{table}

\end{landscape}

    \begin{table}[ht]
    \centering
    \caption{\textbf{Defended PRBCD on Homophily Datasets.} Evaluating PRBCD and adaptive version of PRBCD (\prbcdNA) adversarial attacks on two defense victim models (ElasticGNN, RobustGCN) - Miss-classification rate (\%). Higher is better.  Best performance in \textbf{bold}, second best \underline{underlined}}
    \label{appendixTable:prbcdDefendedHomophily}
    \small
    \renewcommand\arraystretch{3}
    \setlength{\tabcolsep}{2pt}
    \label{tab:adaptive_homo}
        \resizebox{\linewidth}{!}{
                \begin{tabular}{ c | cc | ccccc | ccccc | ccccc | c }
                \toprule \toprule
                \multicolumn{3}{c}{Dataset} & \multicolumn{5}{c}{\cora} & \multicolumn{5}{c}{\citeseer} & \multicolumn{5}{c}{\pubmed} \\
                \hline
                && $\Delta \rightarrow$ & 1 & 2 & 3 & 4 & 5 & 1 & 2 & 3 & 4 & 5 & 1 & 2 & 3 & 4 & 5 & Avg. Rank\\
                \cline{2-19}

                        \multirow{4}{*}{\rotatebox{90}{\textbf{Evasion}}}&\multirow{2}{*}{\rotatebox{90}{\textbf{ElasticGNN}}}& PR-BCD &\textbf{24.80 $\pm$ 2.81} & 30.67 $\pm$ 4.88 & 37.87 $\pm$ 7.27 &\textbf{45.47 $\pm$ 7.76} & 50.13 $\pm$ 8.16 &\textbf{23.60 $\pm$ 7.53} &\textbf{31.87 $\pm$ 8.73} &\textbf{39.20 $\pm$ 10.71} &\textbf{45.47 $\pm$ 10.76} & 49.73 $\pm$ 13.52 &\textbf{22.93 $\pm$ 3.84} &\textbf{25.60 $\pm$ 4.15} & 27.33 $\pm$ 4.94 & 29.60 $\pm$ 5.25 &\textbf{31.73 $\pm$ 5.01}& 1.40\\
                        &&  \prbcdNA & 24.67 $\pm$ 3.75 &\textbf{32.00 $\pm$ 5.76} &\textbf{41.60 $\pm$ 6.85} & 44.80 $\pm$ 7.28 &\textbf{50.40 $\pm$ 9.05} & 23.33 $\pm$ 4.76 & 30.93 $\pm$ 7.44 & 38.40 $\pm$ 9.66 & 45.47 $\pm$ 11.12 &\textbf{51.33 $\pm$ 11.87} & 22.13 $\pm$ 3.16 & 25.60 $\pm$ 4.42 &\textbf{28.00 $\pm$ 4.28} &\textbf{30.53 $\pm$ 5.53} & 30.53 $\pm$ 3.89& 1.60\\
                        \cline{2-19}
                    &\multirow{2}{*}{\rotatebox{90}{\textbf{RobustGCN}}}& PR-BCD & 27.73 $\pm$ 5.44 & 49.33 $\pm$ 5.94 & 56.13 $\pm$ 4.31 & 60.53 $\pm$ 3.58 & 62.93 $\pm$ 4.20 &\textbf{24.27 $\pm$ 4.13} & 40.80 $\pm$ 5.00 & 53.07 $\pm$ 3.20 &\textbf{56.53 $\pm$ 2.77} &\textbf{58.67 $\pm$ 2.79} & 29.60 $\pm$ 3.79 &\textbf{40.40 $\pm$ 5.36} &\textbf{48.53 $\pm$ 3.58} &\textbf{55.20 $\pm$ 2.37} & 56.27 $\pm$ 1.83& 1.60\\
                        &&  \prbcdNA &\textbf{32.40 $\pm$ 4.79} &\textbf{58.00 $\pm$ 3.85} &\textbf{62.00 $\pm$ 3.93} &\textbf{64.67 $\pm$ 3.44} &\textbf{64.93 $\pm$ 2.49} & 23.87 $\pm$ 3.74 &\textbf{42.67 $\pm$ 5.98} &\textbf{53.20 $\pm$ 4.89} & 55.87 $\pm$ 3.25 & 58.27 $\pm$ 3.53 &\textbf{31.47 $\pm$ 2.07} & 38.40 $\pm$ 4.61 & 48.53 $\pm$ 3.42 & 54.00 $\pm$ 3.12 &\textbf{57.07 $\pm$ 2.25}& 1.40\\
                        \cline{2-19}
                    \hline
                        \multirow{4}{*}{\rotatebox{90}{\textbf{Poison}}}&\multirow{2}{*}{\rotatebox{90}{\textbf{ElasticGNN}}}& PR-BCD &\textbf{27.33 $\pm$ 3.18} &\textbf{38.00 $\pm$ 7.29} &\textbf{47.47 $\pm$ 5.32} & 53.20 $\pm$ 4.83 &\textbf{60.53 $\pm$ 5.15} & 27.20 $\pm$ 9.00 &\textbf{41.20 $\pm$ 10.08} & 48.27 $\pm$ 12.40 &\textbf{54.67 $\pm$ 11.97} & 56.93 $\pm$ 12.16 &\textbf{21.87 $\pm$ 4.87} & 24.53 $\pm$ 3.96 & 26.80 $\pm$ 4.52 & 29.47 $\pm$ 5.04 &\textbf{31.47 $\pm$ 4.63}& 1.47\\
                        &&  \prbcdNA & 25.60 $\pm$ 4.15 & 36.53 $\pm$ 5.93 & 47.47 $\pm$ 6.25 &\textbf{55.33 $\pm$ 3.83} & 59.47 $\pm$ 5.04 &\textbf{29.47 $\pm$ 9.21} & 38.93 $\pm$ 11.29 &\textbf{50.93 $\pm$ 11.71} & 53.73 $\pm$ 10.22 &\textbf{57.73 $\pm$ 10.00} & 20.53 $\pm$ 3.07 &\textbf{25.60 $\pm$ 3.94} &\textbf{27.07 $\pm$ 4.83} &\textbf{30.80 $\pm$ 6.13} & 30.80 $\pm$ 4.06& 1.53\\      
                        \cline{2-19}
                    &\multirow{2}{*}{\rotatebox{90}{\textbf{RobustGCN}}}& PR-BCD & 30.93 $\pm$ 4.40 & 50.00 $\pm$ 4.47 & 58.00 $\pm$ 3.78 & 60.13 $\pm$ 2.97 & 62.40 $\pm$ 4.61 &\textbf{29.47 $\pm$ 4.10} & 48.13 $\pm$ 4.98 & 55.60 $\pm$ 2.53 &\textbf{58.27 $\pm$ 3.37} & 59.60 $\pm$ 3.94 & 29.47 $\pm$ 3.34 &\textbf{40.53 $\pm$ 4.56} &\textbf{50.27 $\pm$ 3.45} &\textbf{55.60 $\pm$ 2.95} & 56.27 $\pm$ 1.98& 1.67\\
                        &&  \prbcdNA &\textbf{34.40 $\pm$ 4.08} &\textbf{57.20 $\pm$ 3.19} &\textbf{61.73 $\pm$ 3.37} &\textbf{64.13 $\pm$ 2.97} &\textbf{65.33 $\pm$ 2.99} & 28.67 $\pm$ 5.49 &\textbf{49.60 $\pm$ 6.56} &\textbf{56.00 $\pm$ 4.47} & 57.47 $\pm$ 4.37 &\textbf{60.00 $\pm$ 4.07} &\textbf{31.33 $\pm$ 2.58} & 38.93 $\pm$ 4.89 & 49.47 $\pm$ 3.50 & 54.27 $\pm$ 2.12 &\textbf{56.93 $\pm$ 2.12}& 1.33\\
                        \cline{2-19}
                    \hline
    \bottomrule
    \end{tabular}
    }
    \end{table}

    \begin{table}[ht]
    \centering
    \caption{\textbf{Defended PRBCD on Heterophily Datasets.} Evaluating non adaptive PR-BCD (\prbcdNA), and adaptive version of PR-BCD adversarial attacks on two defense victim models (ElasticGNN, RobustGCN) - Miss-classification rate (\%). Higher is better.  Best performance in \textbf{bold}, second best \underline{underlined}}
     \label{appendixTable:prbcdDefendedHeterophily}
    \small
   \renewcommand\arraystretch{3}
    \setlength{\tabcolsep}{10pt}
    \label{tab:adaptive_heter}
        \resizebox{\linewidth}{!}{
                \begin{tabular}{ c | cc | ccccc | ccccc | c }
                \toprule
                \multicolumn{3}{c}{Dataset} & \multicolumn{5}{c}{\squirrel} & \multicolumn{5}{c}{\chameleon} \\
                \hline
                && $\Delta \rightarrow$ & 1 & 2 & 3 & 4 & 5 & 1 & 2 & 3 & 4 & 5 & Avg. Rank\\
                \cline{2-14}

                        \multirow{4}{*}{\rotatebox{90}{\textbf{Evasion}}}&\multirow{2}{*}{\rotatebox{90}{\textbf{ElasticGNN}}}& PR-BCD &\textbf{12.93 $\pm$ 4.59} &\textbf{16.13 $\pm$ 5.26} & 17.33 $\pm$ 4.51 &\textbf{19.47 $\pm$ 4.56} & 19.07 $\pm$ 6.76 & 11.73 $\pm$ 9.13 & 13.33 $\pm$ 5.11 &\textbf{16.93 $\pm$ 7.52} &\textbf{19.47 $\pm$ 10.18} &\textbf{19.60 $\pm$ 10.18}& 1.40\\
                        &&  \prbcdNA & 12.13 $\pm$ 3.81 & 15.73 $\pm$ 5.70 &\textbf{18.13 $\pm$ 5.83} & 19.33 $\pm$ 6.13 &\textbf{20.13 $\pm$ 6.21} &\textbf{12.27 $\pm$ 7.59} &\textbf{14.40 $\pm$ 8.85} & 16.67 $\pm$ 8.80 & 17.20 $\pm$ 7.55 & 18.40 $\pm$ 8.32& 1.60\\
                        \cline{2-14}
                    &\multirow{2}{*}{\rotatebox{90}{\textbf{RobustGCN}}}& PR-BCD &\textbf{37.73 $\pm$ 7.00} &\textbf{46.40 $\pm$ 6.10} &\textbf{46.27 $\pm$ 8.17} & 44.13 $\pm$ 6.25 & 47.73 $\pm$ 8.34 &\textbf{31.33 $\pm$ 8.47} & 33.07 $\pm$ 8.48 & 36.67 $\pm$ 8.77 & 38.53 $\pm$ 11.07 & 37.47 $\pm$ 7.76& 1.60\\
                        &&  \prbcdNA & 35.47 $\pm$ 7.50 & 41.47 $\pm$ 7.95 & 45.20 $\pm$ 6.36 &\textbf{44.80 $\pm$ 5.54} &\textbf{49.07 $\pm$ 7.28} & 28.93 $\pm$ 7.09 &\textbf{33.33 $\pm$ 9.70} &\textbf{40.13 $\pm$ 10.60} &\textbf{39.33 $\pm$ 10.27} &\textbf{38.67 $\pm$ 10.89}& 1.40\\
                        \cline{2-14}
                    \hline
                        \multirow{4}{*}{\rotatebox{90}{\textbf{Poison}}}&\multirow{2}{*}{\rotatebox{90}{\textbf{ElasticGNN}}}& PR-BCD &\textbf{28.00 $\pm$ 6.37} &\textbf{28.00 $\pm$ 8.04} &\textbf{29.20 $\pm$ 6.92} &\textbf{32.53 $\pm$ 8.57} & 29.07 $\pm$ 8.48 &\textbf{26.40 $\pm$ 9.86} & 28.80 $\pm$ 5.99 &\textbf{34.13 $\pm$ 11.48} & 36.27 $\pm$ 13.54 & 36.53 $\pm$ 12.86& 1.40\\
                        &&  \prbcdNA & 24.00 $\pm$ 8.42 & 24.27 $\pm$ 7.92 & 27.47 $\pm$ 9.46 & 29.87 $\pm$ 8.77 &\textbf{32.40 $\pm$ 9.69} & 24.40 $\pm$ 7.57 &\textbf{30.67 $\pm$ 9.55} & 32.00 $\pm$ 11.34 &\textbf{37.33 $\pm$ 12.11} &\textbf{37.60 $\pm$ 9.45}& 1.60\\
                        \cline{2-14}
                    &\multirow{2}{*}{\rotatebox{90}{\textbf{RobustGCN}}}& PR-BCD &\textbf{44.27 $\pm$ 10.36} &\textbf{51.20 $\pm$ 8.48} & 48.27 $\pm$ 8.51 & 48.27 $\pm$ 9.28 &\textbf{52.00 $\pm$ 10.39} & 36.53 $\pm$ 8.63 & 38.93 $\pm$ 7.96 & 42.80 $\pm$ 7.00 & 44.27 $\pm$ 12.16 & 42.93 $\pm$ 9.44& 1.70\\
                        &&  \prbcdNA & 41.07 $\pm$ 7.89 & 48.00 $\pm$ 11.56 &\textbf{52.27 $\pm$ 7.17} &\textbf{51.33 $\pm$ 8.13} & 49.73 $\pm$ 9.28 &\textbf{37.73 $\pm$ 8.10} &\textbf{41.33 $\pm$ 11.82} &\textbf{44.53 $\pm$ 10.21} &\textbf{48.40 $\pm$ 9.51} &\textbf{44.40 $\pm$ 10.53}& 1.30\\
                        \cline{2-14}
                    \hline
    \bottomrule
    \end{tabular}
    }
    \end{table}

\begin{landscape}

    \begin{table}[ht]
    
    \centering
   \caption{Effect of including node degree as criteria to select the target node on adversarial evaluation results. When evaluating adversarial attacks on target nodes, including node degree as selection criteria, is lower than evaluating those on target nodes without node degree as selection criteria, the results are shown in red; otherwise, the results are shown in blue.}
    \small
    \renewcommand\arraystretch{1.1}
    \setlength{\tabcolsep}{15pt}
    \label{tab:missclass-table-degree-impact}
        \resizebox{\linewidth}{!}{
\begin{tabular}{ c | cc | ccccc | ccccc | ccccc | c }
                \toprule \toprule
                \multicolumn{3}{c}{Dataset} & \multicolumn{5}{c}{\cora} & \multicolumn{5}{c}{\citeseer} & \multicolumn{5}{c}{\pubmed} \\
                \hline
                && $\Delta \rightarrow$ & 1 & 2 & 3 & 4 & 5 & 1 & 2 & 3 & 4 & 5 & 1 & 2 & 3 & 4 & 5 & Avg. Rank Difference \\
                \cline{2-19}

                        \multirow{28}{*}{\rotatebox{90}{\textbf{Evasion}}}&\multirow{7}{*}{\rotatebox{90}{\textbf{GCN}}}& \propseRND & \negative{-3.30} & \negative{-3.63} & \negative{-1.63} & \positive{0.10} & \negative{-5.87} & \negative{-5.13} & \negative{-1.50} & \positive{5.37} & \negative{-2.60} & \positive{0.13} & \negative{-5.07} & \negative{-9.20} & \negative{-11.13} & \negative{-10.03} & \negative{-10.23}& 0.00\\
                        && FGA & \negative{-11.88} & \negative{-7.37} & \negative{-7.27} & \negative{-6.07} & \negative{-5.83} & \negative{-16.78} & \negative{-19.25} & \negative{-10.23} & \negative{-6.47} & \negative{-9.98} & \negative{-21.57} & \negative{-19.57} & \negative{-13.83} & \negative{-13.43} & \negative{-13.83}& \positive{1.00}\\
                        && NETTACK & \negative{-7.15} & \positive{0.60} & \negative{-2.45} & \negative{-2.62} & \negative{-3.02} & \negative{-12.12} & \negative{-5.67} & \negative{-10.13} & \negative{-6.32} & \negative{-8.52} & \negative{-23.93} & \negative{-10.43} & \negative{-10.70} & \negative{-10.50} & \negative{-13.83}& \negative{-1.20}\\
                        &&  PGD & \negative{-11.67} & \negative{-5.50} & \negative{-5.07} & \negative{-4.65} & \negative{-6.80} & \negative{-18.53} & \negative{-13.23} & \negative{-11.87} & \negative{-10.98} & \negative{-10.53} & \negative{-22.70} & \negative{-22.50} & \negative{-26.90} & \negative{-24.20} & \negative{-25.90}& \positive{1.00}\\
\bottomrule \bottomrule
    \end{tabular}
    }
    \end{table}
\end{landscape}

\begin{landscape}
    \begin{table}[ht]
    \centering

    \caption{Effect of model selection on adversarial evaluation results. When evaluating adversarial attacks on victim models with a configuration obtained from model selection, which is higher than evaluating on victim models with a fixed configuration, the results are shown  in blue; the results are shown in red otherwise.}
    \small
    \renewcommand\arraystretch{1}
    \setlength{\tabcolsep}{10pt}
    \label{tab:missclass-table-model-selection-impact}
        \resizebox{\linewidth}{!}{
                \begin{tabular}{ c | cc | ccccc | ccccc | ccccc | c }
                \toprule \toprule
                \multicolumn{3}{c}{Dataset} & \multicolumn{5}{c}{\cora} & \multicolumn{5}{c}{\citeseer} & \multicolumn{5}{c}{\pubmed} \\
                \hline
                && $\Delta \rightarrow$ & 1 & 2 & 3 & 4 & 5 & 1 & 2 & 3 & 4 & 5 & 1 & 2 & 3 & 4 & 5 & Avg. Rank Difference \\
                \cline{2-19}

                        \multirow{28}{*}{\rotatebox{90}{\textbf{Evasion}}}&\multirow{7}{*}{\rotatebox{90}{\textbf{GCN}}}& \propseRND & \negative{-2.67} & \positive{2.80} & 0.00 & \negative{-0.27} & \negative{-2.40} & \positive{4.00} & \positive{6.93} & \positive{2.27} & \positive{1.87} & \negative{-1.07} & 0.00 & \positive{0.13} & \negative{-4.27} & \negative{-9.20} & \negative{-5.73}& 0.00\\     
                        && FGA & \positive{0.67} & \positive{0.80} & \negative{-0.93} & \positive{0.27} & \positive{2.13} & \negative{-1.33} & \negative{-7.20} & \negative{-3.47} & \negative{-1.07} & \negative{-0.67} & \negative{-1.07} & \negative{-0.27} & \negative{-0.53} & \positive{0.40} & \negative{-0.67}& 0.00\\
                        && NETTACK & \positive{0.27} & \negative{-3.07} & \negative{-2.27} & \negative{-0.27} & \positive{0.53} & \negative{-1.87} & \negative{-8.80} & \negative{-8.13} & \negative{-0.53} & \negative{-1.20} & \negative{-2.80} & \positive{2.93} & \negative{-0.40} & \positive{0.27} & \negative{-0.13}& \positive{0.33}\\
                        &&  PGD & \positive{1.60} & \positive{11.47} & \positive{13.60} & \positive{15.33} & \positive{13.87} & \negative{-2.00} & \negative{-0.27} & \negative{-0.80} & \positive{0.80} & \positive{2.13} & \negative{-2.00} & \negative{-2.13} & \negative{-4.80} & \negative{-3.73} & \negative{-4.40}& \negative{-0.13}\\
\bottomrule \bottomrule
    \end{tabular}
    }
    \end{table}

\end{landscape}
\begin{landscape}
\begin{table}[ht]

    \centering
    \caption{Performance of RND attack and our proposed random-based baseline, \propseRND, on defense and non-defense victim models used in evaluating adversarial attacks on three datasets.}
    \small
    \renewcommand\arraystretch{3.2}
    \setlength{\tabcolsep}{15pt}
    \label{tab:rnd_vs_rndv2}
        \resizebox{\linewidth}{!}{
                \begin{tabular}{ c | cc | ccccc | ccccc | ccccc  }
                \toprule \toprule
                \multicolumn{3}{c}{Dataset} & \multicolumn{5}{c}{\cora} & \multicolumn{5}{c}{\citeseer} & \multicolumn{5}{c}{\pubmed} \\
                \hline
                && $\Delta \rightarrow$ & 1 & 2 & 3 & 4 & 5 & 1 & 2 & 3 & 4 & 5 & 1 & 2 & 3 & 4 & 5 \\
                \cline{2-18}

                        \multirow{14}{*}{\rotatebox{90}{\textbf{Evasion}}}&\multirow{2}{*}{\rotatebox{90}{\textbf{GCN}}}& \propseRND &\textbf{13.20 $\pm$ 4.33} &\textbf{22.53 $\pm$ 6.95} &\textbf{29.20 $\pm$ 5.60} &\textbf{32.93 $\pm$ 6.67} &\textbf{32.13 $\pm$ 5.04} &\textbf{15.20 $\pm$ 4.20} &\textbf{28.67 $\pm$ 12.39} &\textbf{36.53 $\pm$ 9.09} &\textbf{37.73 $\pm$ 9.44} &\textbf{42.13 $\pm$ 7.11} &\textbf{10.93 $\pm$ 3.01} &\textbf{15.47 $\pm$ 4.87} &\textbf{17.20 $\pm$ 6.58} &\textbf{18.80 $\pm$ 5.28} &\textbf{23.60 $\pm$ 9.63}\\
                        && Random & 13.20 $\pm$ 4.89 & 20.13 $\pm$ 5.04 & 24.27 $\pm$ 4.95 & 29.47 $\pm$ 5.78 & 30.40 $\pm$ 6.90 & 14.00 $\pm$ 4.34 & 28.27 $\pm$ 8.24 & 35.33 $\pm$ 7.84 & 37.47 $\pm$ 7.03 & 36.00 $\pm$ 8.25 & 10.27 $\pm$ 3.69 & 14.40 $\pm$ 5.77 & 16.27 $\pm$ 3.61 & 17.73 $\pm$ 5.55 & 15.07 $\pm$ 4.77\\
                        \cline{2-18}
                    &\multirow{2}{*}{\rotatebox{90}{\textbf{GIN}}}& \propseRND &\textbf{20.13 $\pm$ 7.35} &\textbf{37.07 $\pm$ 15.27} &\textbf{42.67 $\pm$ 8.80} &\textbf{43.20 $\pm$ 10.05} &\textbf{43.73 $\pm$ 12.02} &\textbf{16.80 $\pm$ 10.87} &\textbf{33.47 $\pm$ 15.76} &\textbf{35.73 $\pm$ 14.94} &\textbf{38.53 $\pm$ 13.17} &\textbf{43.33 $\pm$ 12.57} &\textbf{13.33 $\pm$ 5.64} &\textbf{23.07 $\pm$ 12.89} &\textbf{26.67 $\pm$ 12.75} &\textbf{22.40 $\pm$ 10.26} &\textbf{22.27 $\pm$ 11.54}\\
\bottomrule \bottomrule
    \end{tabular}
    }
    \end{table}
    \end{landscape}

\begin{landscape}

\begin{table}[ht]
    \centering
    \caption{{\textbf{Vanilla Homophily Results.} Evaluating six adversarial attacks( \textbf{features attack variants}) on four vanilla attack victim models - Miss-classification rate (\%). Higher is better.  Best performance in \textbf{bold}, second best \underline{underlined}}}.
    \small
    \renewcommand\arraystretch{1.2}
    \setlength{\tabcolsep}{3pt}
    \label{tab:feat-attack}
        \resizebox{\linewidth}{!}{
                \begin{tabular}{ c | cc | ccccc | ccccc | ccccc | c }
                \toprule
                \multicolumn{3}{c}{Dataset} & \multicolumn{5}{c}{\cora} & \multicolumn{5}{c}{\citeseer} & \multicolumn{5}{c}{\pubmed} \\
                \hline
                && $\Delta \rightarrow$ & 1 & 2 & 3 & 4 & 5 & 1 & 2 & 3 & 4 & 5 & 1 & 2 & 3 & 4 & 5 & Avg. Rank\\
                \cline{2-19}
                    
                        \multirow{20}{*}{\rotatebox{90}{\textbf{Evasion}}}&\multirow{5}{*}{\rotatebox{90}{\textbf{GCN}}}& FGA & 27.20 $\pm$ 4.13 & 46.80 $\pm$ 6.27 & 56.40 $\pm$ 3.87 &\underline{61.87 $\pm$ 3.81} &\underline{64.13 $\pm$ 4.50} & 24.00 $\pm$ 2.73 & 34.27 $\pm$ 6.23 & 51.20 $\pm$ 7.51 & 52.80 $\pm$ 9.22 &\underline{60.67 $\pm$ 3.18} &\textbf{36.80 $\pm$ 3.53} &\underline{49.33 $\pm$ 7.24} &\underline{56.00 $\pm$ 3.70} &\underline{57.07 $\pm$ 2.81} &\textbf{59.20 $\pm$ 1.26}& 2.80\\
                        && NETTACK & 29.33 $\pm$ 4.94 &\underline{49.87 $\pm$ 5.68} &\underline{57.07 $\pm$ 5.44} &\textbf{62.13 $\pm$ 4.75} & 63.33 $\pm$ 4.51 & 25.07 $\pm$ 6.04 &\underline{45.73 $\pm$ 8.28} &\underline{54.13 $\pm$ 5.37} &\underline{57.73 $\pm$ 3.84} & 59.20 $\pm$ 8.84 &\underline{35.20 $\pm$ 2.60} &\textbf{54.27 $\pm$ 2.71} &\textbf{58.27 $\pm$ 1.83} &\textbf{59.07 $\pm$ 1.98} &\underline{58.93 $\pm$ 1.83}& 2.00\\
                        &&  PGD &\underline{29.60 $\pm$ 4.42} & 47.07 $\pm$ 4.20 & 55.87 $\pm$ 4.56 & 57.20 $\pm$ 4.46 & 59.07 $\pm$ 4.06 &\underline{27.73 $\pm$ 5.12} & 40.53 $\pm$ 8.23 & 46.13 $\pm$ 7.11 & 52.93 $\pm$ 5.70 & 54.93 $\pm$ 6.27 & 35.07 $\pm$ 2.37 & 38.80 $\pm$ 4.33 & 43.20 $\pm$ 4.52 & 45.47 $\pm$ 6.57 & 46.27 $\pm$ 5.34& 3.87\\
\hline
    \bottomrule
    \end{tabular}
    }
    \end{table}
    \end{landscape}

    \begin{table}[ht]
    \centering
    \caption{{\textbf{Ablation study}.Evaluating two variants of \propseRND: \propseRND - Add only ( Degree effect) and \propseRND - Remove only ($\mathcal{L}_1$ effect) on PNA and GRAND uner \cora dataset.  Best performance in \textbf{bold}, second best \underline{underlined}}}
    \small
    \renewcommand\arraystretch{1.2}
    \setlength{\tabcolsep}{3pt}
    \label{tab:rnd_ablation}
        \resizebox{\linewidth}{!}{
                \begin{tabular}{ c | cc | ccccc | c }
                \toprule
                \multicolumn{3}{c}{Dataset} & \multicolumn{5}{c}{\cora} \\
                \hline
                && $\Delta \rightarrow$ & 1 & 2 & 3 & 4 & 5 & Avg. Rank\\
                \cline{2-9}

                        \multirow{8}{*}{\rotatebox{90}{\textbf{Evasion}}}&\multirow{4}{*}{\rotatebox{90}{\textbf{PNA}}}& Random & 29.47 $\pm$ 11.72 & 35.33 $\pm$ 9.90 & 31.87 $\pm$ 11.17 & 37.20 $\pm$ 12.87 & 39.07 $\pm$ 10.47& 4.00\\
                        && \propseRND - Add only & 31.73 $\pm$ 11.00 &\underline{46.27 $\pm$ 18.00} &\underline{51.07 $\pm$ 15.69} &\textbf{49.07 $\pm$ 15.34} &\textbf{53.60 $\pm$ 11.89}& 1.80\\
                        && \propseRND - Remove only &\underline{32.40 $\pm$ 12.49} &\textbf{46.80 $\pm$ 17.64} &\textbf{53.87 $\pm$ 16.34} &\underline{47.20 $\pm$ 14.16} & 51.87 $\pm$ 10.99& 1.80\\
                        && \propseRND &\textbf{33.20 $\pm$ 8.81} & 41.60 $\pm$ 12.86 & 47.20 $\pm$ 9.44 & 45.20 $\pm$ 12.14 &\underline{53.07 $\pm$ 15.36}& 2.40\\
                        \cline{2-9}
                    &\multirow{4}{*}{\rotatebox{90}{\textbf{GRAND}}}& Random & 27.33 $\pm$ 7.99 & 41.87 $\pm$ 7.42 & 40.67 $\pm$ 11.48 & 42.93 $\pm$ 9.47 & 45.47 $\pm$ 9.66& 4.00\\ 
                        && \propseRND - Add only &\textbf{62.80 $\pm$ 7.92} &\textbf{70.53 $\pm$ 8.05} &\underline{69.60 $\pm$ 10.26} &\underline{68.27 $\pm$ 12.14} &\underline{64.93 $\pm$ 13.69}& 1.60\\
\hline
    \bottomrule
    \end{tabular}
    }
    \end{table}


\end{document}